\documentclass[letterpaper]{article}
\PassOptionsToPackage{numbers, compress}{natbib}
\usepackage[preprint]{neurips_2025}
\usepackage{times} 
\usepackage{helvet} 
\usepackage{array}
\usepackage{courier} 
\usepackage[hyphens]{url} 
\usepackage{graphicx} 
\usepackage{graphicx} 
\usepackage{natbib} 
\usepackage{caption} 
\usepackage{amsmath, amssymb, graphicx, booktabs} 
\usepackage{algorithm, algpseudocode}
\usepackage{subcaption} 
\usepackage{multirow} 
\usepackage{hyperref}
\title{MultiDiffNet: A Multi-Objective Diffusion Framework for Generalizable Brain Decoding}

\author{
  Mengchun Zhang\thanks{Equal contribution.} \\
  University of Pittsburgh\\
  Pittsburgh, PA, USA \\
  \texttt{mengchuz@andrew.cmu.edu} \\
  \And
  Kateryna Shapovalenko\footnotemark[1] \\
  Carnegie Mellon University \\
  Pittsburgh, PA, USA \\
  \texttt{kshapova@alumni.cmu.edu} \\
  \And
  Yucheng Shao \\
  Carnegie Mellon University \\
  Pittsburgh, PA, USA \\
  \texttt{yshao3@andrew.cmu.edu} \\
  \And
  Eddie Guo \\
  Carnegie Mellon University \\
  Pittsburgh, PA, USA \\
  \texttt{yuzhiguo@andrew.cmu.edu} \\
  \And
  Parusha Pradhan \\
  University of Pittsburgh \\
  Pittsburgh, PA, USA \\
  \texttt{pap203@pitt.edu} \\
}

\begin{document}

\maketitle

\renewcommand{\thefootnote}{\fnsymbol{footnote}}
\footnotetext[1]{A preliminary version appeared at 39th Conference on Neural Information Processing Systems (NeurIPS 2025) Workshop: Foundation Models. Project code: \url{https://github.com/eddieguo-1128/DualDiff}.}

\begin{abstract} 

Neural decoding from electroencephalography (EEG) remains fundamentally limited by poor generalization to unseen subjects, driven by high inter-subject variability and the lack of large-scale datasets to model it effectively. Existing methods often rely on synthetic subject generation or simplistic data augmentation, but these strategies fail to scale or generalize reliably. We introduce \textit{MultiDiffNet}, a diffusion-based framework that bypasses generative augmentation entirely by learning a compact latent space optimized for multiple objectives. We decode directly from this space and achieve state-of-the-art generalization across various neural decoding tasks using subject and session disjoint evaluation. We also curate and release a unified benchmark suite spanning four EEG decoding tasks of increasing complexity (SSVEP, Motor Imagery, P300, and Imagined Speech) and an evaluation protocol that addresses inconsistent split practices in prior EEG research. Finally, we develop a statistical reporting framework tailored for low-trial EEG settings. Our work provides a reproducible and open-source foundation for subject-agnostic EEG decoding in real-world BCI systems.

\end{abstract}

\section{Introduction}

Electroencephalography (EEG) is a widely used modality in brain–computer interfaces (BCIs), supporting applications from assistive communication to cognitive monitoring. Deep learning has improved decoding across motor imagery, SSVEP, and speech tasks~\cite{gu2025cltnet,ahmadi2025universal,lee2022eeg}, yet generalizing to unseen subjects remains challenging due to high inter-subject variability and limited data~\cite{huang2023discrepancy,barmpas2023improving}. 

Subject-specific models require extensive per-user calibration~\cite{hartmann2018eeg,luo2024diffusion}, while multi-subject models struggle to generalize ~\cite{rommel2022data,liu2022subject,wu2016online}. The alternative is to use two-stage pipelines that generate EEG via GANs or diffusion and then train decoders~\citep{hartmann2018eeg,torma2025generative}, but they suffer from low realism, artifact transfer, and inefficiencies.

We propose \textit{MultiDiffNet}, a unified multi-objective diffusion framework that learns a shared latent space, eliminating the need for synthetic augmentation and enhancing generalization. To benchmark progress, we release a curated suite spanning SSVEP, Motor Imagery, P300, and Imagined Speech tasks, with standardized subject- and session-disjoint evaluation. We also develop a statistical reporting protocol tailored for low-trial EEG research, addressing a persistent gap in reproducibility. 

\begin{figure}[htbp]
\centering
\includegraphics[width=0.95\textwidth]{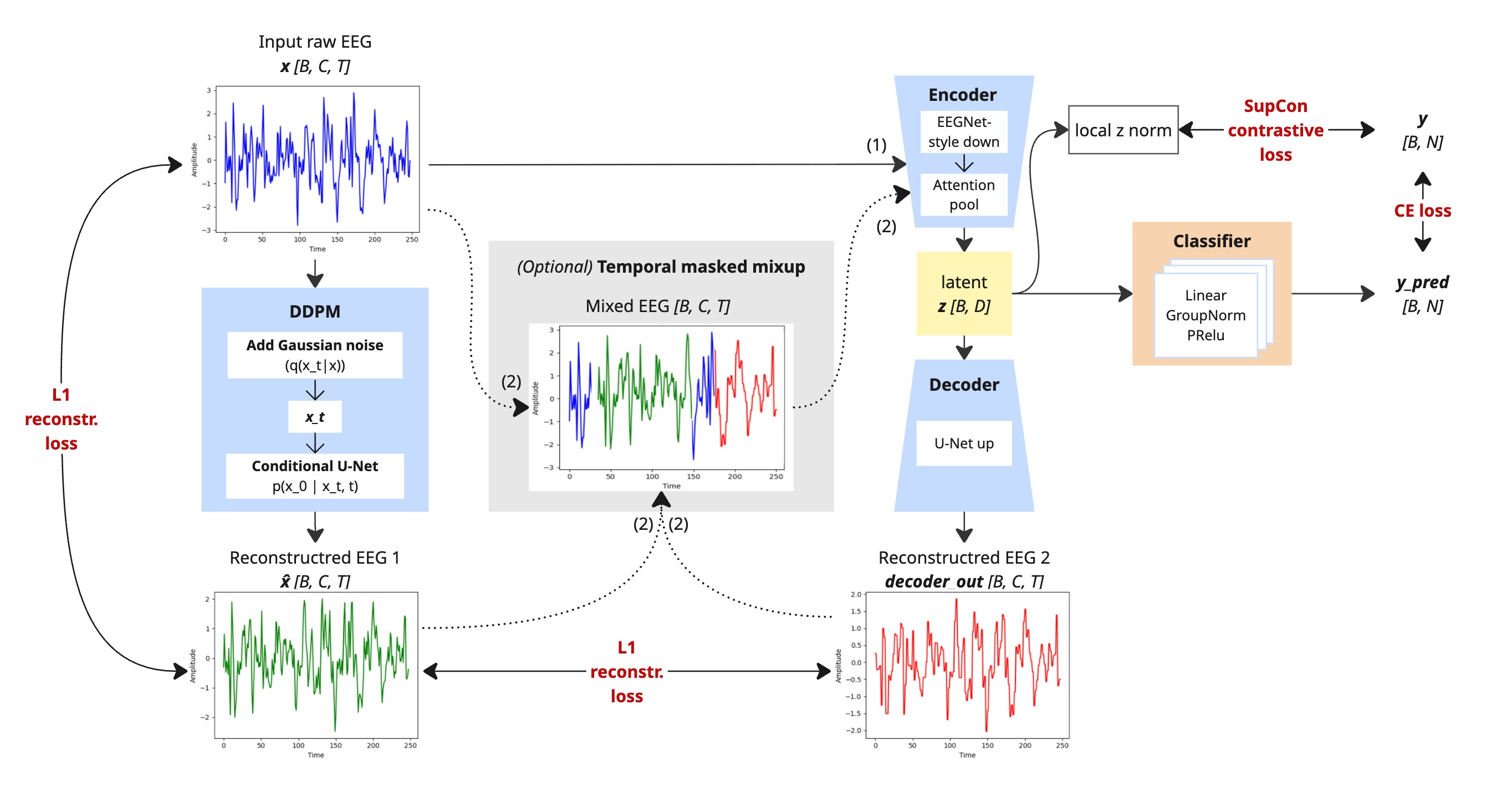}
\caption{Overview of the \textit{MultiDiffNet} that jointly optimizes a conditional DDPM, a contrastive encoder, and a generative decoder through a shared latent space \textit{z}. The encoder produces discriminative features used for both classification and contrastive learning, while the decoder and DDPM reconstruct the input signal. An optional \textit{temporal masked mixup} module stochastically blends the original, DDPM-denoised, and decoder-reconstructed EEG to improve representation quality.}
\label{fig:model}
\end{figure}

\section{Related work} 

\textbf{EEG Decoding and Generalization} EEG decoding has evolved from handcrafted features to deep architectures, with EEGNet emerging as a widely adopted baseline due to its efficient depthwise–separable convolutions and lightweight design~\citep{lawhern2018eegnet, zhang2024improvingssvepbcispellers}. Recent models explore transformers~\citep{auster2025pennythoughtsdecodingspeech, liao2025advancing,song2022eeg} and graph neural networks~\citep{tang2024spatial,hu2023regional}, but EEGNet remains favored for its robustness and simplicity. A key limitation is poor cross-subject generalization, with 20-40\% accuracy drops despite strong within-subject performance~\citep{huang2023discrepancy,barmpas2023improving}. Attempts to address this require expensive calibration~\citep{rommel2022data,liu2022subject,wu2016online}. Scalable BCIs require subject-agnostic models that generalize without per-user retraining.

\textbf{Diffusion Models for EEG} Denoising Diffusion Probabilistic Models (DDPMs) model data distributions via iterative denoising and outperform GANs in EEG synthesis by avoiding mode collapse~\citep{tosato2023eegsyntheticdatageneration,ho2020denoising}. Recent enhancements, such as reinforcement learning~\citep{an2024enhancingeegsignalgeneration} and progressive distillation~\citep{torma2025generative}, have further improved realism and sampling speed. Diff-E~\citep{kim2023diff} extended diffusion to imagined-speech decoding via joint reconstruction and classification, but remained task-specific and did not address cross-subject generalization. Broader research suggests that combining generative and discriminative objectives yields stronger representations~\citep{chow2024unified,grathwohl2019your}, yet EEG models typically optimize only one. We explore this joint learning paradigm across diverse EEG tasks, aiming to learn generalizable representations that capture both signal structure and task-relevant information.

\textbf{Mixup Methods} Signal-level augmentation has evolved from basic jittering and filtering to temporal, spectral, and channel-wise mixup~\citep{luo2025diffusion,liu2025mixeeg,kim2021specmix,pei2021data,zhang2017mixup}, but many variants introduce unrealistic artifacts that hinder generalization~\citep{chen2025decodingeegspeechperception}. This motivates our systematic evaluation of weighted and temporal input mixup across encoder layers, along with latent‑space mixing

\textbf{Evaluation Strategies} Effective cross-subject EEG decoding requires both rigorous training strategies and standardized evaluation. Leave-one-subject-out (LOSO) validation remains common but is computationally intensive and impractical for real-time deployment~\citep{del2025role,chen2025leveraging,zhao2024ctnet,barmpas2023improving,kunjan2021necessity}, while simpler subject splits often neglect session independence and true seen/unseen separation~\citep{zhang2023subject}. We address it in our work by introducing a standardized subject- and session-disjoint evaluation. 

\section{Methodology}

\subsection{MultiDiffNet architecture} 

\textit{MultiDiffNet} is a modular architecture designed to jointly optimize classification, reconstruction, and contrastive structure learning from EEG signals. It consists of a Denoising Diffusion Probabilistic Model (DDPM), a discriminative encoder, a generative decoder, and a classifier (Figure~\ref{fig:model}). 

Given a raw EEG signal \( x \in \mathbb{R}^{C \times T} \), where \( C \) is the number of EEG channels and \( T \) is the number of timepoints, the model processes the input in two parallel paths. First, the DDPM denoises the signal via a learned reverse diffusion process, producing a refined version \( \hat{x} \in \mathbb{R}^{C \times T} \). Simultaneously, the same input \( x \) is passed through an EEGNet-based encoder (See Section~\ref{sec:eegnet-style encoder}) to extract a latent representation \( z \in \mathbb{R}^D \), where \( D \) is the embedding dimension. The latent vector \( z \) is then used for two purposes: (1) it is passed to a lightweight decoder to reconstruct the denoised signal \( \hat{x} \), resulting in a reconstruction \( x_{\text{dec}} \in \mathbb{R}^{C \times T} \); and (2) it is passed to a fully connected classification head to predict class logits \( \hat{y} \in \mathbb{R}^K \), where \( K \) is the number of classes.

To further structure the latent space, \( z \) is locally normalized (Section~\ref{sec:latentnorm}) and then projected to \( z_{\text{proj}} \in \mathbb{R}^{D'} \), which is optimized with a supervised contrastive loss. All classification and reconstruction are performed directly from \( z \), without relying on generated augmentations.

We performed an extensive ablation study across architectural variants, modifying the presence of DDPMs, encoder inputs, decoder pathways, classifier heads, and loss terms. The configuration described here reflects the best-performing combination. 

\subsection{EEGNet-style encoder with attention pool}
\label{sec:eegnet-style encoder}

Given EEGNet's demonstrated effectiveness across multiple EEG decoding tasks, we adapt its architecture as our discriminative encoder, hypothesizing that its proven feature extraction capabilities can produce powerful latent representations $z$ for our multi-objective framework. Our encoder extracts multi-scale features $(dn_1, dn_2, dn_3)$ from different layers and applies attention pooling:
\[
z = \text{AttentionPool}(dn_3) \in \mathbb{R}^D,
\]

\subsection{Subject-wise latent normalization}
\label{sec:latentnorm}

To mitigate inter-subject variability, we apply subject-wise normalization on the encoder output $z$:
\[
z_{\text{norm}} = \frac{z - \mu_s}{\sigma_s},
\]
where $\mu_s$ and $\sigma_s$ denote the mean and standard deviation computed per subject $s$ using a subset of training trials. During evaluation, we adopt a two-mode strategy: for seen subjects, normalization uses pre-computed statistics from their training data; for unseen subjects, statistics are estimated on-the-fly using their own calibration trials, simulating realistic deployment scenarios. 

\begin{algorithm}
\small
\caption{Temporal Masked Mixup}
\label{alg:temporal_mixup}
\begin{algorithmic}[1]
\State Initialize a binary mask $M \in \{0,1\}^{C \times T}$ with all zeros.
\State Flip each $0$ in $M$ to $1$ with probability $p = 0.01$.
\For{each position in $M$ with value $1$}
    \State Expand to a temporal window of random length 
           (uniform between min and max size).
\EndFor
\State Flip each $1$ in $M$ to $-1$ with:
    \begin{itemize}
        \item Fixed probability $0.5$ (\textbf{fixed ratio}), or
        \item Probability drawn from $\text{Beta}(0.2, 0.2)$ each epoch (\textbf{random ratio}).
    \end{itemize}
\State Apply the final mask:
    \begin{itemize}
        \item $0 \rightarrow x$ (original input)
        \item $1 \rightarrow \hat{x}$ (DDPM output)
        \item $-1 \rightarrow x_{\text{dec}}$ (decoder output)
    \end{itemize}
\end{algorithmic}
\end{algorithm}

\subsection{Mixup strategies}

Mixup strategies can improve robustness in low-trial EEG decoding. However, standard mixup techniques may not fully exploit the structure of neural time series. We therefore explore two complementary strategies: \textit{Weighted Average Mixup} and a novel \textit{Temporal Masked Mixup}. \textit{Weighted Average Mixup} performs linear interpolation between the original EEG input $x$, the DDPM-denoised output $\hat{x}$, and the decoder reconstruction $x_{\text{dec}}$. We investigate multiple integration points in the model: \textbf{(0)} Input-level mixup, \textbf{(1-3)} Mixup after encoder layers 1, 2, or 3, respectively, \textbf{(4)} Mixup after the final attention pooling layer. To address the limitations of global interpolation, we propose \textit{Temporal Masked Mixup}, which perturbs only localized segments of the input time series while preserving surrounding structure. See Algorithm~\ref{alg:temporal_mixup} for pseudocode. 

\subsection{Loss functions}

\textit{MultiDiffNet} is trained using a weighted sum of three objectives:

\[
\mathcal{L}_{\text{total}} =
\underbrace{\alpha\, \mathcal{L}_{\text{CE/MSE}}(\hat{y}, y)}_{\text{classification}} +
\underbrace{\beta\, \mathcal{L}_{\text{L1}}(x_{\text{dec}}, \hat{x})}_{\text{reconstruction}} +
\underbrace{\gamma\, \mathcal{L}_{\text{SupCon}}(z_{\text{proj}}, y)}_{\text{contrastive}}
\]

We fix $\alpha = 1.0$ and progressively scale $\beta$ and $\gamma$ to stabilize training:

\[
\beta = \min\left(1.0, \frac{\text{epoch}}{100}\right) \cdot 0.05,\quad
\gamma = \min\left(1.0, \frac{\text{epoch}}{50}\right) \cdot 0.2
\]

Details on loss formulation and weighting strategies are provided in the Appendix.

\subsection{Evaluation metrics} 

We evaluate model performance primarily using downstream classification accuracy, which quantifies the proportion of correctly classified EEG samples. Accuracy is defined as:

\[
\text{Accuracy} = \frac{TP + TN}{TP + TN + FP + FN}
\]

where \( TP \), \( TN \), \( FP \), and \( FN \) denote true positives, true negatives, false positives, and false negatives, respectively.
In addition, we report F1 score, precision, recall, and AUC for a more comprehensive evaluation; detailed formulas and results are provided in the Appendix.    
\subsection{Trend-level statistical reporting framework}

Conventional $p$-values often fail under the high-variance, low-trial, subject-disjoint conditions of EEG decoding. To address this, we introduce a robust trend-level statistical framework (detailed in the Appendix) that synthesizes effect sizes, cross-seed consistency, and Bayesian posterior probabilities. This allows us to detect systematic, reproducible gains even when classical significance tests return null results. Our approach represents a principled shift toward reproducible, evidence-based model evaluation in brain decoding. While this framework enhances reproducibility, it is not meant to substitute conventional $p$-value testing. Instead, it addresses a well-documented limitation: in low-trial, high-variance EEG decoding, even systematic improvements may fail to reach arbitrary significance thresholds. 

\section{Experiments and results}

\subsection{Benchmark dataset suite} 

\begin{figure}[htbp]
\centering
\includegraphics[width=1.0\textwidth]{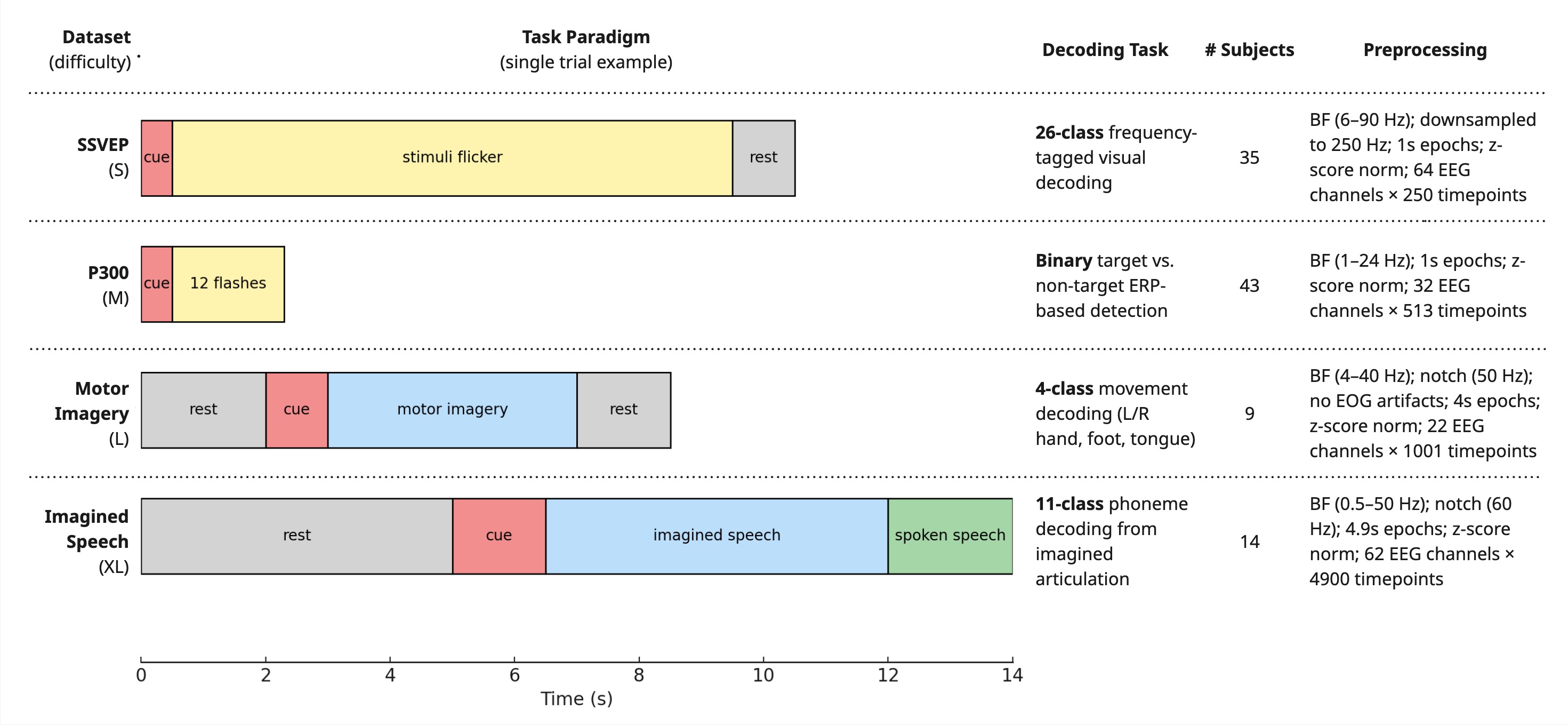}
\caption{Overview of four EEG datasets ranked by task difficulty from easiest (top) to hardest (bottom). Task paradigms and preprocessing details are adapted from the original publications: SSVEP~\cite{wang2017benchmark}, P300~\cite{korczowski2019brain}, Motor Imagery~\cite{tangermann2012review}, and Imagined Speech~\cite{zhao2015classifying}.}
\label{fig:datasets}
\end{figure}

We curated four diverse EEG benchmarks (SSVEP, P300, Motor Imagery, and Imagined Speech), spanning increasing decoding difficulty. Each dataset is split into train, val, and two test sets: a seen-subject (intra-subject) split and an unseen-subject (cross-subject) split. This standardized protocol enables rigorous evaluation of both personalization and generalization, addressing the inconsistent and often unrealistic split practices prevalent in prior EEG research, where models are evaluated on mixed subject data or using computationally expensive LOSO. 

\subsection{Baselines}

We benchmarked our model against a diverse set of carefully selected baselines to ensure robust and fair comparisons. Our selection criteria were twofold: (i) prioritize architectures that are widely used for generalization to unseen subjects or sessions, and (ii) cover the main inductive biases found in EEG decoding, such as spatial filtering, temporal modeling, and attention mechanisms.

Specifically, we include: 
(1) \textbf{EEGNet} \citep{lawhern2018eegnet}, a compact depthwise-separable CNN that is widely adopted for cross-subject generalization due to its strong accuracy–efficiency trade-off; 
(2) \textbf{ShallowFBCSPNet} \citep{shallowfbcspnet_schirrmeister2017}, which implements learnable filter-bank Common Spatial Patterns (CSP) to extract frequency–spatial features; 
(3) \textbf{TIDNet} \citep{tidnet_kostas2020}, which introduces dilated convolutions and residual connections to improve robustness under subject shift; 
(4) \textbf{EEGConformer} \citep{eegconformer_song2022}, which combines a convolutional front-end with self-attention to model both local spatial structure and global temporal context; and 
(5) \textbf{EEGTCNet} \citep{eegtcnet_ingolfsson2020}, a temporal convolutional network tailored for EEG that emphasizes causal and dilated temporal modeling, offering complementary inductive bias to purely spatial–spectral models.

All models are evaluated using identical input windows of shape $(C, T)$, and trained with a unified global training schedule to ensure comparability. Public implementations and recommended hyperparameters are used where available, with no method-specific tuning.

\subsection{Generalization performance} 

\begin{figure}[htbp]
\centering
\includegraphics[width=1.0\textwidth]{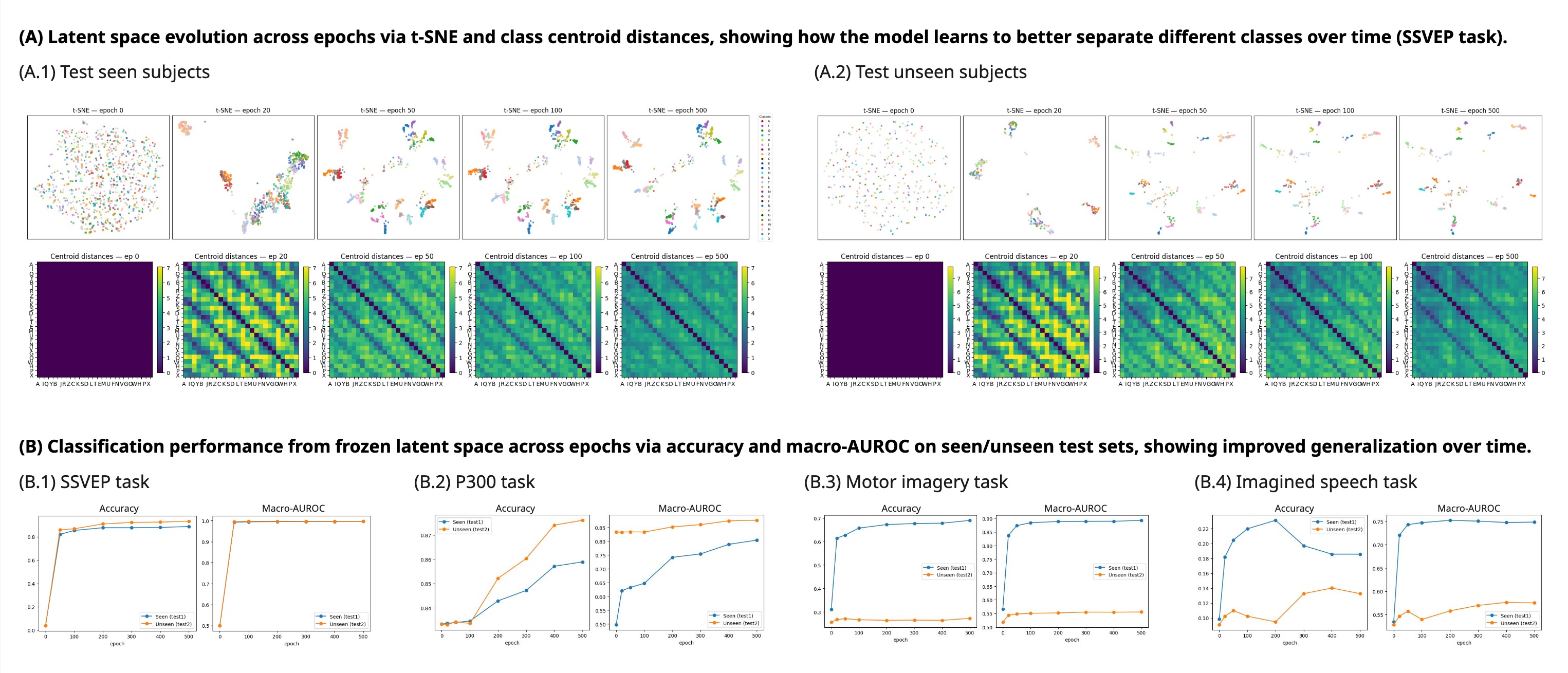}
\caption{(A) Visualization of latent space across training epochs. (B) Downstream classification performance from frozen latent representations.}
\label{fig:embeddings}
\end{figure}

\textit{MultiDiffNet} helps with generalization. Unlike raw EEG representations, where class
boundaries blur due to subject-specific noise, our learned latent space forms clearly separable,
label-aligned clusters (Figure~\ref{fig:embeddings}). This structured representation enables robust
decoding across subjects. As shown in Table~\ref{tab:final_results}, \textit{MultiDiffNet} consistently
reduces the seen--unseen accuracy gap across all tasks. In SSVEP, it lifts cross-subject accuracy
from 81.08\% (EEGNet baseline) to 84.72\%, further boosted to 85.25\% with Temporal Masked
Mixup. For comparison, other representative architectures such as ShallowFBCSPNet (58.87\%),
EEGConformer (51.92\%), TIDNet (25.96\%), and EEGTCNet (49.57\%) fall well behind, highlighting the robustness of our latent-space design.

Even in the low-SNR regime of Imagined Speech, \textit{MultiDiffNet} improves cross-subject
accuracy from 10.61\% (EEGNet) to 12.12\%, while simultaneously achieving a much larger
gain on seen-subject accuracy (11.26\% $\rightarrow$ 17.57\%). Other baselines such as ShallowFBCSPNet
(10.48/13.78\%), EEGConformer (9.21/10.62\%), TIDNet (10.35/9.10\%), and EEGTCNet
(10.10/12.64\%) hover close to chance level on both splits, further highlighting the robustness
of our approach. For such a challenging task, even modest absolute gains are meaningful, as they
can indicate more reliable signal extraction under extreme noise conditions. On Motor Imagery,
\textit{MultiDiffNet} also surpasses most baselines on unseen accuracy, e.g., outperforming TIDNet
(34.42\%) and EEGTCNet (32.99\%), while maintaining competitive seen accuracy (57.69\% vs.\
44.27\% for TIDNet and 58.85\% for EEGTCNet). Although it remains slightly below EEGNet
(46.18/67.01\%), this is likely due to ceiling effects and dataset scale.

\begin{table*}[t]
\centering
\small 
\renewcommand{\arraystretch}{0.8}
\caption{Final results across tasks and models. Accuracy is reported for both seen-subject (intra-subject) and unseen-subject (cross-subject) test splits. Tasks are ranked by task difficulty. \textit{Stars denote win percentage:} \textbf{***}~$\geq$~80\%, \textbf{**}~$\geq$~60\%, \textbf{*}~$\geq$~40\%. Detailed results are in the Appendix.}
\label{tab:final_results}
\begin{tabular}{p{1.4cm} p{3.8cm} c c p{2.1cm} p{2.1cm}}
\toprule
\textbf{Task} & \textbf{Model} & \textbf{Subj.} & \textbf{Classes} & \textbf{Seen Acc. (\%)} & \textbf{Unseen Acc. (\%)} \\
\midrule 
\multirow{4}{*}{\textbf{SSVEP}} 
& ShallowFBCSPNet & 35 & 26 & 69.58 $\pm$ 1.30\textsuperscript{*} & 58.87 $\pm$ 9.37\textsuperscript{*} \\
& EEGConformer & 35 & 26 & 66.98 $\pm$ 2.83 & 51.92 $\pm$ 9.06 \\
& TIDNet & 35 & 26 & 28.01 $\pm$ 4.12 & 25.96 $\pm$ 5.29 \\
& EEGTCNet & 35 & 26 & 58.31 $\pm$ 4.02 & 49.57 $\pm$ 9.14 \\
& EEGNet & 35 & 26 & 89.16 $\pm$ 0.57\textsuperscript{***} & 81.08 $\pm$ 9.16\textsuperscript{**} \\
& \textbf{MultiDiffNet} & 35 & 26 & 85.08 $\pm$ 1.53\textsuperscript{**} & \textbf{84.72 $\pm$ 6.03}\textsuperscript{***} \\
& \textbf{MultiDiffNet + Mixup} & 35 & 26 & 86.79 $\pm$ 1.75\textsuperscript{***} & \textbf{85.25 $\pm$ 6.94}\textsuperscript{***} \\
\midrule
\multirow{3}{*}{\textbf{P300}} 
& ShallowFBCSPNet & 43 & 2 & 87.72 $\pm$ 0.33 & 86.20 $\pm$ 1.45 \\
& EEGConformer & 43 & 2 & 88.54 $\pm$ 0.54\textsuperscript{**} & 86.30 $\pm$ 1.73 \\
& TIDNet & 43 & 2 & 88.24 $\pm$ 0.31\textsuperscript{*} & 85.63 $\pm$ 0.58\textsuperscript{**} \\
& EEGTCNet & 43 & 2 & 88.69 $\pm$ 0.59\textsuperscript{***}  & 87.02 $\pm$ 1.62\textsuperscript{***} \\
& EEGNet & 43 & 2 & 88.79 $\pm$ 0.67\textsuperscript{***} & 87.24 $\pm$ 2.01\textsuperscript{***} \\
& \textbf{MultiDiffNet} & 43 & 2 & 85.35 $\pm$ 1.12 & 79.47 $\pm$ 0.54\textsuperscript{*} \\
& \textbf{MultiDiffNet + Mixup} & 43 & 2 & 85.61 $\pm$ 0.52 & 79.56 $\pm$ 4.43 \\
\midrule
\multirow{3}{*}{\textbf{MI}}
& ShallowFBCSPNet & 9 & 4 & 64.34 $\pm$ 3.61\textsuperscript{***} & 36.46 $\pm$ 6.60 \\
& EEGConformer & 9 & 4 & 59.57 $\pm$ 5.60\textsuperscript{**}  & 36.49 $\pm$ 7.72 \\
& TIDNet & 9 & 4 & 44.27 $\pm$ 2.60 & 34.42 $\pm$ 3.60 \\
& EEGTCNet & 9 & 4 & 58.85 $\pm$ 4.54 & 32.99 $\pm$ 6.94 \\
& EEGNet & 9 & 4 & 67.01 $\pm$ 5.38\textsuperscript{***} & 46.18 $\pm$ 7.20\textsuperscript{***} \\
& \textbf{MultiDiffNet} & 9 & 4 & 55.85 $\pm$ 2.80 & 39.24 $\pm$ 8.00\textsuperscript{***} \\
& \textbf{MultiDiffNet + Mixup} & 9 & 4 & 57.69 $\pm$ 3.27\textsuperscript{*} & 36.78 $\pm$ 5.23 \\
\midrule
\multirow{3}{*}{\textbf{Img. Speech}} 
& ShallowFBCSPNet & 14 & 11 & 13.78 $\pm$ 1.55\textsuperscript{**} & 10.48 $\pm$ 0.64 \\
& EEGConformer & 14 & 11 & 10.62 $\pm$ 0.82 & 9.21 $\pm$ 3.00 \\
& TIDNet & 14 & 11 & 9.10 $\pm$ 0.54 & 10.35 $\pm$ 0.18 \\
& EEGTCNet & 14 & 11 & 12.64 $\pm$ 1.58 & 10.10 $\pm$ 0.64 \\
& EEGNet & 14 & 11 & 11.26 $\pm$ 2.01\textsuperscript{*} & 10.61 $\pm$ 0.93\textsuperscript{*}\\
& \textbf{MultiDiffNet} & 14 & 11 & \textbf{15.55 $\pm$ 0.62}\textsuperscript{***} & \textbf{11.62 $\pm$ 1.29}\textsuperscript{***} \\
& \textbf{MultiDiffNet + Mixup} & 14 & 11 & \textbf{17.57 $\pm$ 1.16}\textsuperscript{***} & \textbf{12.12 $\pm$ 0.38}\textsuperscript{***} \\
\bottomrule
\end{tabular}
\vspace{\baselineskip}
\end{table*}

\subsection{Ablation studies} 

To understand what drives generalization in \textit{MultiDiffNet}, we ran extensive ablation experiments, over 100 controlled configs. All results are reported for both seen- and unseen-subject accuracy, with statistical evidence matrices and trend-level effect sizes in the Appendix.

\textbf{Decoder input.} Feeding only $z$ to the decoder often matches or exceeds more complex
fusion variants. For example, SSVEP unseen accuracy reaches 84.72\% with $z$ alone, further
boosted to 85.25\% with mixup, while more elaborate fusions ($z+x$, $x_{\text{hat}}+$ skips)
show no consistent gains. These findings validate our architectural decision to decode primarily
from $z$. For completeness, the best accuracies achieved in this ablation are 85.86/84.72 on
SSVEP, 85.88/81.41 on P300, 56.89/40.36 on Motor Imagery, and 18.58/12.88 on Imagined
Speech (seen/unseen).

\textbf{Classifier head.} A lightweight FC head on $z$ delivers state-of-the-art generalization
with minimal complexity. It rivals or outperforms EEGNet classifiers trained on $x$, especially
in low-SNR tasks. This supports our choice to use FC as the default classification head. For
completeness, the best accuracies achieved in this ablation are 85.08/84.72 on SSVEP, 85.35/84.12
on P300, 55.85/39.24 on Motor Imagery, and 17.95/11.61 on Imagined Speech (seen/unseen).

\textbf{Encoder and decoder.} Using raw $x$ as encoder input consistently outperforms $\hat{x}$, showing that denoising is useful for regularization. Interestingly, removing the decoder entirely sometimes improves generalization, suggesting that reconstruction may introduce noise if overemphasized. For completeness, the best accuracies in this ablation
are 90.95/85.58 on SSVEP, 85.71/80.93 on P300, 55.85/40.16 on Motor Imagery, and
19.22/13.76 on Imagined Speech (seen/unseen).

\textbf{Loss combinations.} Combining CE with mild MSE or contrastive losses improves stability, particularly when auxiliary weights are gently annealed. The best results use $\beta=0.05$, $\gamma=0.2$—balancing reconstruction as a regularizer without overpowering the classification objective. For completeness, the best accuracies in this ablation
are 86.40/85.58 on SSVEP, 85.69/80.18 on P300, 59.67/41.44 on Motor Imagery, and
19.60/13.51 on Imagined Speech (seen/unseen).

\textbf{Mixup strategies.} Mixup effects are task-specific. For SSVEP, \textit{Temporal Masked Mixup} outperforms all variants. Motor Imagery benefits from \textit{Weighted Average Mixup}, while P300 and Imagined Speech show limited sensitivity, highlighting that mixup is most impactful in high-SNR regimes. For completeness, the best accuracies in this ablation
are 87.84/85.26 on SSVEP, 85.78/79.56 on P300, 63.44/38.83 on Motor Imagery, and
19.47/12.12 on Imagined Speech (seen/unseen).

\section{Conclusions and future work}

We presented \textit{MultiDiffNet}, a diffusion-based neural decoder that learns a compact, multi-objective latent space for EEG decoding without synthetic augmentation. Through unified benchmarks and rigorous cross-subject evaluation, we showed that \textit{MultiDiffNet} achieves strong generalization across diverse BCI paradigms, particularly in challenging low-signal settings such as SSVEP and Imagined Speech. Our statistical analysis framework further addresses reproducibility challenges in low-trial EEG research. Future work will explore scaling \textit{MultiDiffNet} to larger and more diverse EEG datasets and extending the architecture to other neural modalities.

\section*{Acknowledgments}
We would like to thank Professor Bhiksha Raj of Carnegie Mellon University for his guidance and support throughout this project.

\clearpage
\bibliographystyle{abbrvnat}
\bibliography{references}

\appendix
\clearpage
\section*{Appendix}
\addcontentsline{toc}{section}{Appendix}

\section*{Appendix overview}

This appendix provides full experimental details, supporting results, and statistical analyses that complement the main paper:

\begin{itemize}
    \item \textbf{Section A: Implementation and experimental details} \\
    Covers architecture choices, training setups, loss function derivations, and augmentation strategies.
    
    \item \textbf{Section B: Generalization performance and analysis} \\
    Presents extended decoding results across all tasks with seen/unseen splits.

    \item \textbf{Section C: Complete ablation studies} \\
    Contains exhaustive ablations over decoder input types, classifier heads, architectural variants, mixup strategies, and loss function choices. Each is accompanied by multi-task results.

    \item \textbf{Section D: Complete statistical reporting results} \\
    Details our trend-level statistical framework, including effect size estimation, Bayesian comparison, and win-rate matrix construction. 
\end{itemize}

We include additional sections from the main text (e.g., Related Work) that were moved here due to space constraints.

\clearpage

\section{Implementation and experimental details}

\subsection{Loss functions} 

\subsubsection{Classification loss.} We use either cross-entropy (CE) or mean squared error (MSE). 

CE uses softmax over the classifier logits:

\[
\mathcal{L}_{\text{CE}} = -\sum_{c=1}^{N} y_c \log \hat{y}_c,\quad
\hat{y} = \text{Softmax}(\text{FC}(z))
\]

MSE treats classification as regression:

\[
\mathcal{L}_{\text{MSE}} = \| \hat{y} - y \|_2^2
\]

\subsubsection{Supervised contrastive loss.} A projection head maps the normalized embedding \( z \) to \( z_{\text{proj}} \). We then use the SupCon loss:

\[
\mathcal{L}_{\text{SupCon}} = \sum_{i \in I} \frac{-1}{|P(i)|} \sum_{p \in P(i)} \log
\frac{\exp(z_i \cdot z_p / \tau)}{\sum_{a \in A(i)} \exp(z_i \cdot z_a / \tau)}
\]

where \( \tau \) is the temperature, \( P(i) \) are positives, and \( A(i) \) includes all non-anchor samples.

\subsubsection{L1 reconstruction loss.} The decoder learns to reconstruct the DDPM-denoised signal:

\[
\mathcal{L}_{\text{L1}} = \| x_{\text{dec}} - \hat{x} \|_1
\]

We also compute an auxiliary loss \( \| \hat{x} - x \|_1 \) to guide DDPM training but exclude it from the final objective.

\clearpage  

\section{Generalization performance and analysis}

\begin{table*}[htbp]
\centering

\vspace{0.5em}

\resizebox{\textwidth}{!}{
\begin{tabular}{p{2cm} p{3.8cm} *{4}{p{2cm}} *{4}{p{2cm}}}
\toprule
\textbf{Task Paradigm}
& \textbf{Configuration}
& \multicolumn{4}{c}{\textbf{Seen}}
& \multicolumn{4}{c}{\textbf{Unseen}} \\
\cmidrule(lr){3-6} \cmidrule(lr){7-10}
& 
& Acc (\%) & F1 (\%) & Recall (\%)& AUC (\%)
& Acc (\%) & F1 (\%) & Recall (\%)& AUC (\%) \\
\midrule
\multirow{3}{*}{SSVEP}
& \texttt{EEGNet}
  & 89.16 $\pm$ 0.57 & 89.16 $\pm$ 0.55 & 89.16 $\pm$ 0.57 & 99.72 $\pm$ 0.04
  & 81.09 $\pm$ 9.16 & 81.58 $\pm$ 8.21 & 81.09 $\pm$ 9.16 & 98.75 $\pm$ 1.35 \\

& \texttt{MultiDiffNet}
  & 85.08 $\pm$ 1.53 & 84.95 $\pm$ 1.66 & 85.08 $\pm$ 1.53 & 99.37 $\pm$ 0.05
  & \textbf{84.72 $\pm$ 6.04} & \textbf{84.44 $\pm$ 6.21} & \textbf{84.72 $\pm$ 6.04} & \textbf{98.90 $\pm$ 0.76} \\

& \texttt{MultiDiffNet + Temp.\ Masked Mixup}
  & 86.79 $\pm$ 1.75 & 86.00 $\pm$ 0.00 & 86.00 $\pm$ 0.00 & 99.00 $\pm$ 0.00
  & 85.26 $\pm$ 6.94 & 83.00 $\pm$ 0.07 & 83.00 $\pm$ 0.07 & 99.00 $\pm$ 0.01 \\

\midrule
\multirow{2}{*}{P300}
& \texttt{EEGNet}
  & 88.79 $\pm$ 0.68 & 78.09 $\pm$ 1.73 & 76.03 $\pm$ 1.99 & 89.43 $\pm$ 1.06
  & 87.24 $\pm$ 2.01 & 73.72 $\pm$ 5.30 & 71.40 $\pm$ 5.69 & 85.97 $\pm$ 4.19 \\
& \texttt{MultiDiffNet}
  & 85.24 $\pm$ 1.02 & 63.06 $\pm$ 5.33 & 60.79 $\pm$ 4.04 & 77.54 $\pm$ 4.84
  & 76.26 $\pm$ 3.61 & 62.80 $\pm$ 3.19 & 64.95 $\pm$ 2.86 & 71.71 $\pm$ 4.68 \\

\midrule
\multirow{2}{*}{MI}
& \texttt{EEGNet}
  & 64.09 $\pm$ 5.77 & 63.62 $\pm$ 5.79 & 64.09 $\pm$ 5.77 & 87.09 $\pm$ 3.11
  & 43.89 $\pm$ 8.33 & 42.57 $\pm$ 8.35 & 43.89 $\pm$ 8.33 & 71.95 $\pm$ 7.51 \\
\bottomrule
\end{tabular}
}
\vspace{0.5em}
\caption{Final results across tasks and models. Accuracy (mean ± std) reported separately for both seen-subject (intra-subject) and unseen-subject (cross-subject) test splits. Tasks are ranked by task difficulty.}
\label{tab:eeg_results}
\vspace{1em}
\end{table*}

\clearpage  

\section{Complete ablation studies}

 \subsection{Decoder input ablations}

\begin{table*}[htbp]
\centering
\vspace{0.5em}

\resizebox{\textwidth}{!}{
\begin{tabular}{p{3.8cm} p{2cm} p{2cm} p{2cm} p{2cm} p{2cm} p{2cm} p{2cm} p{2cm}}
\toprule
\textbf{Decoder Input} 
& \multicolumn{2}{c}{\textbf{SSVEP}} 
& \multicolumn{2}{c}{\textbf{P300}} 
& \multicolumn{2}{c}{\textbf{Motor Imagery}} 
& \multicolumn{2}{c}{\textbf{Imagined Speech}} \\
\cmidrule(lr){2-3} \cmidrule(lr){4-5} \cmidrule(lr){6-7} \cmidrule(lr){8-9}
& Seen & Unseen & Seen & Unseen & Seen & Unseen & Seen & Unseen \\
\midrule
\texttt{x + x\_hat + skips} & \textbf{85.86 $\pm$ 0.38} & 83.12 $\pm$ 6.90 & 85.07 $\pm$ 1.03 & 79.15 $\pm$ 3.73 & 53.08 $\pm$ 4.68 & 38.74 $\pm$ 6.58 & 17.07 $\pm$ 1.86 & 12.37 $\pm$ 2.79 \\
\texttt{x + x\_hat} & 85.16 $\pm$ 0.95 & 84.40 $\pm$ 6.22 & 85.62 $\pm$ 0.43 & 78.59 $\pm$ 1.56 & 54.56 $\pm$ 1.83 & 40.13 $\pm$ 8.25 & \textbf{18.58 $\pm$ 2.13} & 10.48 $\pm$ 2.87 \\
\texttt{x + skips} & 85.63 $\pm$ 0.29 & 81.73 $\pm$ 7.29 & 85.75 $\pm$ 0.09 & 77.93 $\pm$ 1.43 & 54.66 $\pm$ 2.58 & \textbf{40.36 $\pm$ 7.30} & 17.32 $\pm$ 1.40 & 11.74 $\pm$ 2.42 \\
\texttt{x\_hat + skips} & 85.70 $\pm$ 0.14 & 83.55 $\pm$ 7.93 & 85.60 $\pm$ 0.57 & \textbf{81.41 $\pm$ 0.98} & 54.27 $\pm$ 2.88 & 40.31 $\pm$ 7.48 & 17.19 $\pm$ 0.47 & 12.50 $\pm$ 1.72 \\
\texttt{skips only} & 84.03 $\pm$ 1.33 & 82.59 $\pm$ 4.69 & 84.53 $\pm$ 0.82 & 75.52 $\pm$ 4.93 & 53.77 $\pm$ 2.80 & 37.33 $\pm$ 6.58 & 17.70 $\pm$ 2.30 & 11.11 $\pm$ 2.24 \\
\texttt{z only} & 85.08 $\pm$ 1.53 & \textbf{84.72 $\pm$ 6.04} & 85.35 $\pm$ 1.12 & 79.47 $\pm$ 0.54 & 55.85 $\pm$ 2.80 & 39.24 $\pm$ 7.95 & 15.55 $\pm$ 0.62 & 11.62 $\pm$ 1.29 \\
\bottomrule
\end{tabular}
}
\vspace{0.5em}
\caption{Ablation study of decoder input combinations for \textit{MultiDiffNet}. Accuracy (mean ± std) reported separately for seen- and unseen-subject test splits across four tasks.}
\label{tab:decoder_ablation_results}
\vspace{1em}

\end{table*}

\begin{table*}[htbp]
\centering
\vspace{0.5em}

\resizebox{\textwidth}{!}{
\begin{tabular}{p{3.8cm} *{8}{p{2cm}} }
\toprule
\textbf{Configuration}
& \multicolumn{4}{c}{\textbf{SSVEP (Seen)}}
& \multicolumn{4}{c}{\textbf{SSVEP (Unseen)}} \\
\cmidrule(lr){2-5} \cmidrule(lr){6-9}
& Acc (\%) & F1 (\%) & Recall (\%)& AUC (\%)
& Acc (\%) & F1 (\%) & Recall (\%)& AUC (\%) \\
\midrule
\texttt{x + x\_hat + skips}
  & \textbf{85.86 $\pm$ 0.38} & \textbf{85.84 $\pm$ 0.40} & \textbf{85.86 $\pm$ 0.38} & \textbf{99.50 $\pm$ 0.03}
  & 83.12 $\pm$ 6.91 & 82.85 $\pm$ 6.92 & 83.12 $\pm$ 6.91 & 98.79 $\pm$ 0.94 \\
\texttt{x + x\_hat}
  & 85.16 $\pm$ 0.95 & 85.14 $\pm$ 0.96 & 85.16 $\pm$ 0.95 & 99.40 $\pm$ 0.06
  & 84.40 $\pm$ 6.22 & 84.14 $\pm$ 6.36 & 84.40 $\pm$ 6.22 & \textbf{98.94 $\pm$ 0.80} \\
\texttt{x\_hat + skips}
  & 85.70 $\pm$ 0.15 & 85.67 $\pm$ 0.14 & 85.70 $\pm$ 0.15 & 99.45 $\pm$ 0.01
  & 83.55 $\pm$ 7.93 & 83.33 $\pm$ 8.13 & 83.55 $\pm$ 7.93 & 98.92 $\pm$ 0.84 \\
\texttt{x + skips}
  & 85.63 $\pm$ 0.29 & 85.61 $\pm$ 0.32 & 85.63 $\pm$ 0.29 & 99.40 $\pm$ 0.03
  & 81.73 $\pm$ 7.29 & 81.52 $\pm$ 7.44 & 81.73 $\pm$ 7.29 & 98.85 $\pm$ 0.95 \\
\texttt{skips}
  & 84.03 $\pm$ 1.33 & 83.86 $\pm$ 1.46 & 84.03 $\pm$ 1.33 & 99.41 $\pm$ 0.02
  & 82.59 $\pm$ 4.69 & 82.22 $\pm$ 4.66 & 82.59 $\pm$ 4.69 & 98.80 $\pm$ 0.91 \\
\texttt{z only}
  & 85.08 $\pm$ 1.53 & 84.95 $\pm$ 1.66 & 85.08 $\pm$ 1.53 & 99.37 $\pm$ 0.05
  & \textbf{84.72 $\pm$ 6.04} & \textbf{84.44 $\pm$ 6.21} & \textbf{84.72 $\pm$ 6.04} & 98.90 $\pm$ 0.76 \\
\bottomrule
\end{tabular}
}
\vspace{0.5em}
\caption{Ablation study of decoder input combinations for \textit{MultiDiffNet} in the ssvep task. Mean\,±\,std is reported separately for Accuracy, F1, Recall, and AUC on seen- and unseen-subject test splits across four tasks.}
\label{tab:ssvep_results}
\vspace{1em}
\end{table*}

\begin{table*}[htbp]
\centering

\vspace{0.5em}

\resizebox{\textwidth}{!}{
\begin{tabular}{p{3.8cm} p{2cm} p{2cm} p{2cm} p{2cm} p{2cm} p{2cm} p{2cm} p{2cm}}
\toprule
\textbf{Configuration}
& \multicolumn{4}{c}{\textbf{P300 (Seen)}}
& \multicolumn{4}{c}{\textbf{P300 (Unseen)}} \\
\cmidrule(lr){2-5} \cmidrule(lr){6-9}
& Acc (\%) & F1 (\%) & Recall (\%)& AUC (\%)
& Acc (\%) & F1 (\%) & Recall (\%)& AUC (\%) \\
\midrule
\texttt{x + x\_hat + skips}
  & 85.07 $\pm$ 1.03 & \textbf{67.21 $\pm$ 1.80} & \textbf{64.55 $\pm$ 1.39} & 77.87 $\pm$ 2.81
  & 79.15 $\pm$ 3.73 & 63.75 $\pm$ 4.99 & 64.99 $\pm$ 6.27 & 71.06 $\pm$ 7.65 \\
\texttt{x + x\_hat}
  & 85.62 $\pm$ 0.43 & 64.19 $\pm$ 2.47 & 61.38 $\pm$ 1.99 & 77.86 $\pm$ 2.52
  & 78.59 $\pm$ 1.56 & 63.56 $\pm$ 3.74 & 64.68 $\pm$ 4.86 & 71.54 $\pm$ 5.86 \\
\texttt{x\_hat + skips}
  & 85.60 $\pm$ 0.57 & 65.95 $\pm$ 1.10 & 63.00 $\pm$ 1.28 & \textbf{79.70 $\pm$ 0.77}
  & \textbf{81.41 $\pm$ 0.98} & 64.91 $\pm$ 4.68 & 64.99 $\pm$ 6.23 & \textbf{72.75 $\pm$ 7.16} \\
\texttt{x + skips}
  & 85.75 $\pm$ 0.09 & 65.96 $\pm$ 0.60 & 62.87 $\pm$ 0.57 & 79.66 $\pm$ 1.29
  & 77.93 $\pm$ 1.43 & 62.76 $\pm$ 1.61 & 64.23 $\pm$ 3.87 & 70.32 $\pm$ 2.97 \\
\texttt{skips}
  & 84.53 $\pm$ 0.82 & 65.14 $\pm$ 0.95 & 62.79 $\pm$ 1.48 & 77.33 $\pm$ 1.39
  & 75.52 $\pm$ 4.93 & 62.34 $\pm$ 4.91 & 64.58 $\pm$ 4.17 & 69.91 $\pm$ 6.11 \\
\texttt{z only}
  & 85.35 $\pm$ 1.12 & 64.39 $\pm$ 4.95 & 61.88 $\pm$ 4.10 & 77.14 $\pm$ 4.35
  & 79.47 $\pm$ 0.54 & 63.67 $\pm$ 1.58 & 63.95 $\pm$ 1.86 & 70.79 $\pm$ 3.27 \\
\bottomrule
\end{tabular}
}
\vspace{0.5em}

\caption{Ablation study of decoder input combinations for \textit{MultiDiffNet} in the P300 task. Mean\,±\,std is reported separately for Accuracy, F1, Recall, and AUC on seen- and unseen-subject test splits across four tasks.}
\label{tab:p300_results}
\vspace{1em}
\end{table*}

\begin{table*}[htbp]
\centering
\vspace{0.5em}

\resizebox{\textwidth}{!}{
\begin{tabular}{p{3.8cm} p{2cm} p{2cm} p{2cm} p{2cm} p{2cm} p{2cm} p{2cm} p{2cm}}
\toprule
\textbf{Configuration}
& \multicolumn{4}{c}{\textbf{MI (Seen)}}
& \multicolumn{4}{c}{\textbf{MI (Unseen)}} \\
\cmidrule(lr){2-5} \cmidrule(lr){6-9}
& Acc (\%) & F1 (\%) & Recall (\%)& AUC (\%)
& Acc (\%) & F1 (\%) & Recall (\%)& AUC (\%) \\
\midrule
\texttt{x + x\_hat + skips}
  & 53.08 $\pm$ 4.68 & 48.79 $\pm$ 5.69 & 53.08 $\pm$ 4.68 & 80.78 $\pm$ 4.91
  & 38.74 $\pm$ 6.58 & 37.74 $\pm$ 6.87 & 38.74 $\pm$ 6.58 & 65.41 $\pm$ 6.77 \\
\texttt{x + x\_hat}
  & 54.56 $\pm$ 1.83 & 52.83 $\pm$ 1.61 & 54.56 $\pm$ 1.83 & 83.00 $\pm$ 0.71
  & 40.13 $\pm$ 8.25 & 38.69 $\pm$ 8.47 & 40.13 $\pm$ 8.25 & 66.58 $\pm$ 6.26 \\
\texttt{x\_hat + skips}
  & 54.27 $\pm$ 2.88 & 52.66 $\pm$ 4.08 & 54.27 $\pm$ 2.88 & 81.45 $\pm$ 1.68
  & 40.31 $\pm$ 7.48 & 39.13 $\pm$ 7.91 & 40.31 $\pm$ 7.48 & 66.60 $\pm$ 6.85 \\
\texttt{x + skips}
  & 54.66 $\pm$ 2.58 & 52.09 $\pm$ 4.13 & 54.66 $\pm$ 2.58 & 82.00 $\pm$ 1.47
  & \textbf{40.36 $\pm$ 7.30} & 39.25 $\pm$ 7.58 &\textbf{ 40.36 $\pm$ 7.30} & 66.33 $\pm$ 6.98 \\
\texttt{skips}
  & 53.77 $\pm$ 2.80 & 51.12 $\pm$ 3.68 & 53.77 $\pm$ 2.80 & \textbf{83.80 $\pm$ 1.37}
  & 37.33 $\pm$ 6.58 & 36.21 $\pm$ 7.14 & 37.33 $\pm$ 6.58 & 63.85 $\pm$ 5.14 \\
\texttt{z only}
  & 55.85 $\pm$ 2.80 & 54.25 $\pm$ 3.15 & 55.85 $\pm$ 2.80 & 82.72 $\pm$ 0.86
  & 39.24 $\pm$ 7.95 & 38.07 $\pm$ 7.86 & 39.24 $\pm$ 7.95 & \textbf{66.70 $\pm$ 7.32} \\
\bottomrule
\end{tabular}
}
\vspace{0.5em}
\caption{Ablation study of decoder input combinations for \textit{MultiDiffNet} in the MI task. Mean\,±\,std is reported separately for Accuracy, F1, Recall, and AUC on seen- and unseen-subject test splits across four tasks.}
\label{tab:mi_results}
\vspace{1em}
\end{table*}

\begin{table*}[htbp]
\centering

\vspace{0.5em}

\resizebox{\textwidth}{!}{
\begin{tabular}{p{3.8cm} p{2cm} p{2cm} p{2cm} p{2cm} p{2cm} p{2cm} p{2cm} p{2cm}}
\toprule
\textbf{Configuration}
& \multicolumn{4}{c}{\textbf{Imag. Speech (Seen)}}
& \multicolumn{4}{c}{\textbf{Imag. Speech (Unseen)}} \\
\cmidrule(lr){2-5} \cmidrule(lr){6-9}
& Acc (\%) & F1 (\%) & Recall (\%)& AUC (\%)
& Acc (\%) & F1 (\%) & Recall (\%)& AUC (\%) \\
\midrule
\texttt{x + x\_hat + skips}
  & 17.07 $\pm$ 1.86 & 13.35 $\pm$ 2.60 & 17.05 $\pm$ 1.86 & 71.15 $\pm$ 3.16
  & 12.37 $\pm$ 2.79 &  9.76 $\pm$ 2.17 & 12.37 $\pm$ 2.79 & 57.98 $\pm$ 4.50 \\
\texttt{x + x\_hat}
  & \textbf{18.58 $\pm$ 2.13} & 13.39 $\pm$ 2.72 & \textbf{18.56 $\pm$ 2.14} & 72.58 $\pm$ 1.68
  & 10.48 $\pm$ 2.87 &  7.63 $\pm$ 0.81 & 10.48 $\pm$ 2.87 & 56.64 $\pm$ 6.04 \\
\texttt{x\_hat + skips}
  & 17.19 $\pm$ 0.47 & 12.90 $\pm$ 1.56 & 17.17 $\pm$ 0.47 & \textbf{73.16 $\pm$ 0.59}
  & 12.50 $\pm$ 1.72 &  9.02 $\pm$ 0.18 & 12.50 $\pm$ 1.72 & 56.86 $\pm$ 4.74 \\
\texttt{x + skips}
  & 17.32 $\pm$ 1.40 & 13.92 $\pm$ 1.93 & 17.30 $\pm$ 1.39 & 71.93 $\pm$ 2.02
  & 11.74 $\pm$ 2.42 &  8.57 $\pm$ 1.97 & 11.74 $\pm$ 2.42 & 56.47 $\pm$ 3.87 \\
\texttt{skips}
  & 17.70 $\pm$ 2.30 & 12.91 $\pm$ 2.52 & 17.68 $\pm$ 2.32 & 72.61 $\pm$ 1.56
  & 11.11 $\pm$ 2.24 &  9.59 $\pm$ 2.13 & 11.11 $\pm$ 2.24 & 56.24 $\pm$ 2.03 \\
\texttt{z only}
  & 15.55 $\pm$ 0.62 &  9.61 $\pm$ 1.23 & 15.53 $\pm$ 0.62 & 71.24 $\pm$ 1.66
  & 11.62 $\pm$ 1.29 &  7.84 $\pm$ 1.32 & 11.62 $\pm$ 1.29 & 57.15 $\pm$ 3.18 \\
\bottomrule
\end{tabular}
}
\vspace{0.5em}

\caption{Ablation study of decoder input combinations for \textit{MultiDiffNet} in the imagined speech task. Mean\,±\,std is reported separately for Accuracy, F1, Recall, and AUC on seen- and unseen-subject test splits across four tasks.}
\label{tab:kara_one_results}
\vspace{1em}
\end{table*}

\clearpage  

\subsection{Classifier head ablations}

\begin{table*}[htbp]
\centering

\vspace{0.5em}

\resizebox{\textwidth}{!}{
\begin{tabular}{p{3.8cm} p{2cm} p{2cm} p{2cm} p{2cm} p{2cm} p{2cm} p{2cm} p{2cm}}
\toprule
\textbf{Classifier Variant} 
& \multicolumn{2}{c}{\textbf{SSVEP}} 
& \multicolumn{2}{c}{\textbf{P300}} 
& \multicolumn{2}{c}{\textbf{Motor Imagery}} 
& \multicolumn{2}{c}{\textbf{Imagined Speech}} \\
\cmidrule(lr){2-3} \cmidrule(lr){4-5} \cmidrule(lr){6-7} \cmidrule(lr){8-9}
& Seen & Unseen & Seen & Unseen & Seen & Unseen & Seen & Unseen \\
\midrule
\texttt{x $\rightarrow$ EEGNet} & 
\textbf{89.16 $\pm$ 0.57} & \textbf{81.09 $\pm$ 9.16} &
\textbf{88.79 $\pm$ 0.68} & \textbf{87.24 $\pm$ 2.01} &
\textbf{67.01 $\pm$ 5.38} & \textbf{46.18 $\pm$ 7.21} & 
11.26 $\pm$ 2.01 & 10.61 $\pm$ 0.93 \\

\texttt{x\_hat $\rightarrow$ EEGNet} & 
10.61 $\pm$ 4.96 & 9.19 $\pm$ 7.57 &
79.18 $\pm$ 1.32 & 79.08 $\pm$ 1.49 &
26.63 $\pm$ 0.21 & 25.84 $\pm$ 0.43 & 
9.86 $\pm$ 3.27 & 7.70 $\pm$ 1.09 \\

\texttt{deco\_out $\rightarrow$ EEGNet} & 
80.22 $\pm$ 0.99 & 70.62 $\pm$ 6.38 &
84.29 $\pm$ 0.77 & 83.52 $\pm$ 1.07 &
52.63 $\pm$ 3.56 & 35.16 $\pm$ 5.48 & 
\textbf{16.94 $\pm$ 1.28} & \textbf{10.73 $\pm$ 1.59} \\

\texttt{x $\rightarrow$ FC} & 
10.72 $\pm$ 2.01 & 11.43 $\pm$ 5.31 &
83.30 $\pm$ 0.05 & 83.33 $\pm$ 0.00 &
47.12 $\pm$ 0.92 & 29.48 $\pm$ 2.58 & 
16.94 $\pm$ 1.97 & 8.71 $\pm$ 0.93 \\

\texttt{x\_hat $\rightarrow$ FC} & 
3.73 $\pm$ 0.25 & 4.06 $\pm$ 0.54 &
83.33 $\pm$ 0.00 & 83.33 $\pm$ 0.00 &
25.30 $\pm$ 0.42 & 24.77 $\pm$ 0.33 & 
10.62 $\pm$ 0.02 & 8.96 $\pm$ 0.94 \\

\texttt{deco\_out $\rightarrow$ FC} & 
77.62 $\pm$ 1.44 & 69.55 $\pm$ 6.77 &
85.10 $\pm$ 0.35 & 84.11 $\pm$ 1.01 &
50.64 $\pm$ 2.08 & 33.74 $\pm$ 4.12 & 
\textbf{17.95 $\pm$ 1.39} & 8.71 $\pm$ 1.07 \\

\texttt{z $\rightarrow$ FC} & 
\textbf{85.08 $\pm$ 1.53} & \textbf{84.72 $\pm$ 6.04} &
\textbf{85.35 $\pm$ 0.35} & \textbf{84.12 $\pm$ 1.01} &
\textbf{55.85 $\pm$ 2.80} & \textbf{39.24 $\pm$ 7.95} & 
15.55 $\pm$ 0.62 & \textbf{11.61 $\pm$ 1.29} \\
\bottomrule
\end{tabular}
}
\vspace{0.5em}

\caption{Ablation study of classifier variants.Accuracy (mean ± std) reported separately for seen- and unseen-subject test splits across four tasks.}
\label{tab:classifier_ablation_results}
\vspace{1em}
\end{table*}

\begin{table*}[htbp]
\centering
\vspace{0.5em}

\resizebox{\textwidth}{!}{
\begin{tabular}{p{3.5cm}*{8}{p{2cm}}}
\toprule
\textbf{Configuration}
& \multicolumn{4}{c}{\textbf{SSVEP (Seen)}}
& \multicolumn{4}{c}{\textbf{SSVEP (Unseen)}} \\
\cmidrule(lr){2-5} \cmidrule(lr){6-9}
& Acc (\%) & F1 (\%) & Recall (\%)& AUC (\%)
& Acc (\%) & F1 (\%) & Recall (\%)& AUC (\%)  \\
\midrule
\texttt{eegn\_clsf\_\_x}
  & \textbf{89.16 $\pm$ 0.57} & \textbf{89.16 $\pm$ 0.55} & \textbf{89.16 $\pm$ 0.57} & \textbf{99.72 $\pm$ 0.04}
  & 81.09 $\pm$ 9.16 & 81.58 $\pm$ 8.21 & 81.09 $\pm$ 9.16 & 98.75 $\pm$ 1.35 \\
\texttt{eegn\_clsf\_\_x\_hat}
  & 10.61 $\pm$ 4.96 & 8.73 $\pm$ 4.32 & 10.61 $\pm$ 4.96 & 59.86 $\pm$ 5.14
  & 9.19 $\pm$ 7.57 & 8.15 $\pm$ 6.91 & 9.19 $\pm$ 7.57 & 59.39 $\pm$ 8.07 \\
\texttt{eegn\_clsf\_\_deco\_out}
  & 80.23 $\pm$ 0.99 & 80.19 $\pm$ 1.01 & 80.23 $\pm$ 0.99 & 99.35 $\pm$ 0.03
  & 70.62 $\pm$ 6.38 & 69.69 $\pm$ 6.48 & 70.62 $\pm$ 6.38 & 98.21 $\pm$ 1.26 \\
\texttt{fc\_clsf\_\_x}
  & 10.72 $\pm$ 2.01 & 8.91 $\pm$ 1.93 & 10.72 $\pm$ 2.01 & 73.43 $\pm$ 3.35
  & 11.43 $\pm$ 5.31 & 9.63 $\pm$ 5.44 & 11.43 $\pm$ 5.31 & 70.57 $\pm$ 5.03 \\
\texttt{fc\_clsf\_\_x\_hat}
  & 3.73 $\pm$ 0.25 & 0.68 $\pm$ 0.30 & 3.73 $\pm$ 0.25 & 49.74 $\pm$ 0.48
  & 4.06 $\pm$ 0.54 & 1.17 $\pm$ 0.32 & 4.06 $\pm$ 0.54 & 49.43 $\pm$ 1.77 \\
\texttt{fc\_clsf\_\_deco\_out}
  & 77.62 $\pm$ 1.44 & 77.49 $\pm$ 1.59 & 77.62 $\pm$ 1.44 & 99.25 $\pm$ 0.06
  & 69.55 $\pm$ 6.77 & 69.56 $\pm$ 6.81 & 69.55 $\pm$ 6.77 & 98.03 $\pm$ 1.17 \\
\texttt{fc\_clsf\_\_z}
  & 85.08 $\pm$ 1.53 & 84.95 $\pm$ 1.66 & 85.08 $\pm$ 1.53 & 99.37 $\pm$ 0.05
  & \textbf{84.72 $\pm$ 6.04} & \textbf{84.44 $\pm$ 6.21} & \textbf{84.72 $\pm$ 6.04} & \textbf{98.90 $\pm$ 0.76} \\
\bottomrule
\end{tabular}
}
\vspace{0.5em}
\caption{
Ablation study of \texttt{classifier\_variant} and \texttt{classifier\_input} combinations on the SSVEP task. Mean\,±\,std is reported separately for Accuracy, F1, Recall, and AUC on both seen- and unseen-subject test splits.  Results are averaged across four tasks (“eegn” = EEGNet, “clsf” = classifier, “deco” = decoder).}
\label{tab:ssvep_eegnet_fc_results}
\vspace{1em}
\end{table*}

\begin{table*}[htbp]
\centering

\vspace{0.5em}

\resizebox{\textwidth}{!}{
\begin{tabular}{p{3.8cm} *{8}{p{2cm}} }
\toprule
\textbf{Configuration}
& \multicolumn{4}{c}{\textbf{P300 (Seen)}}
& \multicolumn{4}{c}{\textbf{P300 (Unseen)}} \\
\cmidrule(lr){2-5} \cmidrule(lr){6-9}
& Acc (\%) & F1 (\%) & Recall (\%)& AUC (\%)
& Acc (\%) & F1 (\%) & Recall (\%)& AUC (\%)  \\
\midrule
\texttt{eegn\_clsf\_\_x}
  & \textbf{88.79 $\pm$ 0.68} & \textbf{78.09 $\pm$ 1.73} & \textbf{76.03 $\pm$ 1.99} & \textbf{89.43 $\pm$ 1.06}
  & \textbf{87.24 $\pm$ 2.01} & \textbf{73.72 $\pm$ 5.30} & \textbf{71.40 $\pm$ 5.69} & \textbf{85.97 $\pm$ 4.19} \\
\texttt{eegn\_clsf\_\_x\_hat}
  & 79.18 $\pm$ 1.32 & 48.09 $\pm$ 2.56 & 49.70 $\pm$ 1.82 & 46.94 $\pm$ 8.58
  & 79.08 $\pm$ 1.49 & 47.72 $\pm$ 1.89 & 49.41 $\pm$ 1.50 & 45.95 $\pm$ 6.80 \\
\texttt{eegn\_clsf\_\_deco\_out}
  & 84.29 $\pm$ 0.77 & 56.01 $\pm$ 7.51 & 56.08 $\pm$ 4.34 & 66.90 $\pm$ 12.00
  & 83.52 $\pm$ 1.07 & 54.02 $\pm$ 8.11 & 54.95 $\pm$ 5.26 & 64.43 $\pm$ 9.42 \\
\texttt{fc\_clsf\_\_x}
  & 83.30 $\pm$ 0.05 & 45.50 $\pm$ 0.06 & 50.00 $\pm$ 0.00 & 50.08 $\pm$ 0.23
  & 83.33 $\pm$ 0.00 & 45.45 $\pm$ 0.00 & 50.00 $\pm$ 0.00 & 50.32 $\pm$ 0.38 \\
\texttt{fc\_clsf\_\_x\_hat}
  & 83.33 $\pm$ 0.00 & 45.45 $\pm$ 0.00 & 50.00 $\pm$ 0.00 & 49.85 $\pm$ 0.50
  & 83.33 $\pm$ 0.00 & 45.45 $\pm$ 0.00 & 50.00 $\pm$ 0.00 & 50.31 $\pm$ 0.30 \\
\texttt{fc\_clsf\_\_deco\_out}
  & 85.10 $\pm$ 0.35 & 60.43 $\pm$ 2.05 & 58.48 $\pm$ 1.48 & 74.88 $\pm$ 2.73
  & 84.12 $\pm$ 1.01 & 56.01 $\pm$ 5.84 & 55.80 $\pm$ 3.95 & 70.24 $\pm$ 6.64 \\
\texttt{fc\_clsf\_\_z}
  & 85.35 $\pm$ 1.12 & 64.39 $\pm$ 4.95 & 61.88 $\pm$ 4.10 & 77.14 $\pm$ 4.35
  & 79.47 $\pm$ 0.54 & 63.67 $\pm$ 1.58 & 63.95 $\pm$ 1.86 & 70.79 $\pm$ 3.27 \\
\bottomrule
\end{tabular}
}
\vspace{0.5em}
\caption{Ablation study of \texttt{classifier\_variant}, and \texttt{classifier\_input} combinations in the P300 task. Mean\,±\,std is reported separately for Accuracy, F1, Recall, and AUC on seen- and unseen-subject test splits across four tasks (“eegn” = EEGNet, “clsf” = classifier, “deco” = decoder).}
\label{tab:p300_eegnet_fc_results}
\vspace{1em}
\end{table*}

\begin{table*}[htbp]
\centering
\vspace{0.5em}

\resizebox{\textwidth}{!}{
\begin{tabular}{p{3.8cm} *{8}{p{2cm}} }
\toprule
\textbf{Configuration}
& \multicolumn{4}{c}{\textbf{MI (Seen)}}
& \multicolumn{4}{c}{\textbf{MI (Unseen)}} \\
\cmidrule(lr){2-5} \cmidrule(lr){6-9}
& Acc (\%) & F1 (\%) & Recall (\%)& AUC (\%)
& Acc (\%) & F1 (\%) & Recall (\%)& AUC (\%) \\
\midrule
\texttt{eegn\_clsf\_\_x}
  & \textbf{67.01 $\pm$ 5.38} & \textbf{67.01 $\pm$ 5.39} & \textbf{67.01 $\pm$ 5.38} & \textbf{88.66 $\pm$ 3.28}
  & \textbf{46.18 $\pm$ 7.21} & \textbf{45.69 $\pm$ 7.30} & \textbf{46.18 $\pm$ 7.21} & \textbf{72.75 $\pm$ 5.62} \\
\texttt{eegn\_clsf\_\_x\_hat}
  & 26.64 $\pm$ 0.21 & 20.09 $\pm$ 3.08 & 26.64 $\pm$ 0.21 & 53.43 $\pm$ 1.52
  & 25.84 $\pm$ 0.43 & 18.80 $\pm$ 4.43 & 25.84 $\pm$ 0.43 & 50.57 $\pm$ 0.75 \\
\texttt{eegn\_clsf\_\_deco\_out}
  & 52.63 $\pm$ 3.56 & 51.30 $\pm$ 3.43 & 52.63 $\pm$ 3.56 & 79.26 $\pm$ 2.99
  & 35.16 $\pm$ 5.48 & 31.97 $\pm$ 4.78 & 35.16 $\pm$ 5.48 & 60.71 $\pm$ 5.32 \\
\texttt{fc\_clsf\_\_x}
  & 47.12 $\pm$ 0.92 & 46.83 $\pm$ 1.08 & 47.12 $\pm$ 0.92 & 73.82 $\pm$ 0.72
  & 29.48 $\pm$ 2.59 & 26.58 $\pm$ 2.96 & 29.48 $\pm$ 2.59 & 57.16 $\pm$ 3.26 \\
\texttt{fc\_clsf\_\_x\_hat}
  & 25.30 $\pm$ 0.42 & 12.15 $\pm$ 3.04 & 25.30 $\pm$ 0.42 & 50.92 $\pm$ 1.51
  & 24.77 $\pm$ 0.33 & 11.37 $\pm$ 1.93 & 24.77 $\pm$ 0.33 & 49.21 $\pm$ 1.85 \\
\texttt{fc\_clsf\_\_deco\_out}
  & 50.64 $\pm$ 2.08 & 49.99 $\pm$ 2.38 & 50.64 $\pm$ 2.08 & 76.84 $\pm$ 2.48
  & 33.74 $\pm$ 4.12 & 30.64 $\pm$ 5.43 & 33.74 $\pm$ 4.12 & 60.94 $\pm$ 4.40 \\
\texttt{fc\_clsf\_\_z}
  & 55.85 $\pm$ 2.80 & 54.25 $\pm$ 3.15 & 55.85 $\pm$ 2.80 & 82.72 $\pm$ 0.86
  & 39.24 $\pm$ 7.95 & 38.07 $\pm$ 7.86 & 39.24 $\pm$ 7.95 & 66.70 $\pm$ 7.32 \\
\bottomrule
\end{tabular}
}
\vspace{0.5em}

\caption{Ablation study of \texttt{classifier\_variant}, and \texttt{classifier\_input} combinations in the MI task. Mean\,±\,std is reported separately for Accuracy, F1, Recall, and AUC on seen- and unseen-subject test splits across four tasks (“eegn” = EEGNet, “clsf” = classifier, “deco” = decoder).}
\label{tab:mi_eegnet_fc_results}
\vspace{1em}
\end{table*}

\begin{table*}[htbp]
\centering

\vspace{0.5em}

\resizebox{\textwidth}{!}{
\begin{tabular}{p{3.8cm} *{8}{p{2cm}} }
\toprule
\textbf{Configuration}
& \multicolumn{4}{c}{\textbf{Imag. Speech (Seen)}}
& \multicolumn{4}{c}{\textbf{Imag. Speech (Unseen)}} \\
\cmidrule(lr){2-5} \cmidrule(lr){6-9}
&  Acc (\%) & F1 (\%) & Recall (\%)& AUC (\%)
& Acc (\%) & F1 (\%) & Recall (\%)& AUC (\%) \\
\midrule
\texttt{eegn\_clsf\_\_x}
  & 11.26 $\pm$ 2.01 & 11.19 $\pm$ 1.98 & 11.26 $\pm$ 2.02 & 52.02 $\pm$ 0.42
  & 10.61 $\pm$ 0.93 & \textbf{10.17 $\pm$ 0.75} & 10.61 $\pm$ 0.93 & 50.33 $\pm$ 1.25 \\
\texttt{eegn\_clsf\_\_x\_hat}
  &  9.86 $\pm$ 3.27 &  8.82 $\pm$ 2.53 &  9.85 $\pm$ 3.27 & 50.06 $\pm$ 1.59
  &  7.70 $\pm$ 1.09 &  6.15 $\pm$ 0.85 &  7.70 $\pm$ 1.09 & 49.10 $\pm$ 3.36 \\
\texttt{eegn\_clsf\_\_deco\_out}
  & \textbf{16.94 $\pm$ 1.28} & \textbf{14.73 $\pm$ 0.81} & \textbf{16.92 $\pm$ 1.26} & \textbf{73.33 $\pm$ 0.13}
  & 10.73 $\pm$ 1.59 &  6.33 $\pm$ 1.77 & 10.73 $\pm$ 1.59 & 51.50 $\pm$ 1.99 \\
\texttt{fc\_clsf\_\_x}
  & 16.94 $\pm$ 1.97 & 16.68 $\pm$ 1.78 & 16.93 $\pm$ 1.98 & 68.36 $\pm$ 0.93
  &  8.71 $\pm$ 0.93 &  6.31 $\pm$ 0.76 &  8.71 $\pm$ 0.93 & 50.99 $\pm$ 0.69 \\
\texttt{fc\_clsf\_\_x\_hat}
  & 10.62 $\pm$ 0.02 &  4.48 $\pm$ 1.18 & 10.61 $\pm$ 0.00 & 51.62 $\pm$ 1.79
  &  8.96 $\pm$ 0.94 &  3.48 $\pm$ 0.57 &  8.96 $\pm$ 0.94 & 51.20 $\pm$ 0.41 \\
\texttt{fc\_clsf\_\_deco\_out}
  & 17.95 $\pm$ 1.39 & 17.02 $\pm$ 1.18 & 17.95 $\pm$ 1.39 & 74.32 $\pm$ 0.97
  &  8.71 $\pm$ 1.07 &  4.46 $\pm$ 0.35 &  8.71 $\pm$ 1.07 & 49.30 $\pm$ 1.62 \\
\texttt{fc\_clsf\_\_z}
  & 15.55 $\pm$ 0.62 &  9.61 $\pm$ 1.23 & 15.53 $\pm$ 0.62 & 71.24 $\pm$ 1.66
  & \textbf{11.62 $\pm$ 1.29} &  7.84 $\pm$ 1.32 & \textbf{11.62 $\pm$ 1.29} & \textbf{57.15 $\pm$ 3.18} \\
\bottomrule
\end{tabular}
}
\vspace{0.5em}

\caption{Ablation study of \texttt{classifier\_variant}, and \texttt{classifier\_input} combinations in the imagened speech task. Mean\,±\,std is reported separately for Accuracy, F1, Recall, and AUC on seen- and unseen-subject test splits across four tasks.}
\label{tab:kara_one_eegnet_fc_results}
\vspace{1em}
\end{table*}

\clearpage  

\subsection{Encoder and decoder ablations}

\begin{table*}[htbp]
\centering

\vspace{0.5em}

\resizebox{\textwidth}{!}{
\begin{tabular}{p{3.2cm} p{2cm} p{2cm} p{2cm} p{2cm} p{2cm} p{2cm} p{2cm} p{2cm}}
\toprule
\textbf{Configuration} 
& \multicolumn{2}{c}{\textbf{SSVEP}} 
& \multicolumn{2}{c}{\textbf{P300}} 
& \multicolumn{2}{c}{\textbf{Motor Imagery}} 
& \multicolumn{2}{c}{\textbf{Imagained Speech}} \\
\cmidrule(lr){2-3} \cmidrule(lr){4-5} \cmidrule(lr){6-7} \cmidrule(lr){8-9}
& Seen & Unseen & Seen & Unseen & Seen & Unseen & Seen & Unseen \\
\midrule
\texttt{u\_dp-x-u\_deco} & 
85.08 $\pm$ 1.53 & 84.72 $\pm$ 6.04 &
85.35 $\pm$ 1.12 & 79.47 $\pm$ 0.54 & 
\textbf{55.85 $\pm$ 2.80} & 39.24 $\pm$ 7.95 & 
15.55 $\pm$ 0.62 & 11.62 $\pm$ 1.29 \\
\texttt{u\_dp-x-n\_deco} & 
85.51 $\pm$ 0.63 & 83.55 $\pm$ 5.60 &
\textbf{85.71 $\pm$ 0.38} & \textbf{80.93 $\pm$ 1.52} & 
53.22 $\pm$ 4.58 & \textbf{40.16 $\pm$ 6.17} & 
18.71 $\pm$ 1.43 & 10.98 $\pm$ 1.61 \\
\texttt{u\_dp-x\_h-u\_deco} & 
6.53 $\pm$ 1.07 & 7.58 $\pm$ 2.35 &
80.02 $\pm$ 4.40 & 69.43 $\pm$ 4.04 & 
25.20 $\pm$ 0.86 & 26.04 $\pm$ 0.07 & 
8.85 $\pm$ 0.37 & 9.47 $\pm$ 1.35 \\
\texttt{u\_dp-x\_h-n\_deco} & 
6.72 $\pm$ 1.35 & 7.05 $\pm$ 1.63 &
83.11 $\pm$ 0.18 & 66.39 $\pm$ 6.24 & 
28.12 $\pm$ 0.99 & 26.04 $\pm$ 0.28 & 
8.97 $\pm$ 0.16 & 8.33 $\pm$ 2.75 \\
\texttt{n\_dp-x-u\_deco} & 
90.29 $\pm$ 0.29 & 84.94 $\pm$ 8.16 &
85.36 $\pm$ 1.28 & 78.84 $\pm$ 5.44 & 
55.06 $\pm$ 5.31 & 36.49 $\pm$ 2.70 & 
\textbf{19.22 $\pm$ 2.13} & \textbf{13.76 $\pm$ 1.25} \\
\texttt{n\_dp-x-n\_deco} & 
\textbf{90.95 $\pm$ 0.83} & \textbf{85.58 $\pm$ 6.15} &
85.49 $\pm$ 0.66 & 79.46 $\pm$ 0.72 & 
53.22 $\pm$ 4.44 & 38.54 $\pm$ 3.87 & 
19.22 $\pm$ 0.20 & 11.36 $\pm$ 1.93 \\
\bottomrule
\end{tabular}
}
\vspace{0.5em}
\caption{Ablation study of \texttt{ddpm\_variant}, \texttt{encoder\_input}, and \texttt{decoder\_variant} combinations. Accuracy (mean ± std) reported separately for seen/unseen-subject test splits across tasks (``dp'' = ddpm\_variant, ``x\_h'' = x\_hat, ``deco'' = decoder\_variant,``u'' = use,``n'' = not use,).}
\label{tab:ddpm_ablation_results}
\vspace{1em}
\end{table*}

\begin{table*}[htbp]
\centering

\vspace{0.5em}

\resizebox{\textwidth}{!}{
\begin{tabular}{p{3.5cm} *{8}{p{2cm}} }
\toprule
\textbf{Configuration}
& \multicolumn{4}{c}{\textbf{SSVEP (Seen)}}
& \multicolumn{4}{c}{\textbf{SSVEP (Unseen)}} \\
\cmidrule(lr){2-5} \cmidrule(lr){6-9}
& Acc (\%) & F1 (\%) & Recall (\%)& AUC (\%)
& Acc (\%) & F1 (\%) & Recall (\%)& AUC (\%) \\
\midrule
\texttt{u\_dp\_\_x\_\_u\_deco}
  & 85.08 $\pm$ 1.53 & 84.95 $\pm$ 1.66 & 85.08 $\pm$ 1.53 & 99.37 $\pm$ 0.05
  & 84.72 $\pm$ 6.04 & 84.44 $\pm$ 6.21 & 84.72 $\pm$ 6.04 & 98.90 $\pm$ 0.76 \\
\texttt{u\_dp\_\_x\_\_n\_deco}
  & 85.51 $\pm$ 0.63 & 85.48 $\pm$ 0.69 & 85.51 $\pm$ 0.63 & 99.44 $\pm$ 0.02
  & 83.55 $\pm$ 5.60 & 83.25 $\pm$ 6.02 & 83.55 $\pm$ 5.60 & 98.92 $\pm$ 0.99 \\
\texttt{u\_dp\_\_x\_h\_\_u\_deco}
  & 6.53 $\pm$ 1.07 & 4.89 $\pm$ 0.91 & 6.53 $\pm$ 1.07 & 55.27 $\pm$ 1.84
  & 7.59 $\pm$ 2.35 & 7.11 $\pm$ 2.35 & 7.59 $\pm$ 2.35 & 56.22 $\pm$ 2.99 \\
\texttt{u\_dp\_\_x\_h\_\_n\_deco}
  & 6.72 $\pm$ 1.35 & 4.38 $\pm$ 1.43 & 6.72 $\pm$ 1.35 & 56.19 $\pm$ 3.11
  & 7.05 $\pm$ 1.63 & 6.48 $\pm$ 1.48 & 7.05 $\pm$ 1.63 & 54.26 $\pm$ 2.37 \\
\texttt{n\_dp\_\_x\_\_u\_deco}
  & 90.29 $\pm$ 0.29 & 90.36 $\pm$ 0.28 & 90.29 $\pm$ 0.29 & \textbf{99.68 $\pm$ 0.04}
  & 84.94 $\pm$ 8.16 & 84.53 $\pm$ 8.58 & 84.94 $\pm$ 8.16 & \textbf{99.11 $\pm$ 0.88} \\
\texttt{n\_dp\_\_x\_\_n\_deco}
  & \textbf{90.95 $\pm$ 0.83} & \textbf{91.02 $\pm$ 0.84} & \textbf{90.95 $\pm$ 0.83} & 99.68 $\pm$ 0.01
  & \textbf{85.58 $\pm$ 6.15} & \textbf{85.32 $\pm$ 6.37} & \textbf{85.58 $\pm$ 6.15} & 99.07 $\pm$ 0.83 \\
\bottomrule
\end{tabular}
}
\vspace{0.5em}

\caption{Ablation study of \texttt{ddpm\_variant}, \texttt{encoder\_input} and \texttt{decoder\_variant} combinations in the SSVEP task. Mean\,±\,std is reported separately for Accuracy, F1, Recall, and AUC on seen- and unseen-subject test splits across four tasks (``dp'' = ddpm\_variant, ``x\_h'' = x\_hat, ``deco'' = decoder\_variant,``u'' = use,``n'' = not use).}
\label{tab:ssvep_decoder_ablation_results}

\vspace{1em}
\end{table*}

\begin{table*}[htbp]
\centering
\vspace{0.5em}

\resizebox{\textwidth}{!}{
\begin{tabular}{p{3.8cm} *{8}{p{2cm}} }
\toprule
\textbf{Configuration}
& \multicolumn{4}{c}{\textbf{P300 (Seen)}}
& \multicolumn{4}{c}{\textbf{P300 (Unseen)}} \\
\cmidrule(lr){2-5} \cmidrule(lr){6-9}
& Acc (\%) & F1 (\%) & Recall (\%)& AUC (\%)
& Acc (\%) & F1 (\%) & Recall (\%)& AUC (\%) \\
\midrule
\texttt{u\_dp\_\_x\_\_u\_deco}
  & 85.35 $\pm$ 1.12 & 64.39 $\pm$ 4.95 & 61.88 $\pm$ 4.10 & 77.14 $\pm$ 4.35
  & 79.47 $\pm$ 0.54 & 63.67 $\pm$ 1.58 & 63.95 $\pm$ 1.86 & 70.79 $\pm$ 3.27 \\
\texttt{u\_dp\_\_x\_\_n\_deco}
  & \textbf{85.71 $\pm$ 0.38} & 67.06 $\pm$ 2.38 & 64.12 $\pm$ 2.35 & \textbf{79.67 $\pm$ 1.76}
  & \textbf{80.94 $\pm$ 1.53} & 64.74 $\pm$ 5.62 & 65.57 $\pm$ 6.99 & 72.64 $\pm$ 7.24 \\
\texttt{u\_dp\_\_x\_h\_\_u\_deco}
  & 80.02 $\pm$ 4.40 & 48.44 $\pm$ 3.74 & 51.30 $\pm$ 1.76 & 52.24 $\pm$ 4.16
  & 69.43 $\pm$ 4.04 & 50.33 $\pm$ 1.63 & 50.64 $\pm$ 1.36 & 50.71 $\pm$ 2.41 \\
\texttt{u\_dp\_\_x\_h\_\_n\_deco}
  & 83.11 $\pm$ 0.18 & 45.76 $\pm$ 0.42 & 50.02 $\pm$ 0.11 & 48.80 $\pm$ 1.49
  & 66.39 $\pm$ 6.24 & 47.84 $\pm$ 2.84 & 48.03 $\pm$ 2.67 & 46.97 $\pm$ 3.74 \\
\texttt{n\_dp\_\_x\_\_u\_deco}
  & 85.36 $\pm$ 1.28 & \textbf{68.45 $\pm$ 3.02} & \textbf{66.09 $\pm$ 3.16} & 79.59 $\pm$ 2.08
  & 78.84 $\pm$ 5.44 & \textbf{65.21 $\pm$ 5.06} & \textbf{66.28 $\pm$ 3.31} & \textbf{73.36 $\pm$ 6.13} \\
\texttt{n\_dp\_\_x\_\_n\_deco}
  & 85.50 $\pm$ 0.66 & 62.95 $\pm$ 4.25 & 60.57 $\pm$ 3.45 & 76.39 $\pm$ 3.83
  & 79.46 $\pm$ 0.72 & 64.63 $\pm$ 2.54 & 65.53 $\pm$ 3.49 & 72.86 $\pm$ 3.49 \\
\bottomrule
\end{tabular}
}
\vspace{0.5em}
\caption{Ablation study of \texttt{ddpm\_variant}, \texttt{encoder\_input} and \texttt{decoder\_variant} combinations in the P300 task. Mean\,±\,std is reported separately for Accuracy, F1, Recall, and AUC on seen- and unseen-subject test splits across four tasks (``dp'' = ddpm\_variant, ``x\_h'' = x\_hat, ``deco'' = decoder\_variant,``u'' = use,``n'' = not use).}
\label{tab:p300_decoder_ablation_results}
\vspace{1em}
\end{table*}

\begin{table*}[htbp]
\centering
\vspace{0.5em}

\resizebox{\textwidth}{!}{
\begin{tabular}{p{3.8cm} *{8}{p{2cm}} }
\toprule
\textbf{Configuration}
& \multicolumn{4}{c}{\textbf{MI (Seen)}}
& \multicolumn{4}{c}{\textbf{MI (Unseen)}} \\
\cmidrule(lr){2-5} \cmidrule(lr){6-9}
& Acc (\%) & F1 (\%) & Recall (\%)& AUC (\%)
& Acc (\%) & F1 (\%) & Recall (\%)& AUC (\%)\\
\midrule
\texttt{u\_dp\_\_x\_\_u\_deco}
  & \textbf{55.85 $\pm$ 2.80} & \textbf{54.25 $\pm$ 3.15} & \textbf{55.85 $\pm$ 2.80} & \textbf{82.72 $\pm$ 0.86}
  & 39.24 $\pm$ 7.95 & \textbf{38.07 $\pm$ 7.86} & 39.24 $\pm$ 7.95 & \textbf{66.70 $\pm$ 7.32} \\
\texttt{u\_dp\_\_x\_\_n\_deco}
  & 53.22 $\pm$ 4.58 & 51.30 $\pm$ 5.58 & 53.22 $\pm$ 4.58 & 81.97 $\pm$ 1.51
  & \textbf{40.16 $\pm$ 6.18} & 37.50 $\pm$ 4.49 & \textbf{40.16 $\pm$ 6.18} & 65.69 $\pm$ 6.26 \\
\texttt{u\_dp\_\_x\_h\_\_u\_deco}
  & 25.20 $\pm$ 0.86 & 16.31 $\pm$ 1.77 & 25.20 $\pm$ 0.86 & 51.24 $\pm$ 0.50
  & 26.04 $\pm$ 0.07 & 22.09 $\pm$ 1.31 & 26.04 $\pm$ 0.07 & 49.78 $\pm$ 0.32 \\
\texttt{u\_dp\_\_x\_h\_\_n\_deco}
  & 28.12 $\pm$ 0.99 & 23.52 $\pm$ 3.09 & 28.12 $\pm$ 0.99 & 52.46 $\pm$ 0.57
  & 26.04 $\pm$ 0.28 & 19.57 $\pm$ 1.42 & 26.04 $\pm$ 0.28 & 53.04 $\pm$ 2.04 \\
\texttt{n\_dp\_\_x\_\_u\_deco}
  & 55.06 $\pm$ 5.31 & 52.65 $\pm$ 8.20 & 55.06 $\pm$ 5.31 & 81.31 $\pm$ 5.04
  & 36.49 $\pm$ 2.70 & 35.02 $\pm$ 2.24 & 36.49 $\pm$ 2.70 & 62.69 $\pm$ 2.92 \\
\texttt{n\_dp\_\_x\_\_n\_deco}
  & 53.22 $\pm$ 4.44 & 50.50 $\pm$ 5.52 & 53.22 $\pm$ 4.44 & 81.79 $\pm$ 1.09
  & 38.54 $\pm$ 3.87 & 36.72 $\pm$ 3.79 & 38.54 $\pm$ 3.87 & 64.42 $\pm$ 4.25 \\
\bottomrule
\end{tabular}
}
\vspace{0.5em}
\caption{Ablation study of \texttt{ddpm\_variant}, \texttt{encoder\_input} and \texttt{decoder\_variant} combinations in the MI task. Mean\,±\,std is reported separately for Accuracy, F1, Recall, and AUC on seen- and unseen-subject test splits across four tasks (``dp'' = ddpm\_variant, ``x\_h'' = x\_hat, ``deco'' = decoder\_variant,``u'' = use,``n'' = not use).}
\label{tab:mi_decoder_ablation_results}
\vspace{1em}
\end{table*}

\begin{table*}[htbp]
\centering

\vspace{0.5em}

\resizebox{\textwidth}{!}{
\begin{tabular}{p{3.8cm} *{8}{p{2cm}} }
\toprule
\textbf{Configuration}
& \multicolumn{4}{c}{\textbf{Imag. Speech (Seen)}}
& \multicolumn{4}{c}{\textbf{Imag. Speech (Unseen)}} \\
\cmidrule(lr){2-5} \cmidrule(lr){6-9}
& Acc (\%) & F1 (\%) & Recall (\%)& AUC (\%)
& Acc (\%) & F1 (\%) & Recall (\%)& AUC (\%) \\
\midrule
\texttt{u\_dp\_\_x\_\_u\_deco}
  & 15.55 $\pm$ 0.62 &  9.61 $\pm$ 1.23 & 15.53 $\pm$ 0.62 & 71.24 $\pm$ 1.66
  & 11.62 $\pm$ 1.29 &  7.84 $\pm$ 1.32 & 11.62 $\pm$ 1.29 & 57.15 $\pm$ 3.18 \\
\texttt{u\_dp\_\_x\_\_n\_deco}
  & 18.71 $\pm$ 1.43 & 13.92 $\pm$ 2.57 & 18.69 $\pm$ 1.46 & 72.81 $\pm$ 0.13
  & 10.98 $\pm$ 1.61 &  8.06 $\pm$ 0.55 & 10.98 $\pm$ 1.61 & 56.99 $\pm$ 4.34 \\
\texttt{u\_dp\_\_x\_h\_\_use\_deco}
  &  8.85 $\pm$ 0.37 &  1.72 $\pm$ 0.28 &  8.84 $\pm$ 0.36 & 51.77 $\pm$ 1.23
  &  9.47 $\pm$ 1.35 &  7.09 $\pm$ 0.63 &  9.47 $\pm$ 1.35 & 52.19 $\pm$ 2.60 \\
\texttt{u\_dp\_\_x\_h\_\_n\_deco}
  &  8.98 $\pm$ 0.16 &  2.22 $\pm$ 0.99 &  8.96 $\pm$ 0.18 & 49.93 $\pm$ 0.32
  &  8.33 $\pm$ 2.75 &  6.21 $\pm$ 1.83 &  8.33 $\pm$ 2.75 & 51.19 $\pm$ 3.38 \\
\texttt{n\_dp\_\_x\_\_u\_deco}
  & \textbf{19.22 $\pm$ 2.13} & 14.48 $\pm$ 1.41 & \textbf{19.20 $\pm$ 2.11} & \textbf{74.23 $\pm$ 1.24}
  & \textbf{13.76 $\pm$ 1.25} & \textbf{10.59 $\pm$ 0.42} & \textbf{13.76 $\pm$ 1.25} & \textbf{57.67 $\pm$ 3.22} \\
\texttt{n\_dp\_\_x\_\_n\_deco}
  & 19.22 $\pm$ 0.20 & \textbf{14.78 $\pm$ 0.77} & 19.19 $\pm$ 0.18 & 73.51 $\pm$ 1.08
  & 11.36 $\pm$ 1.93 &  7.95 $\pm$ 2.65 & 11.36 $\pm$ 1.93 & 54.54 $\pm$ 0.85 \\
\bottomrule
\end{tabular}
}
\vspace{0.5em}
\caption{Ablation study of \texttt{ddpm\_variant}, \texttt{encoder\_input} and \texttt{decoder\_variant} combinations in the imagined speech task. Mean\,±\,std is reported separately for Accuracy, F1, Recall, and AUC on seen- and unseen-subject test splits across four tasks (``dp'' = ddpm\_variant, ``x\_h'' = x\_hat, ``deco'' = decoder\_variant,``u'' = use,``n'' = not use).}
\label{tab:imagined_speech_results}
\vspace{1em}
\label{tab:imagined_speech_results}
\end{table*}

\clearpage  

\subsection{Loss combinations ablations}

\begin{table*}[htbp]
\centering
\vspace{0.5em}

\resizebox{\textwidth}{!}{
\begin{tabular}{p{6.3cm} *{8}{p{2cm}}}
\toprule
\textbf{Configuration}
& \multicolumn{4}{c}{\textbf{SSVEP (Seen)}}
& \multicolumn{4}{c}{\textbf{SSVEP (Unseen)}} \\
\cmidrule(lr){2-5} \cmidrule(lr){6-9}
& Acc (\%) & F1 (\%) & Recall (\%)& AUC (\%)
& Acc (\%) & F1 (\%) & Recall (\%)& AUC (\%)  \\
\midrule
\texttt{CE\_a0.5\_b0\_g0}
  & 84.27 $\pm$ 1.34 & 84.20 $\pm$ 1.39 & 84.27 $\pm$ 1.34 & 99.39 $\pm$ 0.12
  & 82.80 $\pm$ 5.98 & 82.45 $\pm$ 6.16 & 82.80 $\pm$ 5.98 & 98.95 $\pm$ 0.79 \\
\texttt{CE\_a0.5\_b0\_gsched 0.2}
  & 84.27 $\pm$ 1.65 & 84.14 $\pm$ 1.72 & 84.27 $\pm$ 1.65 & 99.39 $\pm$ 0.13
  & 81.62 $\pm$ 5.61 & 81.17 $\pm$ 5.81 & 81.62 $\pm$ 5.61 & 98.87 $\pm$ 0.81 \\
\texttt{CE\_a0.5\_bsched 0.05\_g0}
  & 86.01 $\pm$ 0.86 & 85.99 $\pm$ 0.89 & 86.01 $\pm$ 0.86 & 99.48 $\pm$ 0.02
  & 83.12 $\pm$ 5.75 & 82.68 $\pm$ 6.03 & 83.12 $\pm$ 5.75 & 98.89 $\pm$ 0.72 \\
\texttt{CE\_a0.5\_bsched 0.05\_gsched 0.2}
  & \textbf{86.40 $\pm$ 0.86} & \textbf{86.36 $\pm$ 0.92} & \textbf{86.40 $\pm$ 0.86} & \textbf{99.51 $\pm$ 0.04}
  & 83.23 $\pm$ 4.69 & 82.79 $\pm$ 5.00 & 83.23 $\pm$ 4.69 & \textbf{99.03 $\pm$ 0.64} \\
\texttt{CE\_a1\_b0\_g0}
  & 83.49 $\pm$ 1.72 & 83.38 $\pm$ 1.79 & 83.49 $\pm$ 1.72 & 99.32 $\pm$ 0.06
  & 84.83 $\pm$ 4.91 & 84.56 $\pm$ 5.01 & 84.83 $\pm$ 4.91 & 99.02 $\pm$ 0.73 \\
\texttt{CE\_a1\_b0\_gsched 0.2}
  & 84.46 $\pm$ 0.78 & 84.33 $\pm$ 0.87 & 84.46 $\pm$ 0.78 & 99.39 $\pm$ 0.08
  & 83.44 $\pm$ 5.53 & 82.93 $\pm$ 5.88 & 83.44 $\pm$ 5.53 & 98.89 $\pm$ 0.82 \\
\texttt{CE\_a1\_bsched 0.05\_g0}
  & 85.16 $\pm$ 0.95 & 85.14 $\pm$ 0.96 & 85.16 $\pm$ 0.95 & 99.40 $\pm$ 0.06
  & 84.40 $\pm$ 6.22 & 84.14 $\pm$ 6.36 & 84.40 $\pm$ 6.22 & 98.94 $\pm$ 0.80 \\
\texttt{CE\_a1\_bsched 0.05\_gsched 0.2}
  & 85.08 $\pm$ 1.53 & 84.95 $\pm$ 1.66 & 85.08 $\pm$ 1.53 & 99.37 $\pm$ 0.05
  & 84.72 $\pm$ 6.04 & 84.44 $\pm$ 6.21 & 84.72 $\pm$ 6.04 & 98.90 $\pm$ 0.76 \\
\texttt{MSE\_a0.5\_b0\_g0}
  & 85.39 $\pm$ 0.68 & 85.18 $\pm$ 0.90 & 85.39 $\pm$ 0.68 & 98.49 $\pm$ 0.10
  & \textbf{85.58 $\pm$ 8.39} & \textbf{85.49 $\pm$ 8.47} & \textbf{85.58 $\pm$ 8.39} & 98.19 $\pm$ 1.02 \\
\texttt{MSE\_a0.5\_b0\_gsched 0.2}
  & 84.97 $\pm$ 0.62 & 84.77 $\pm$ 0.61 & 84.97 $\pm$ 0.62 & 98.20 $\pm$ 0.16
  & 84.83 $\pm$ 7.03 & 84.67 $\pm$ 7.26 & 84.83 $\pm$ 7.03 & 97.60 $\pm$ 1.65 \\
\texttt{MSE\_a0.5\_bsched 0.05\_g0}
  & 84.50 $\pm$ 0.81 & 84.52 $\pm$ 0.81 & 84.50 $\pm$ 0.81 & 97.55 $\pm$ 0.45
  & 80.24 $\pm$ 5.75 & 79.88 $\pm$ 5.74 & 80.24 $\pm$ 5.75 & 96.89 $\pm$ 1.73 \\
\texttt{MSE\_a0.5\_bsched 0.05\_gsched 0.2}
  & 85.08 $\pm$ 0.76 & 85.09 $\pm$ 0.75 & 85.08 $\pm$ 0.76 & 97.30 $\pm$ 0.37
  & 81.09 $\pm$ 6.84 & 80.92 $\pm$ 6.93 & 81.09 $\pm$ 6.84 & 96.69 $\pm$ 1.10 \\
\texttt{MSE\_a1\_b0\_g0}
  & 85.66 $\pm$ 0.59 & 85.58 $\pm$ 0.68 & 85.66 $\pm$ 0.59 & 98.06 $\pm$ 0.11
  & 85.26 $\pm$ 7.57 & 85.00 $\pm$ 7.62 & 85.26 $\pm$ 7.57 & 98.12 $\pm$ 1.14 \\
\texttt{MSE\_a1\_b0\_gsched 0.2}
  & 85.70 $\pm$ 0.67 & 85.66 $\pm$ 0.68 & 85.70 $\pm$ 0.67 & 98.34 $\pm$ 0.12
  & 84.62 $\pm$ 8.44 & 84.26 $\pm$ 8.72 & 84.62 $\pm$ 8.44 & 97.70 $\pm$ 1.60 \\
\texttt{MSE\_a1\_bsched .05\_g0}
  & 86.21 $\pm$ 0.15 & 86.20 $\pm$ 0.11 & 86.21 $\pm$ 0.15 & 98.08 $\pm$ 0.27
  & 83.44 $\pm$ 6.83 & 83.18 $\pm$ 6.94 & 83.44 $\pm$ 6.83 & 97.07 $\pm$ 2.09 \\
\texttt{MSE\_a1\_bsched 0.05\_gsched 0.2}
  & 85.39 $\pm$ 0.15 & 85.37 $\pm$ 0.17 & 85.39 $\pm$ 0.15 & 98.10 $\pm$ 0.12
  & 82.80 $\pm$ 6.19 & 82.48 $\pm$ 6.34 & 82.80 $\pm$ 6.19 & 97.45 $\pm$ 1.51 \\
\bottomrule
\end{tabular}
}
\vspace{0.5em}
\caption{Ablation study of loss in the SSVEP task. Mean\,±\,std is reported separately for Accuracy, F1, Recall, and AUC on seen- and unseen-subject test splits across four tasks (``sched'' = scheduler to).}
\label{tab:ssvep_loss_ablation_results}
\vspace{1em}
\end{table*}

\begin{table*}[htbp]
\centering
\vspace{0.5em}

\resizebox{\textwidth}{!}{
\begin{tabular}{p{6.4cm} *{8}{p{2cm}} }
\toprule
\textbf{Configuration}
& \multicolumn{4}{c}{\textbf{P300 (Seen)}}
& \multicolumn{4}{c}{\textbf{P300 (Unseen)}} \\
\cmidrule(lr){2-5} \cmidrule(lr){6-9}
& Acc (\%) & F1 (\%) & Recall (\%)& AUC (\%)
& Acc (\%) & F1 (\%) & Recall (\%)& AUC (\%) \\
\midrule
\texttt{CE\_a0.5\_b0\_g0}
  & 85.15 $\pm$ 1.25 & 61.34 $\pm$ 9.17 & 60.20 $\pm$ 6.39 & 72.13 $\pm$ 8.49
  & 77.61 $\pm$ 5.30 & 59.52 $\pm$ 3.58 & 59.38 $\pm$ 2.61 & 66.68 $\pm$ 3.94 \\
\texttt{CE\_a0.5\_b0\_gsched 0.2}
  & 84.96 $\pm$ 1.07 & 61.36 $\pm$ 8.67 & 60.18 $\pm$ 6.04 & 72.52 $\pm$ 8.40
  & 77.15 $\pm$ 4.27 & 59.83 $\pm$ 3.08 & 60.14 $\pm$ 2.78 & 65.42 $\pm$ 3.51 \\
\texttt{CE\_a0.5\_bsched 0.05\_g0}
  & 85.36 $\pm$ 1.01 & 62.19 $\pm$ 7.03 & 60.36 $\pm$ 4.98 & 76.10 $\pm$ 5.36
  & 75.06 $\pm$ 2.99 & 61.63 $\pm$ 1.48 & 64.22 $\pm$ 1.48 & 70.90 $\pm$ 2.06 \\
\texttt{CE\_a0.5\_bsched 0.05\_gsched 0.2}
  & 85.51 $\pm$ 1.09 & 65.01 $\pm$ 3.74 & 62.21 $\pm$ 3.11 & 78.48 $\pm$ 4.32
  & 76.46 $\pm$ 1.12 & \textbf{63.76 $\pm$ 0.81} & \textbf{66.84 $\pm$ 2.65} & \textbf{73.06 $\pm$ 2.84} \\
\texttt{CE\_a1\_b0\_g0}
  & 84.56 $\pm$ 0.95 & 65.49 $\pm$ 2.69 & 63.26 $\pm$ 2.83 & 77.34 $\pm$ 1.00
  & \textbf{80.18 $\pm$ 0.50} & 63.07 $\pm$ 3.07 & 63.13 $\pm$ 4.36 & 71.95 $\pm$ 4.05 \\
\texttt{CE\_a1\_b0\_gsched 0.2}
  & 83.91 $\pm$ 1.49 & 65.53 $\pm$ 1.16 & \textbf{63.65 $\pm$ 1.92} & 75.12 $\pm$ 0.99
  & 78.19 $\pm$ 3.49 & 63.21 $\pm$ 3.97 & 63.99 $\pm$ 3.60 & 70.67 $\pm$ 5.66 \\
\texttt{CE\_a1\_bsched 0.05\_g0}
  & 85.46 $\pm$ 0.75 & \textbf{66.28 $\pm$ 2.44} & 63.42 $\pm$ 2.23 & \textbf{78.86 $\pm$ 2.41}
  & 79.22 $\pm$ 2.47 & 63.35 $\pm$ 3.07 & 63.66 $\pm$ 2.95 & 71.34 $\pm$ 4.79 \\
\texttt{CE\_a1\_bsched 0.05\_gsched 0.2}
  & 85.35 $\pm$ 1.12 & 64.39 $\pm$ 4.95 & 61.88 $\pm$ 4.10 & 77.14 $\pm$ 4.35
  & 79.47 $\pm$ 0.54 & 63.67 $\pm$ 1.58 & 63.95 $\pm$ 1.86 & 70.79 $\pm$ 3.27 \\
\texttt{MSE\_a0.5\_b0\_g0}
  & 84.70 $\pm$ 1.14 & 58.84 $\pm$ 9.64 & 58.44 $\pm$ 6.15 & 67.96 $\pm$ 9.11
  & 67.08 $\pm$ 13.35 & 54.56 $\pm$ 7.66 & 57.93 $\pm$ 4.25 & 60.24 $\pm$ 4.85 \\
\texttt{MSE\_a0.5\_b0\_gsched 0.2}
  & 83.90 $\pm$ 0.40 & 55.75 $\pm$ 7.38 & 55.99 $\pm$ 4.34 & 67.58 $\pm$ 8.14
  & 63.50 $\pm$ 9.96 & 52.50 $\pm$ 6.04 & 56.70 $\pm$ 3.72 & 59.72 $\pm$ 4.79 \\
\texttt{MSE\_a0.5\_bsched 0.05\_g0}
  & \textbf{85.69 $\pm$ 0.65} & 63.41 $\pm$ 3.27 & 60.73 $\pm$ 2.50 & 75.94 $\pm$ 2.75
  & 74.81 $\pm$ 3.74 & 61.87 $\pm$ 1.89 & 65.58 $\pm$ 4.44 & 69.66 $\pm$ 5.34 \\
\texttt{MSE\_a0.5\_bsched 0.05\_gsched 0.2}
  & 85.67 $\pm$ 0.52 & 65.01 $\pm$ 2.49 & 62.09 $\pm$ 2.09 & 75.30 $\pm$ 5.37
  & 78.17 $\pm$ 1.62 & 63.36 $\pm$ 1.79 & 64.83 $\pm$ 3.56 & 68.73 $\pm$ 4.26 \\
\texttt{MSE\_a1\_b0\_g0}
  & 84.24 $\pm$ 0.69 & 58.50 $\pm$ 9.22 & 58.31 $\pm$ 5.88 & 68.51 $\pm$ 8.11
  & 68.46 $\pm$ 10.36 & 54.56 $\pm$ 6.82 & 56.33 $\pm$ 5.40 & 59.00 $\pm$ 6.62 \\
\texttt{MSE\_a1\_b0\_gsched 0.2}
  & 84.68 $\pm$ 0.29 & 61.65 $\pm$ 1.34 & 59.45 $\pm$ 1.03 & 71.05 $\pm$ 1.04
  & 72.24 $\pm$ 2.85 & 59.71 $\pm$ 3.64 & 63.04 $\pm$ 4.17 & 67.29 $\pm$ 5.38 \\
\texttt{MSE\_a1\_bsched 0.05\_g0}
  & 85.20 $\pm$ 1.15 & 66.12 $\pm$ 1.40 & 63.29 $\pm$ 1.00 & 75.87 $\pm$ 2.28
  & 72.83 $\pm$ 2.19 & 61.17 $\pm$ 1.95 & 65.55 $\pm$ 3.36 & 69.81 $\pm$ 3.62 \\
\texttt{MSE\_a1\_bsched 0.05\_gsched 0.2}
  & 85.31 $\pm$ 0.56 & 64.17 $\pm$ 1.70 & 61.41 $\pm$ 1.32 & 75.70 $\pm$ 3.09
  & 74.35 $\pm$ 2.84 & 61.78 $\pm$ 2.66 & 65.57 $\pm$ 5.00 & 69.59 $\pm$ 5.21 \\
\bottomrule
\end{tabular}
}
\vspace{0.5em}
\caption{Ablation study of loss in the P300 task. Mean\,±\,std is reported separately for Accuracy, F1, Recall, and AUC on seen- and unseen-subject test splits across four tasks (``sched'' = scheduler to).}
\label{tab:p300_loss_ablation_results}
\vspace{1em}
\end{table*}

\begin{table*}[htbp]
\centering
\vspace{0.5em}

\resizebox{\textwidth}{!}{
\begin{tabular}{p{6.4cm} *{8}{p{2cm}} }

\toprule
\textbf{Configuration}
& \multicolumn{4}{c}{\textbf{MI (Seen)}}
& \multicolumn{4}{c}{\textbf{MI (Unseen)}} \\
\cmidrule(lr){2-5} \cmidrule(lr){6-9}
& Acc (\%) & F1 (\%) & Recall (\%)& AUC (\%)
& Acc (\%) & F1 (\%) & Recall (\%)& AUC (\%)  \\
\midrule
\texttt{CE\_a0.5\_b0\_g0}
  & 51.84 $\pm$ 2.36 & 47.19 $\pm$ 3.13 & 51.84 $\pm$ 2.36 & 81.38 $\pm$ 2.42
  & 39.15 $\pm$ 7.00 & 37.98 $\pm$ 6.49 & 39.15 $\pm$ 7.00 & 66.57 $\pm$ 6.63 \\
\texttt{CE\_a0.5\_b0\_gsched 0.2}
  & 53.87 $\pm$ 2.48 & 50.13 $\pm$ 3.19 & 53.87 $\pm$ 2.48 & 81.44 $\pm$ 1.65
  & 39.99 $\pm$ 6.23 & 39.21 $\pm$ 5.71 & 39.99 $\pm$ 6.23 & 66.25 $\pm$ 6.73 \\
\texttt{CE\_a0.5\_bsched 0.05\_g0}
  & 55.36 $\pm$ 2.11 & 53.21 $\pm$ 2.62 & 55.36 $\pm$ 2.11 & \textbf{83.07 $\pm$ 1.54}
  & 40.80 $\pm$ 6.38 & 39.21 $\pm$ 5.99 & 40.80 $\pm$ 6.38 & 67.29 $\pm$ 6.94 \\
\texttt{CE\_a0.5\_bsched 0.05\_gsched 0.2}
  & 54.76 $\pm$ 3.08 & 52.67 $\pm$ 4.43 & 54.76 $\pm$ 3.08 & 83.06 $\pm$ 1.33
  & 40.48 $\pm$ 6.61 & 39.39 $\pm$ 6.28 & 40.48 $\pm$ 6.61 & \textbf{67.68 $\pm$ 6.98} \\
\texttt{CE\_a1\_b0\_g0}
  & 52.38 $\pm$ 3.19 & 48.02 $\pm$ 3.37 & 52.38 $\pm$ 3.19 & 81.50 $\pm$ 2.66
  & 39.15 $\pm$ 4.69 & 37.77 $\pm$ 4.16 & 39.15 $\pm$ 4.69 & 65.42 $\pm$ 5.19 \\
\texttt{CE\_a1\_b0\_gsched 0.2}
  & 51.44 $\pm$ 3.86 & 46.19 $\pm$ 4.98 & 51.44 $\pm$ 3.86 & 80.63 $\pm$ 3.05
  & 39.18 $\pm$ 4.31 & 37.85 $\pm$ 3.56 & 39.18 $\pm$ 4.31 & 65.29 $\pm$ 5.10 \\
\texttt{CE\_a1\_bsched 0.05\_g0}
  & 54.46 $\pm$ 2.59 & 52.02 $\pm$ 3.63 & 54.46 $\pm$ 2.59 & 82.65 $\pm$ 1.20
  & 40.05 $\pm$ 7.45 & 38.91 $\pm$ 7.38 & 40.05 $\pm$ 7.45 & 67.01 $\pm$ 6.95 \\
\texttt{CE\_a1\_bsched 0.05\_gsched 0.2}
  & 55.85 $\pm$ 2.80 & 54.25 $\pm$ 3.15 & 55.85 $\pm$ 2.80 & 82.72 $\pm$ 0.86
  & 39.24 $\pm$ 7.95 & 38.07 $\pm$ 7.86 & 39.24 $\pm$ 7.95 & 66.70 $\pm$ 7.32 \\
\texttt{MSE\_a0.5\_b0\_g0}
  & 53.92 $\pm$ 5.58 & 51.22 $\pm$ 8.61 & 53.92 $\pm$ 5.58 & 79.86 $\pm$ 4.12
  & 37.33 $\pm$ 2.98 & 34.85 $\pm$ 1.05 & 37.33 $\pm$ 2.98 & 63.60 $\pm$ 4.09 \\
\texttt{MSE\_a0.5\_b0\_gsched 0.2}
  & 53.62 $\pm$ 5.23 & 50.63 $\pm$ 7.83 & 53.62 $\pm$ 5.23 & 79.80 $\pm$ 4.09
  & 37.59 $\pm$ 2.99 & 34.90 $\pm$ 1.34 & 37.59 $\pm$ 2.99 & 63.60 $\pm$ 3.91 \\
\texttt{MSE\_a0.5\_bsched 0.05\_g0}
  & 57.74 $\pm$ 1.48 & 55.92 $\pm$ 2.94 & 57.74 $\pm$ 1.48 & 79.98 $\pm$ 2.41
  & 40.89 $\pm$ 5.00 & 40.20 $\pm$ 4.83 & 40.89 $\pm$ 5.00 & 63.96 $\pm$ 3.68 \\
\texttt{MSE\_a0.5\_bsched 0.05\_gsched 0.2}
  & \textbf{59.67 $\pm$ 1.72} & \textbf{58.45 $\pm$ 2.70} & \textbf{59.67 $\pm$ 1.72} & 81.22 $\pm$ 1.05
  & 40.60 $\pm$ 4.03 & 39.56 $\pm$ 3.52 & 40.60 $\pm$ 4.03 & 63.57 $\pm$ 3.07 \\
\texttt{MSE\_a1\_b0\_g0}
  & 54.07 $\pm$ 5.19 & 51.81 $\pm$ 8.05 & 54.07 $\pm$ 5.19 & 80.34 $\pm$ 3.07
  & 37.30 $\pm$ 4.89 & 36.10 $\pm$ 4.59 & 37.30 $\pm$ 4.89 & 62.84 $\pm$ 5.47 \\
\texttt{MSE\_a1\_b0\_gsched 0.2}
  & 54.91 $\pm$ 5.02 & 52.24 $\pm$ 7.49 & 54.91 $\pm$ 5.02 & 81.03 $\pm$ 3.28
  & 37.33 $\pm$ 4.36 & 36.21 $\pm$ 4.11 & 37.33 $\pm$ 4.36 & 63.15 $\pm$ 4.89 \\
\texttt{MSE\_a1\_bsched 0.05\_g0}
  & 57.79 $\pm$ 1.30 & 57.15 $\pm$ 1.89 & 57.79 $\pm$ 1.30 & 81.44 $\pm$ 2.13
  & 39.32 $\pm$ 3.39 & 38.31 $\pm$ 2.84 & 39.32 $\pm$ 3.39 & 63.34 $\pm$ 3.57 \\
\texttt{MSE\_a1\_bsched 0.05\_gsched 0.2}
  & 59.23 $\pm$ 1.58 & 57.85 $\pm$ 3.33 & 59.23 $\pm$ 1.58 & 82.16 $\pm$ 1.11
  & \textbf{41.44 $\pm$ 6.03} & \textbf{40.37 $\pm$ 6.26} & \textbf{41.44 $\pm$ 6.03} & 65.18 $\pm$ 4.76 \\
\bottomrule
\end{tabular}
}
\vspace{0.5em}
\caption{Ablation study of loss in the MI task. Mean\,±\,std is reported separately for Accuracy, F1, Recall, and AUC on seen- and unseen-subject test splits across four tasks (``sched'' = scheduler to).}
\label{tab:mi_loss_ablation_results}

\vspace{1em}
\end{table*}

\begin{table*}[htbp]
\centering
\vspace{0.5em}

\resizebox{\textwidth}{!}{
\begin{tabular}{p{6.4cm} *{8}{p{2cm}} }
\toprule
\textbf{Configuration}
& \multicolumn{4}{c}{\textbf{Imag. Speech (Seen)}}
& \multicolumn{4}{c}{\textbf{Imag. Speech (Unseen)}} \\
\cmidrule(lr){2-5} \cmidrule(lr){6-9}
& Acc & F1 & Recall & AUC
& Acc & F1 & Recall & AUC \\
\midrule
\texttt{CE\_a0.5\_b0\_g0}
  & 17.83 $\pm$ 1.38 & 13.40 $\pm$ 2.54 & 17.80 $\pm$ 1.35 & 70.88 $\pm$ 1.42
  & 13.26 $\pm$ 0.82 &  9.98 $\pm$ 1.14 & 13.26 $\pm$ 0.82 & 57.90 $\pm$ 2.93 \\
\texttt{CE\_a0.5\_b0\_gsched 0.2}
  & 19.09 $\pm$ 0.75 & \textbf{15.08 $\pm$ 0.99} & 19.07 $\pm$ 0.78 & 72.13 $\pm$ 1.37
  & 11.99 $\pm$ 2.87 &  9.08 $\pm$ 2.30 & 11.99 $\pm$ 2.87 & \textbf{59.72 $\pm$ 1.83} \\
\texttt{CE\_a0.5\_bsched 0.05\_g0}
  & 18.46 $\pm$ 0.68 & 15.04 $\pm$ 0.64 & 18.43 $\pm$ 0.64 & 71.98 $\pm$ 0.72
  & 11.24 $\pm$ 0.99 &  8.54 $\pm$ 1.38 & 11.24 $\pm$ 0.99 & 56.68 $\pm$ 3.58 \\
\texttt{CE\_a0.5\_bsched 0.05\_gsched 0.2}
  & \textbf{19.60 $\pm$ 1.62} & 13.61 $\pm$ 1.07 & \textbf{19.57 $\pm$ 1.59} & \textbf{73.49 $\pm$ 0.24}
  &  9.34 $\pm$ 3.16 &  5.83 $\pm$ 1.34 &  9.34 $\pm$ 3.16 & 56.33 $\pm$ 3.92 \\
\texttt{CE\_a1\_b0\_g0}
  & 18.08 $\pm$ 1.28 & 12.39 $\pm$ 0.35 & 18.06 $\pm$ 1.25 & 72.35 $\pm$ 0.89
  & 11.49 $\pm$ 1.96 &  9.18 $\pm$ 2.13 & 11.49 $\pm$ 1.96 & 59.17 $\pm$ 1.89 \\
\texttt{CE\_a1\_b0\_gsched 0.2}
  & 17.95 $\pm$ 0.70 & 13.52 $\pm$ 1.18 & 17.93 $\pm$ 0.71 & 73.07 $\pm$ 0.40
  & 11.87 $\pm$ 1.53 &  8.18 $\pm$ 2.56 & 11.87 $\pm$ 1.53 & 59.12 $\pm$ 2.40 \\
\texttt{CE\_a1\_bsched 0.05\_g0}
  & 17.44 $\pm$ 1.70 & 11.50 $\pm$ 0.76 & 17.42 $\pm$ 1.72 & 71.56 $\pm$ 2.03
  & 12.12 $\pm$ 2.23 &  8.78 $\pm$ 1.11 & 12.12 $\pm$ 2.23 & 57.33 $\pm$ 3.33 \\
\texttt{CE\_a1\_bsched 0.05\_gsched 0.2}
  & 15.55 $\pm$ 0.62 &  9.61 $\pm$ 1.23 & 15.53 $\pm$ 0.62 & 71.24 $\pm$ 1.66
  & 11.62 $\pm$ 1.29 &  7.84 $\pm$ 1.32 & 11.62 $\pm$ 1.29 & 57.15 $\pm$ 3.18 \\
\texttt{MSE\_a0.5\_b0\_g0}
  & 16.69 $\pm$ 1.32 & 11.20 $\pm$ 0.62 & 16.67 $\pm$ 1.35 & 66.83 $\pm$ 5.71
  & 11.99 $\pm$ 0.47 &  8.54 $\pm$ 1.26 & 11.99 $\pm$ 0.47 & 55.95 $\pm$ 1.92 \\
\texttt{MSE\_a0.5\_b0\_gsched 0.2}
  & 16.82 $\pm$ 1.98 & 11.22 $\pm$ 0.64 & 16.79 $\pm$ 1.96 & 64.87 $\pm$ 4.18
  & 11.87 $\pm$ 1.39 &  7.68 $\pm$ 2.71 & 11.87 $\pm$ 1.39 & 54.02 $\pm$ 2.17 \\
\texttt{MSE\_a0.5\_bsched 0.05\_g0}
  & 17.19 $\pm$ 0.81 & 10.68 $\pm$ 1.75 & 17.17 $\pm$ 0.78 & 70.09 $\pm$ 5.01
  &  9.60 $\pm$ 1.53 &  5.44 $\pm$ 0.55 &  9.60 $\pm$ 1.53 & 52.43 $\pm$ 4.56 \\
\texttt{MSE\_a0.5\_bsched 0.05\_gsched 0.2}
  & 17.32 $\pm$ 0.80 & 12.48 $\pm$ 1.70 & 17.33 $\pm$ 0.81 & 61.80 $\pm$ 7.84
  & 11.62 $\pm$ 1.17 &  8.85 $\pm$ 1.36 & 11.62 $\pm$ 1.17 & 53.32 $\pm$ 0.93 \\
\texttt{MSE\_a1\_b0\_g0}
  & 16.43 $\pm$ 0.98 & 12.53 $\pm$ 0.65 & 16.43 $\pm$ 0.99 & 62.13 $\pm$ 5.47
  & \textbf{13.51 $\pm$ 2.36} & \textbf{10.05 $\pm$ 1.92} & \textbf{13.51 $\pm$ 2.36} & 53.55 $\pm$ 2.59 \\
\texttt{MSE\_a1\_b0\_gsched 0.2}
  & 16.06 $\pm$ 2.01 & 11.90 $\pm$ 1.40 & 16.10 $\pm$ 2.04 & 65.92 $\pm$ 4.70
  & 13.26 $\pm$ 1.93 &  9.80 $\pm$ 1.77 & 13.26 $\pm$ 1.93 & 54.53 $\pm$ 1.07 \\
\texttt{MSE\_a1\_bsched 0.05\_g0}
  & 17.32 $\pm$ 1.11 & 12.89 $\pm$ 1.93 & 17.31 $\pm$ 1.11 & 61.35 $\pm$ 4.49
  &  9.97 $\pm$ 2.52 &  6.64 $\pm$ 1.24 &  9.97 $\pm$ 2.52 & 50.25 $\pm$ 1.56 \\
\texttt{MSE\_a1\_bsched 0.05\_gsched 0.2}
  & 17.57 $\pm$ 0.67 & 13.19 $\pm$ 1.56 & 17.58 $\pm$ 0.68 & 61.76 $\pm$ 4.09
  & 12.12 $\pm$ 2.70 &  7.66 $\pm$ 2.48 & 12.12 $\pm$ 2.70 & 48.50 $\pm$ 1.67 \\
\bottomrule
\end{tabular}
}
\vspace{0.5em}
\caption{Ablation study of loss in the imagined speech task. Mean\,±\,std is reported separately for Accuracy, F1, Recall, and AUC on seen- and unseen-subject test splits across four tasks (``sched'' = scheduler to).}
\label{tab:kara_one_loss_ablation_results}
\vspace{1em}
\end{table*}

\clearpage  

\subsection{Mixup strategies ablations}

\begin{table*}[htbp]
\centering
\vspace{0.5em}

\resizebox{\textwidth}{!}{
\begin{tabular}{p{2.2cm} *{8}{p{2cm}} }
\toprule
\textbf{Configuration (mixup layer/ warmup epoch/ random ratio)} 
& \multicolumn{4}{c}{\textbf{SSVEP (Seen)}} 
& \multicolumn{4}{c}{\textbf{SSVEP (Unseen)}} \\
\cmidrule(lr){2-5} \cmidrule(lr){6-9}
& Acc (\%) & F1 (\%) & Recall (\%) & AUC (\%) 
& Acc (\%) & F1 (\%) & Recall (\%) & AUC (\%) \\
\midrule
-1 / 100 / No 
  & 86.79 $\pm$ 1.75 & 86.75 $\pm$ 1.83 & 86.79 $\pm$ 1.75 & \textbf{99.62 $\pm$ 0.02}
  & \textbf{85.26 $\pm$ 6.94} & 85.02 $\pm$ 7.05 & \textbf{85.26 $\pm$ 6.94} & \textbf{99.19 $\pm$ 0.81} \\
-1 / 100 / Yes
  & 86.75 $\pm$ 0.78 & 86.74 $\pm$ 0.78 & 86.75 $\pm$ 0.78 & 99.60 $\pm$ 0.02 
  & 85.15 $\pm$ 7.46 & \textbf{85.09 $\pm$ 7.32} & 85.15 $\pm$ 7.46 & 99.16 $\pm$ 0.92 \\
-1 / 150 / No 
  & 86.87 $\pm$ 1.77 & 86.86 $\pm$ 1.81 & 86.87 $\pm$ 1.77 & 99.55 $\pm$ 0.07 
  & 84.40 $\pm$ 7.65 & 84.07 $\pm$ 7.87 & 84.40 $\pm$ 7.65 & 99.02 $\pm$ 1.13 \\
-1 / 150 / Yes
  & 86.60 $\pm$ 0.42 & 86.62 $\pm$ 0.44 & 86.60 $\pm$ 0.42 & 99.54 $\pm$ 0.07 
  & 84.29 $\pm$ 7.09 & 84.08 $\pm$ 7.13 & 84.29 $\pm$ 7.09 & 99.06 $\pm$ 1.09 \\
0 / 100 / No 
  & 86.60 $\pm$ 1.07 & 86.59 $\pm$ 1.08 & 86.60 $\pm$ 1.07 & 99.46 $\pm$ 0.13 
  & 84.51 $\pm$ 7.68 & 84.14 $\pm$ 7.92 & 84.51 $\pm$ 7.68 & 98.99 $\pm$ 1.12 \\
0 / 100 / Yes
  & 86.60 $\pm$ 1.12 & 86.56 $\pm$ 1.09 & 86.60 $\pm$ 1.12 & 99.58 $\pm$ 0.04 
  & 84.72 $\pm$ 6.48 & 84.54 $\pm$ 6.51 & 84.72 $\pm$ 6.48 & 99.15 $\pm$ 0.92 \\
0 / 150 / No 
  & 87.36 $\pm$ 1.34 & 86.63 $\pm$ 0.54 & 86.64 $\pm$ 0.53 & 99.46 $\pm$ 0.11 
  & 83.65 $\pm$ 8.38 & 83.46 $\pm$ 8.44 & 83.65 $\pm$ 8.38 & 99.00 $\pm$ 1.11 \\
0 / 150 / Yes
  & \textbf{87.84 $\pm$ 0.86} & \textbf{87.89 $\pm$ 0.83} & \textbf{87.84 $\pm$ 0.86} & 99.58 $\pm$ 0.01 
  & 85.04 $\pm$ 7.94 & 84.85 $\pm$ 8.10 & 85.04 $\pm$ 7.94 & 99.11 $\pm$ 1.05 \\
1 / 100 / No 
  & 85.08 $\pm$ 1.62 & 85.11 $\pm$ 1.62 & 85.08 $\pm$ 1.62 & 99.34 $\pm$ 0.13 
  & 84.51 $\pm$ 6.85 & 84.35 $\pm$ 6.91 & 84.51 $\pm$ 6.85 & 99.03 $\pm$ 0.97 \\
1 / 100 / Yes
  & 84.65 $\pm$ 2.36 & 84.57 $\pm$ 2.55 & 84.65 $\pm$ 2.36 & 99.40 $\pm$ 0.11 
  & 82.69 $\pm$ 5.94 & 82.52 $\pm$ 5.95 & 82.69 $\pm$ 5.94 & 98.98 $\pm$ 0.93 \\
1 / 150 / No 
  & 85.47 $\pm$ 0.74 & 85.43 $\pm$ 0.82 & 85.47 $\pm$ 0.74 & 99.42 $\pm$ 0.07 
  & 83.33 $\pm$ 6.99 & 83.04 $\pm$ 7.16 & 83.33 $\pm$ 6.99 & 98.95 $\pm$ 1.12 \\
1 / 150 / Yes
  & 85.16 $\pm$ 0.66 & 85.09 $\pm$ 0.72 & 85.16 $\pm$ 0.66 & 99.42 $\pm$ 0.07 
  & 83.55 $\pm$ 7.21 & 83.31 $\pm$ 7.45 & 83.55 $\pm$ 7.21 & 99.00 $\pm$ 1.16 \\
2 / 100 / No 
  & 85.12 $\pm$ 1.55 & 85.17 $\pm$ 1.52 & 85.12 $\pm$ 1.55 & 99.36 $\pm$ 0.12 
  & 84.19 $\pm$ 6.59 & 83.99 $\pm$ 6.61 & 84.19 $\pm$ 6.59 & 99.00 $\pm$ 0.95 \\
2 / 100 / Yes
  & 85.12 $\pm$ 2.06 & 84.90 $\pm$ 2.59 & 85.12 $\pm$ 2.06 & 99.44 $\pm$ 0.15 
  & 84.40 $\pm$ 7.59 & 84.26 $\pm$ 7.71 & 84.40 $\pm$ 7.59 & 99.12 $\pm$ 0.79 \\
2 / 150 / No 
  & 85.47 $\pm$ 0.74 & 85.43 $\pm$ 0.82 & 85.47 $\pm$ 0.74 & 99.42 $\pm$ 0.07 
  & 83.33 $\pm$ 6.99 & 82.38 $\pm$ 6.55 & 83.33 $\pm$ 6.99 & 98.95 $\pm$ 1.12 \\
2 / 150 / Yes
  & 84.89 $\pm$ 0.91 & 84.80 $\pm$ 0.97 & 84.89 $\pm$ 0.91 & 99.43 $\pm$ 0.05 
  & 83.97 $\pm$ 7.70 & 83.74 $\pm$ 7.94 & 83.97 $\pm$ 7.70 & 98.99 $\pm$ 1.15 \\
3 / 100 / No 
  & 84.77 $\pm$ 2.16 & 84.72 $\pm$ 2.28 & 84.77 $\pm$ 2.16 & 99.36 $\pm$ 0.12 
  & 83.01 $\pm$ 5.99 & 82.84 $\pm$ 6.01 & 83.01 $\pm$ 5.99 & 99.03 $\pm$ 0.97 \\
3 / 100 / Yes
  & 84.65 $\pm$ 2.36 & 84.57 $\pm$ 2.55 & 84.65 $\pm$ 2.36 & 99.40 $\pm$ 0.11 
  & 82.69 $\pm$ 5.94 & 82.52 $\pm$ 5.95 & 82.69 $\pm$ 5.94 & 98.98 $\pm$ 0.93 \\
3 / 150 / No 
  & 85.47 $\pm$ 0.74 & 85.43 $\pm$ 0.82 & 85.47 $\pm$ 0.74 & 99.42 $\pm$ 0.07 
  & 83.33 $\pm$ 6.99 & 83.04 $\pm$ 7.16 & 83.33 $\pm$ 6.99 & 98.95 $\pm$ 1.12 \\
3 / 150 / Yes
  & 85.47 $\pm$ 0.74 & 85.43 $\pm$ 0.82 & 85.47 $\pm$ 0.74 & 99.42 $\pm$ 0.07 
  & 83.33 $\pm$ 6.99 & 83.04 $\pm$ 7.16 & 83.33 $\pm$ 6.99 & 98.95 $\pm$ 1.12 \\
4 / 100 / No 
  & 85.70 $\pm$ 1.50 & 85.70 $\pm$ 1.61 & 85.70 $\pm$ 1.50 & 99.42 $\pm$ 0.13 
  & 83.76 $\pm$ 8.14 & 83.60 $\pm$ 8.15 & 83.76 $\pm$ 8.14 & 98.98 $\pm$ 1.07 \\
4 / 100 / Yes
  & 84.85 $\pm$ 2.53 & 84.84 $\pm$ 2.78 & 84.85 $\pm$ 2.53 & 99.39 $\pm$ 0.10 
  & 82.26 $\pm$ 6.60 & 82.14 $\pm$ 6.54 & 82.26 $\pm$ 6.60 & 99.02 $\pm$ 0.86 \\
4 / 150 / No 
  & 85.78 $\pm$ 1.34 & 85.81 $\pm$ 1.44 & 85.78 $\pm$ 1.34 & 99.43 $\pm$ 0.08 
  & 83.23 $\pm$ 6.91 & 83.01 $\pm$ 6.78 & 83.23 $\pm$ 6.91 & 99.02 $\pm$ 1.03 \\
4 / 150 / Yes
  & 85.55 $\pm$ 0.82 & 85.56 $\pm$ 0.95 & 85.55 $\pm$ 0.82 & 99.46 $\pm$ 0.11 
  & 83.33 $\pm$ 6.99 & 82.99 $\pm$ 7.24 & 83.33 $\pm$ 6.99 & 99.08 $\pm$ 0.91 \\

\bottomrule
\end{tabular}
}
\vspace{0.5em}
\caption{Ablation study of mixup in the SSVEP task. Mean$\pm$std is reported separately for Accuracy, F1, Recall, and AUC on seen- and unseen-subject test splits across four tasks (mixup layer: -1 = temporal mixup at input, 0 = weighted average at input, 1/2/3 = weighted average after first/second/third encoder layer, 4 = weighted average after attention pooling. warmup epoch: number of epochs to train the generators before training the classifier. random ratio: No = equal possibility on choosing ddpm or decoder out for temporal mixup/equal weight for weighted average mixup, Yes = beta/dirichlet distribution with b=0.2 for a random ratio more heavily tilted towards one of the mixup candidates).}
\label{tab:ssvep_mixup_results}

\vspace{1em}
\end{table*}

\begin{table*}[htbp]
\centering
\vspace{0.5em}

\resizebox{\textwidth}{!}{
\begin{tabular}{p{2.2cm} *{8}{p{2cm}} }
\toprule
\textbf{Configuration (mixup layer/ warmup epoch/ random ratio)} 
& \multicolumn{4}{c}{\textbf{P300 (Seen)}} 
& \multicolumn{4}{c}{\textbf{P300 (Unseen)}} \\
\cmidrule(lr){2-5} \cmidrule(lr){6-9}
& Acc (\%) & F1 (\%) & Recall (\%) & AUC (\%) 
& Acc (\%) & F1 (\%) & Recall (\%) & AUC (\%) \\
\midrule
-1 / 100 / No 
  & 85.61 $\pm$ 0.52 & 67.30 $\pm$ 1.54 & 64.42 $\pm$ 2.02 & \textbf{80.00 $\pm$ 1.39}
  & \textbf{79.56 $\pm$ 4.43} & 65.19 $\pm$ 7.51 & 66.10 $\pm$ 7.91 & 72.72 $\pm$ 9.74 \\
-1 / 100 / Yes
  & 85.58 $\pm$ 0.47 & \textbf{67.89 $\pm$ 0.71} & \textbf{64.98 $\pm$ 0.71} & 79.94 $\pm$ 1.61 
  & 79.51 $\pm$ 4.51 & \textbf{65.22 $\pm$ 7.57} & 66.18 $\pm$ 7.97 & 73.41 $\pm$ 9.22 \\
-1 / 150 / No 
  & 85.70 $\pm$ 0.28 & 66.38 $\pm$ 0.92 & 63.30 $\pm$ 0.84 & 78.55 $\pm$ 1.24 
  & 78.05 $\pm$ 6.49 & 64.40 $\pm$ 7.89 & 65.74 $\pm$ 7.33 & 71.71 $\pm$ 9.98 \\
-1 / 150 / Yes
  & \textbf{85.78 $\pm$ 0.13} & 66.30 $\pm$ 0.51 & 63.18 $\pm$ 0.56 & 79.05 $\pm$ 0.35 
  & 79.10 $\pm$ 2.04 & 64.31 $\pm$ 5.87 & 65.65 $\pm$ 7.85 & 71.52 $\pm$ 9.12 \\
0 / 100 / No 
  & 85.70 $\pm$ 0.44 & 66.60 $\pm$ 1.84 & 63.59 $\pm$ 1.88 & 78.87 $\pm$ 1.90 
  & 78.25 $\pm$ 6.85 & 64.98 $\pm$ 8.43 & 66.65 $\pm$ 8.39 & 72.34 $\pm$ 10.36 \\
0 / 100 / Yes
  & 85.54 $\pm$ 0.19 & 66.38 $\pm$ 1.54 & 63.43 $\pm$ 1.47 & 79.63 $\pm$ 1.54 
  & 75.11 $\pm$ 5.28 & 63.41 $\pm$ 6.66 & 67.46 $\pm$ 8.57 & 73.52 $\pm$ 10.37 \\
0 / 150 / No 
  & \textbf{85.78 $\pm$ 0.41} & 67.00 $\pm$ 1.29 & 63.91 $\pm$ 1.27 & 78.73 $\pm$ 1.45 
  & 78.10 $\pm$ 6.48 & 64.64 $\pm$ 8.28 & 66.21 $\pm$ 8.24 & 72.06 $\pm$ 10.14 \\
0 / 150 / Yes
  & 85.66 $\pm$ 0.25 & 66.60 $\pm$ 0.89 & 63.55 $\pm$ 0.81 & 79.48 $\pm$ 1.97 
  & 75.81 $\pm$ 3.96 & 63.38 $\pm$ 6.33 & 66.85 $\pm$ 8.63 & 72.89 $\pm$ 10.04 \\
1 / 100 / No 
  & 85.03 $\pm$ 0.33 & 62.05 $\pm$ 2.81 & 59.77 $\pm$ 2.19 & 76.15 $\pm$ 1.21 
  & 70.27 $\pm$ 3.61 & 59.38 $\pm$ 3.95 & 64.29 $\pm$ 4.65 & 68.67 $\pm$ 5.30 \\
1 / 100 / Yes
  & 85.48 $\pm$ 0.19 & 66.53 $\pm$ 1.64 & 63.66 $\pm$ 1.84 & 79.16 $\pm$ 1.17 
  & 66.82 $\pm$ 3.51 & 58.12 $\pm$ 1.03 & 66.45 $\pm$ 5.63 & 72.31 $\pm$ 8.49 \\
1 / 150 / No 
  & 85.02 $\pm$ 0.91 & 61.30 $\pm$ 5.42 & 59.35 $\pm$ 4.08 & 74.17 $\pm$ 5.60 
  & 72.35 $\pm$ 3.03 & 60.51 $\pm$ 3.10 & 64.56 $\pm$ 3.92 & 69.57 $\pm$ 5.41 \\
1 / 150 / Yes
  & 85.23 $\pm$ 0.99 & 64.38 $\pm$ 5.42 & 61.93 $\pm$ 4.45 & 77.76 $\pm$ 3.93 
  & 71.72 $\pm$ 2.41 & 60.98 $\pm$ 3.39 & 66.22 $\pm$ 4.78 & 71.21 $\pm$ 6.19 \\
2 / 100 / No 
  & 84.55 $\pm$ 0.97 & 57.31 $\pm$ 7.76 & 56.75 $\pm$ 4.83 & 71.05 $\pm$ 10.48 
  & 70.83 $\pm$ 1.12 & 58.46 $\pm$ 0.39 & 62.09 $\pm$ 1.45 & 66.24 $\pm$ 2.34 \\
2 / 100 / Yes
  & 84.84 $\pm$ 0.80 & 61.74 $\pm$ 5.07 & 59.71 $\pm$ 3.79 & 76.45 $\pm$ 3.50 
  & 72.20 $\pm$ 1.37 & 60.54 $\pm$ 2.63 & 64.91 $\pm$ 4.27 & 70.24 $\pm$ 5.46 \\
2 / 150 / No 
  & 84.73 $\pm$ 0.75 & 60.25 $\pm$ 4.67 & 58.52 $\pm$ 3.42 & 74.52 $\pm$ 5.66 
  & 74.35 $\pm$ 0.65 & 61.03 $\pm$ 2.28 & 63.97 $\pm$ 4.40 & 68.81 $\pm$ 5.90 \\
2 / 150 / Yes
  & 84.87 $\pm$ 0.79 & 62.09 $\pm$ 5.56 & 60.04 $\pm$ 4.15 & 74.71 $\pm$ 5.79 
  & 73.87 $\pm$ 0.18 & 61.51 $\pm$ 2.21 & 65.20 $\pm$ 4.06 & 70.76 $\pm$ 5.20 \\
3 / 100 / No 
  & 84.86 $\pm$ 0.44 & 61.05 $\pm$ 1.52 & 58.93 $\pm$ 1.09 & 76.08 $\pm$ 0.75 
  & 70.95 $\pm$ 4.27 & 60.24 $\pm$ 4.92 & 65.37 $\pm$ 5.98 & 69.56 $\pm$ 7.12 \\
3 / 100 / Yes
  & 85.16 $\pm$ 0.17 & 65.07 $\pm$ 2.10 & 62.40 $\pm$ 2.02 & 78.23 $\pm$ 1.34 
  & 71.84 $\pm$ 6.59 & 62.04 $\pm$ 7.10 & \textbf{67.89 $\pm$ 7.38} & 73.22 $\pm$ 9.40 \\
3 / 150 / No 
  & 84.99 $\pm$ 0.41 & 62.08 $\pm$ 1.94 & 59.77 $\pm$ 1.62 & 76.95 $\pm$ 1.70 
  & 72.01 $\pm$ 4.05 & 61.35 $\pm$ 5.41 & 66.56 $\pm$ 6.89 & 71.09 $\pm$ 8.62 \\
3 / 150 / Yes
  & 85.27 $\pm$ 0.45 & 62.82 $\pm$ 3.39 & 60.39 $\pm$ 2.83 & 77.27 $\pm$ 1.96 
  & 73.99 $\pm$ 4.67 & 62.18 $\pm$ 5.96 & 66.33 $\pm$ 7.83 & 71.19 $\pm$ 10.09 \\
4 / 100 / No 
  & 85.47 $\pm$ 0.11 & 65.16 $\pm$ 2.15 & 62.30 $\pm$ 1.91 & 78.26 $\pm$ 1.36 
  & 72.84 $\pm$ 5.44 & 62.07 $\pm$ 5.67 & 67.37 $\pm$ 7.83 & 72.55 $\pm$ 8.94 \\
4 / 100 / Yes
  & 85.48 $\pm$ 0.07 & 65.64 $\pm$ 1.41 & 62.72 $\pm$ 1.34 & 79.56 $\pm$ 0.34 
  & 76.03 $\pm$ 0.72 & 63.23 $\pm$ 4.99 & 67.16 $\pm$ 9.38 & \textbf{73.91 $\pm$ 9.71} \\
4 / 150 / No 
  & 85.00 $\pm$ 0.91 & 62.81 $\pm$ 6.22 & 60.69 $\pm$ 4.74 & 75.11 $\pm$ 6.29 
  & 74.24 $\pm$ 2.88 & 61.90 $\pm$ 3.60 & 65.40 $\pm$ 4.55 & 71.28 $\pm$ 5.96 \\
4 / 150 / Yes
  & 85.26 $\pm$ 0.59 & 64.19 $\pm$ 3.92 & 61.60 $\pm$ 3.19 & 77.58 $\pm$ 2.92 
  & 75.38 $\pm$ 4.29 & 62.57 $\pm$ 4.95 & 65.28 $\pm$ 4.82 & 71.59 $\pm$ 6.93 \\

\bottomrule
\end{tabular}
}
\vspace{0.5em}
\caption{Ablation study of mixup in the P300 task. Mean$\pm$std is reported separately for Accuracy, F1, Recall, and AUC on seen- and unseen-subject test splits across four tasks (mixup layer: -1 = temporal mixup at input, 0 = weighted average at input, 1/2/3 = weighted average after first/second/third encoder layer, 4 = weighted average after attention pooling. warmup epoch: number of epochs to train the generators before training the classifier. random ratio: No = equal possibility on choosing ddpm or decoder out for temporal mixup/equal weight for weighted average mixup, Yes = beta/dirichlet distribution with b=0.2 for a random ratio more heavily tilted towards one of the mixup candidates).}
\label{tab:p300_mixup_results}
\vspace{1em}
\end{table*}

\begin{table*}[htbp]
\vspace{0.5em}

\resizebox{\textwidth}{!}{
\begin{tabular}{p{2.2cm} *{8}{p{2cm}} }
\toprule
\textbf{Configuration (mixup layer/ warmup epoch/ random ratio)} 
& \multicolumn{4}{c}{\textbf{MI (Seen)}} 
& \multicolumn{4}{c}{\textbf{MI (Unseen)}} \\
\cmidrule(lr){2-5} \cmidrule(lr){6-9}
& Acc (\%) & F1 (\%) & Recall (\%) & AUC (\%) 
& Acc (\%) & F1 (\%) & Recall (\%) & AUC (\%) \\
\midrule
-1 / 100 / No 
  & 57.69 $\pm$ 3.27 & 55.70 $\pm$ 4.59 & 57.69 $\pm$ 3.27 & 84.97 $\pm$ 1.19
  & 36.78 $\pm$ 5.22 & 35.94 $\pm$ 5.62 & 36.78 $\pm$ 5.22 & 63.44 $\pm$ 5.17 \\
-1 / 100 / Yes
  & 58.33 $\pm$ 2.10 & 56.58 $\pm$ 3.53 & 58.33 $\pm$ 2.10 & 84.52 $\pm$ 1.20 
  & 36.23 $\pm$ 5.56 & 35.56 $\pm$ 5.90 & 36.23 $\pm$ 5.56 & 63.44 $\pm$ 5.01 \\
-1 / 150 / No 
  & 58.13 $\pm$ 4.14 & 56.10 $\pm$ 5.03 & 58.13 $\pm$ 4.14 & 85.32 $\pm$ 1.64 
  & 35.53 $\pm$ 3.03 & 34.51 $\pm$ 3.21 & 35.53 $\pm$ 3.03 & 62.92 $\pm$ 3.83 \\
-1 / 150 / Yes
  & 57.44 $\pm$ 4.36 & 55.81 $\pm$ 5.18 & 57.44 $\pm$ 4.36 & 85.03 $\pm$ 2.00 
  & 36.49 $\pm$ 5.74 & 35.68 $\pm$ 6.36 & 36.49 $\pm$ 5.74 & 63.62 $\pm$ 5.23 \\
0 / 100 / No 
  & 62.75 $\pm$ 6.07 & 62.71 $\pm$ 5.90 & 62.75 $\pm$ 6.07 & 85.84 $\pm$ 4.03 
  & 36.83 $\pm$ 8.41 & 36.01 $\pm$ 8.52 & 36.83 $\pm$ 8.41 & 63.51 $\pm$ 6.71 \\
0 / 100 / Yes
  & 62.25 $\pm$ 3.33 & 62.32 $\pm$ 3.18 & 62.25 $\pm$ 3.33 & 86.28 $\pm$ 2.31 
  & 38.14 $\pm$ 8.56 & 37.99 $\pm$ 8.69 & 38.14 $\pm$ 8.56 & 63.64 $\pm$ 5.84 \\
0 / 150 / No 
  & \textbf{63.44 $\pm$ 3.44} & \textbf{63.44 $\pm$ 3.36} & \textbf{63.44 $\pm$ 3.44} & \textbf{87.20 $\pm$ 2.69} 
  & 35.45 $\pm$ 6.82 & 34.63 $\pm$ 6.56 & 35.45 $\pm$ 6.82 & 62.40 $\pm$ 5.07 \\
0 / 150 / Yes
  & 62.95 $\pm$ 5.94 & 62.97 $\pm$ 5.95 & 62.95 $\pm$ 5.94 & 86.71 $\pm$ 3.13 
  & 36.17 $\pm$ 6.50 & 35.66 $\pm$ 6.94 & 36.17 $\pm$ 6.50 & 63.10 $\pm$ 5.78 \\
1 / 100 / No 
  & 59.38 $\pm$ 3.80 & 59.23 $\pm$ 3.95 & 59.38 $\pm$ 3.80 & 83.61 $\pm$ 2.36 
  & 36.08 $\pm$ 4.55 & 35.17 $\pm$ 5.20 & 36.08 $\pm$ 4.55 & 63.93 $\pm$ 5.77 \\
1 / 100 / Yes
  & 61.56 $\pm$ 3.06 & 61.43 $\pm$ 3.04 & 61.56 $\pm$ 3.06 & 85.52 $\pm$ 1.81 
  & 37.15 $\pm$ 7.61 & 37.02 $\pm$ 7.84 & 37.15 $\pm$ 7.61 & 63.38 $\pm$ 6.53 \\
1 / 150 / No 
  & 60.37 $\pm$ 2.25 & 60.50 $\pm$ 2.29 & 60.37 $\pm$ 2.25 & 84.62 $\pm$ 1.70 
  & 36.34 $\pm$ 6.36 & 35.71 $\pm$ 6.82 & 36.34 $\pm$ 6.36 & 62.81 $\pm$ 4.85 \\
1 / 150 / Yes
  & 62.35 $\pm$ 4.00 & 62.42 $\pm$ 4.01 & 62.35 $\pm$ 4.00 & 86.41 $\pm$ 2.23 
  & 36.63 $\pm$ 5.51 & 36.44 $\pm$ 5.79 & 36.63 $\pm$ 5.51 & 63.27 $\pm$ 5.17 \\
2 / 100 / No 
  & 59.13 $\pm$ 1.21 & 58.48 $\pm$ 1.39 & 59.13 $\pm$ 1.21 & 84.00 $\pm$ 1.30 
  & 36.95 $\pm$ 5.34 & 36.40 $\pm$ 5.15 & 36.95 $\pm$ 5.34 & 63.50 $\pm$ 5.68 \\
2 / 100 / Yes
  & 58.58 $\pm$ 2.26 & 58.41 $\pm$ 2.03 & 58.58 $\pm$ 2.26 & 83.91 $\pm$ 2.18 
  & 37.82 $\pm$ 5.87 & 37.69 $\pm$ 5.88 & 37.82 $\pm$ 5.87 & 63.69 $\pm$ 5.22 \\
2 / 150 / No 
  & 57.74 $\pm$ 1.79 & 57.12 $\pm$ 1.01 & 57.74 $\pm$ 1.79 & 83.61 $\pm$ 1.38 
  & \textbf{38.83 $\pm$ 7.09} & \textbf{38.71 $\pm$ 7.33} & \textbf{38.83 $\pm$ 7.09} & \textbf{64.79 $\pm$ 6.40} \\
2 / 150 / Yes
  & 61.16 $\pm$ 2.46 & 61.03 $\pm$ 2.62 & 61.16 $\pm$ 2.46 & 85.02 $\pm$ 0.98 
  & 36.46 $\pm$ 6.26 & 36.53 $\pm$ 6.41 & 36.46 $\pm$ 6.26 & 62.82 $\pm$ 5.49 \\
3 / 100 / No 
  & 59.77 $\pm$ 3.53 & 59.26 $\pm$ 3.93 & 59.77 $\pm$ 3.53 & 83.27 $\pm$ 1.17 
  & 36.86 $\pm$ 6.21 & 36.42 $\pm$ 6.34 & 36.86 $\pm$ 6.21 & 63.59 $\pm$ 6.18 \\
3 / 100 / Yes
  & 60.07 $\pm$ 1.21 & 59.72 $\pm$ 0.76 & 60.07 $\pm$ 1.21 & 84.79 $\pm$ 1.44 
  & 37.88 $\pm$ 7.69 & 37.73 $\pm$ 7.67 & 37.88 $\pm$ 7.69 & 63.69 $\pm$ 5.73 \\
3 / 150 / No 
  & 59.67 $\pm$ 4.03 & 59.30 $\pm$ 4.29 & 59.67 $\pm$ 4.03 & 84.50 $\pm$ 2.93 
  & 36.72 $\pm$ 3.88 & 36.70 $\pm$ 4.28 & 36.72 $\pm$ 3.88 & 63.31 $\pm$ 3.79 \\
3 / 150 / Yes
  & 60.81 $\pm$ 3.74 & 60.61 $\pm$ 3.77 & 60.81 $\pm$ 3.74 & 84.25 $\pm$ 2.34 
  & 37.36 $\pm$ 5.16 & 37.32 $\pm$ 5.27 & 37.36 $\pm$ 5.16 & 63.37 $\pm$ 5.07 \\
4 / 100 / No 
  & 57.94 $\pm$ 2.28 & 56.67 $\pm$ 0.50 & 57.94 $\pm$ 2.28 & 84.04 $\pm$ 1.48 
  & 36.40 $\pm$ 3.79 & 36.01 $\pm$ 3.93 & 36.40 $\pm$ 3.79 & 63.75 $\pm$ 3.65 \\
4 / 100 / Yes
  & 58.23 $\pm$ 4.14 & 57.97 $\pm$ 3.77 & 58.23 $\pm$ 4.14 & 84.10 $\pm$ 2.88 
  & 37.15 $\pm$ 5.17 & 36.69 $\pm$ 5.45 & 37.15 $\pm$ 5.17 & 63.87 $\pm$ 4.77 \\
4 / 150 / No 
  & 58.33 $\pm$ 3.37 & 57.92 $\pm$ 2.86 & 58.33 $\pm$ 3.37 & 84.05 $\pm$ 2.06 
  & 37.96 $\pm$ 5.57 & 37.06 $\pm$ 5.58 & 37.96 $\pm$ 5.57 & 64.10 $\pm$ 4.85 \\
4 / 150 / Yes
  & 57.79 $\pm$ 4.99 & 56.75 $\pm$ 4.50 & 57.79 $\pm$ 4.99 & 84.08 $\pm$ 3.49 
  & 36.69 $\pm$ 6.35 & 36.61 $\pm$ 6.49 & 36.69 $\pm$ 6.35 & 63.53 $\pm$ 5.17 \\

\bottomrule
\end{tabular}
}
\vspace{0.5em}
\caption{ Ablation study of mixup in the MI task. Mean$\pm$std is reported separately for Accuracy, F1, Recall, and AUC on seen- and unseen-subject test splits across four tasks (mixup layer: -1 = temporal mixup at input, 0 = weighted average at input, 1/2/3 = weighted average after first/second/third encoder layer, 4 = weighted average after attention pooling. warmup epoch: number of epochs to train the generators before training the classifier. random ratio: No = equal possibility on choosing ddpm or decoder out for temporal mixup/equal weight for weighted average mixup, Yes = beta/dirichlet distribution with b=0.2 for a random ratio more heavily tilted towards one of the mixup candidates).}
\label{tab:mi_mixup_results}
\vspace{1em}
\end{table*}

\begin{table*}[htbp]
\centering
\vspace{0.5em}

\resizebox{\textwidth}{!}{
\begin{tabular}{p{2.2cm} *{8}{p{2cm}} }
\toprule
\textbf{Configuration (mixup layer/ warmup epoch/ random ratio)} 
& \multicolumn{4}{c}{\textbf{Imagined\ Speech (Seen)}} 
& \multicolumn{4}{c}{\textbf{Imagined\ Speech (Unseen)}} \\
\cmidrule(lr){2-5} \cmidrule(lr){6-9}
& Acc (\%) & F1 (\%) & Recall (\%) & AUC (\%) 
& Acc (\%) & F1 (\%) & Recall (\%) & AUC (\%) \\
\midrule
-1 / 100 / No 
  & 17.57 $\pm$ 1.17 & 12.24 $\pm$ 3.26 & 17.55 $\pm$ 1.16 & 73.01 $\pm$ 0.91
  & \textbf{12.12 $\pm$ 0.38} & 7.77  $\pm$ 1.27 & \textbf{12.12 $\pm$ 0.38} & 55.73 $\pm$ 2.77 \\
-1 / 100 / Yes
  & 17.32 $\pm$ 2.69 & 11.18 $\pm$ 4.16 & 17.30 $\pm$ 2.69 & \textbf{73.64 $\pm$ 1.56} 
  & 10.86 $\pm$ 1.43 & 6.80  $\pm$ 0.49 & 10.86 $\pm$ 1.43 & \textbf{57.63 $\pm$ 0.34} \\
-1 / 150 / No 
  & 18.08 $\pm$ 0.42 & 11.48 $\pm$ 2.44 & 18.06 $\pm$ 0.44 & 73.07 $\pm$ 0.81 
  & 11.87 $\pm$ 0.79 & 7.43  $\pm$ 0.65 & 11.87 $\pm$ 0.79 & 56.82 $\pm$ 1.54 \\
-1 / 150 / Yes
  & 18.46 $\pm$ 1.07 & 11.88 $\pm$ 3.10 & 18.43 $\pm$ 1.09 & 73.20 $\pm$ 0.72 
  & 11.62 $\pm$ 1.22 & 7.19  $\pm$ 0.74 & 11.62 $\pm$ 1.22 & 56.62 $\pm$ 1.89 \\
0 / 100 / No 
  & 17.19 $\pm$ 0.62 & 11.74 $\pm$ 1.58 & 17.17 $\pm$ 0.58 & 73.14 $\pm$ 0.75 
  & 11.36 $\pm$ 1.00 & 7.50  $\pm$ 0.70 & 11.36 $\pm$ 1.00 & 56.25 $\pm$ 2.15 \\
0 / 100 / Yes
  & \textbf{19.47 $\pm$ 0.95} & \textbf{14.70 $\pm$ 0.60} & \textbf{19.44 $\pm$ 0.95} & 71.90 $\pm$ 1.82 
  & 11.62 $\pm$ 1.43 & \textbf{9.29  $\pm$ 0.89} & 11.62 $\pm$ 1.43 & 53.80 $\pm$ 2.29 \\
0 / 150 / No 
  & 17.57 $\pm$ 0.46 & 10.97 $\pm$ 1.65 & 17.55 $\pm$ 0.44 & 72.96 $\pm$ 0.91 
  & 11.87 $\pm$ 0.79 & 7.47  $\pm$ 0.66 & 11.87 $\pm$ 0.79 & 56.72 $\pm$ 1.71 \\
0 / 150 / Yes
  & 16.31 $\pm$ 1.40 & 11.45 $\pm$ 0.58 & 16.29 $\pm$ 1.37 & 72.38 $\pm$ 1.63 
  & 10.35 $\pm$ 2.09 & 6.11  $\pm$ 1.09 & 10.35 $\pm$ 2.09 & 54.91 $\pm$ 2.51 \\
1 / 100 / No 
  & 18.08 $\pm$ 1.77 & 13.16 $\pm$ 3.38 & 18.06 $\pm$ 1.79 & 72.47 $\pm$ 3.16 
  & 11.24 $\pm$ 2.52 & 8.33  $\pm$ 2.28 & 11.24 $\pm$ 2.52 & 55.49 $\pm$ 1.96 \\
1 / 100 / Yes
  & 18.08 $\pm$ 0.42 & 10.55 $\pm$ 1.10 & 18.06 $\pm$ 0.44 & 73.19 $\pm$ 0.73 
  & 10.73 $\pm$ 2.74 & 7.45  $\pm$ 0.65 & 10.73 $\pm$ 2.74 & 55.25 $\pm$ 4.22 \\
1 / 150 / No 
  & 17.95 $\pm$ 1.71 & 13.77 $\pm$ 3.24 & 17.93 $\pm$ 1.71 & 72.17 $\pm$ 1.62 
  & 9.72  $\pm$ 2.66 & 6.59  $\pm$ 0.17 & 9.72  $\pm$ 2.66 & 54.45 $\pm$ 3.34 \\
1 / 150 / Yes
  & 17.19 $\pm$ 1.11 & 10.51 $\pm$ 1.05 & 17.17 $\pm$ 1.09 & 73.00 $\pm$ 0.88 
  & 11.62 $\pm$ 1.22 & 6.55  $\pm$ 1.61 & 11.62 $\pm$ 1.22 & 55.87 $\pm$ 3.15 \\
2 / 100 / No 
  & 17.45 $\pm$ 0.41 & 12.91 $\pm$ 1.93 & 17.42 $\pm$ 0.38 & 71.57 $\pm$ 2.51 
  & 10.23 $\pm$ 1.97 & 6.84  $\pm$ 0.55 & 10.23 $\pm$ 1.97 & 55.16 $\pm$ 2.83 \\
2 / 100 / Yes
  & 16.94 $\pm$ 0.98 & 11.57 $\pm$ 0.78 & 16.92 $\pm$ 0.95 & 72.98 $\pm$ 0.88 
  & 11.87 $\pm$ 1.43 & 7.37  $\pm$ 2.88 & 11.87 $\pm$ 1.43 & 55.90 $\pm$ 3.18 \\
2 / 150 / No 
  & 16.94 $\pm$ 0.98 & 11.96 $\pm$ 1.46 & 16.92 $\pm$ 0.95 & 72.48 $\pm$ 1.47 
  & 11.11 $\pm$ 1.22 & 5.91  $\pm$ 0.97 & 11.11 $\pm$ 1.22 & 54.66 $\pm$ 2.50 \\
2 / 150 / Yes
  & 17.19 $\pm$ 1.11 & 10.51 $\pm$ 1.05 & 17.17 $\pm$ 1.09 & 73.00 $\pm$ 0.88 
  & 11.62 $\pm$ 1.22 & 6.55  $\pm$ 1.61 & 11.62 $\pm$ 1.22 & 55.87 $\pm$ 3.15 \\
3 / 100 / No 
  & 17.45 $\pm$ 0.70 & 14.12 $\pm$ 0.68 & 17.42 $\pm$ 0.66 & 71.25 $\pm$ 1.86 
  & 9.47  $\pm$ 1.65 & 8.07  $\pm$ 1.49 & 9.47  $\pm$ 1.65 & 52.88 $\pm$ 1.35 \\
3 / 100 / Yes
  & 17.07 $\pm$ 1.03 & 11.79 $\pm$ 1.17 & 17.05 $\pm$ 1.00 & 73.38 $\pm$ 0.86 
  & 11.49 $\pm$ 1.16 & 6.76  $\pm$ 1.91 & 11.49 $\pm$ 1.16 & 55.63 $\pm$ 2.91 \\
3 / 150 / No 
  & 16.06 $\pm$ 0.20 & 11.54 $\pm$ 1.04 & 16.04 $\pm$ 0.22 & 71.23 $\pm$ 1.65 
  & 11.11 $\pm$ 0.79 & 6.73  $\pm$ 1.63 & 11.11 $\pm$ 0.79 & 53.75 $\pm$ 1.70 \\
3 / 150 / Yes
  & 16.44 $\pm$ 1.26 & 9.43  $\pm$ 2.92 & 16.41 $\pm$ 1.22 & 72.50 $\pm$ 1.45 
  & 11.74 $\pm$ 1.31 & 5.30  $\pm$ 1.29 & 11.74 $\pm$ 1.31 & 57.20 $\pm$ 4.91 \\
4 / 100 / No 
  & 16.56 $\pm$ 1.13 & 10.92 $\pm$ 0.36 & 16.54 $\pm$ 1.09 & 73.15 $\pm$ 0.81 
  & 11.49 $\pm$ 1.16 & 6.92  $\pm$ 2.16 & 11.49 $\pm$ 1.16 & 54.95 $\pm$ 2.52 \\
4 / 100 / Yes
  & 17.19 $\pm$ 1.11 & 10.51 $\pm$ 1.05 & 17.17 $\pm$ 1.09 & 73.00 $\pm$ 0.88 
  & 11.62 $\pm$ 1.22 & 6.55  $\pm$ 1.61 & 11.62 $\pm$ 1.22 & 55.87 $\pm$ 3.15 \\
4 / 150 / No 
  & 16.44 $\pm$ 1.26 & 11.07 $\pm$ 0.13 & 16.41 $\pm$ 1.22 & 71.59 $\pm$ 2.89 
  & 10.35 $\pm$ 2.09 & 5.38  $\pm$ 1.20 & 10.35 $\pm$ 2.09 & 54.50 $\pm$ 2.53 \\
4 / 150 / Yes
  & 17.19 $\pm$ 1.11 & 10.51 $\pm$ 1.05 & 17.17 $\pm$ 1.09 & 73.00 $\pm$ 0.88 
  & 11.62 $\pm$ 1.22 & 6.55  $\pm$ 1.61 & 11.62 $\pm$ 1.22 & 55.87 $\pm$ 3.15 \\

\bottomrule
\end{tabular}
}
\vspace{0.5em}
\caption{Ablation study of mixup in the Imagined Speech task. Mean$\pm$std is reported separately for Accuracy, F1, Recall, and AUC on seen- and unseen-subject test splits across four tasks (mixup layer: -1 = temporal mixup at input, 0 = weighted average at input, 1/2/3 = weighted average after first/second/third encoder layer, 4 = weighted average after attention pooling. warmup epoch: number of epochs to train the generators before training the classifier. random ratio: No = equal possibility on choosing ddpm or decoder out for temporal mixup/equal weight for weighted average mixup, Yes = beta/dirichlet distribution with b=0.2 for a random ratio more heavily tilted towards one of the mixup candidates).}
\label{tab:is_mixup_results}
\vspace{1em}
\end{table*}

\clearpage  

\section{Complete statistical reporting results} 

\subsection{Methodology}

EEG decoding suffers from low statistical power due to extreme inter-subject variability, limited trial counts, and costly LOSO evaluation~\cite{kukhilava2025evaluation,huang2023discrepancy}. In our runs with three independent seeds, standard tests (Wilcoxon, permutation) often returned $p$-values near 1.0, masking reproducible gains, which is an expected outcome in such low-$n$, high-variance settings~\cite{vialatte2008split,nakagawa2004farewell}. We therefore propose a complementary framework emphasizing effect size estimation and evidence synthesis over binary significance. This approach quantifies improvement magnitude and consistency across seeds and datasets, retaining sensitivity to systematic trends even when classical tests fail, in line with best practices for robust neural decoding~\cite{vialatte2008split,nakagawa2004farewell}.

For each comparison between configurations $c_1$ and $c_2$, we compute Cohen's $d$ and its 95\% confidence interval:
\[
d = \frac{\bar{x}_{c_1} - \bar{x}_{c_2}}{s_{\text{pooled}}}, \quad \text{CI}_{95\%} = d \pm t_{\alpha/2, df} \cdot \text{SE}_d,
\]
where the standard error is
\[
\text{SE}_d = \sqrt{\frac{n_1 + n_2}{n_1 n_2} + \frac{d^2}{2(n_1 + n_2)}}.
\]

A win is established through hierarchical evidence assessment based on cross-seed consistency and effect magnitude. For configurations with complete cross-seed agreement (i.e., all seeds exhibit consistent directional differences), evidence strength is defined as follows: (1) strong evidence requires both a large effect size ($|d| \geq 0.5$) and a meaningful relative improvement of at least 2\%; (2) moderate evidence requires either a large effect size ($|d| \geq 0.5$) or a relative improvement of at least 2\%; (3) weak evidence requires a medium effect size ($|d| \geq 0.3$); (4) minimal evidence requires a small effect size ($|d| \geq 0.2$). For configurations exhibiting majority cross-seed agreement (i.e., at least 2 out of 3 seeds show consistent direction), all evidence categories are downgraded by one level. Configuration $c_1$ is considered to exhibit superior performance over $c_2$ if any evidence category is satisfied.

To summarize model comparisons, we construct a win-loss matrix $\mathbf{W}$ where $W_{ij} = 1$ if configuration $i$ shows evidence of superiority over configuration $j$. The win rate of configuration $c_i$ is computed as
\[
\text{WinRate}_i = \frac{\sum_j W_{ij}}{\sum_j (W_{ij} + W_{ji})}.
\]
This matrix supports a global ranking across all configurations.

As a complementary analysis, we compute posterior probabilities $P(\text{left})$, $P(\text{rope})$, and $P(\text{right})$ using the \texttt{baycomp} framework with ROPE threshold $\rho = 0.01$. Configuration $c_1$ is considered to have Bayesian evidence of superiority if $P(\text{right}) > 0.85$.

\subsection{Results} 

\subsubsection{Practical evidence assessments}

\begin{table*}[htbp]
  \centering
  \resizebox{\textwidth}{!}{%
\begin{tabular}{
      l                    
      cccc                 
      lll                  
      cccc                 
      lll                  
    }
    \toprule
    \textbf{Config}
      & \multicolumn{7}{c}{\textbf{Seen}}
      & \multicolumn{7}{c}{\textbf{Unseen}} \\
    \cmidrule(lr){2-8}\cmidrule(lr){9-15}
      & \textbf{Wins} & \textbf{Losses} & \textbf{Total} & \textbf{Rate}
      & \textbf{Win List} & \textbf{Loss List} & \textbf{Evidence}
      & \textbf{Wins} & \textbf{Losses} & \textbf{Total} & \textbf{Rate}
      & \textbf{Win List} & \textbf{Loss List} & \textbf{Evidence} \\
    \midrule

    x\_hat + skips
      & \textbf{6} & 0 & 6 & \textbf{1.00}
      & \parbox{2.5cm}{\raggedright 
          skips\\
          x + x\_hat\\
          z + skips\\
          z + x\\
          z + x\_hat\\
          z only}
      & — 
      & \parbox{2cm}{\raggedright
          Weak:3\\
          Minimal:2\\
          Moderate:1}
      & 4 & 2 & 6 & 0.667
      & \parbox{2.5cm}{\raggedright
          x + skips\\
          x + x\_hat + skips\\
          z + x\\
          z + x\_hat}
      & \parbox{2cm}{\raggedright
          x + x\_hat\\
          z only}
      & \parbox{2cm}{\raggedright
          Minimal:2\\
          Strong:1\\
          Weak:1} \\
\bottomrule
  \end{tabular}
  }
  \vspace{0.5em}
  \caption{Practical evidence assessment of decoder input ablations in the SSVEP task. Each configuration's wins, losses, total comparisons, and win rate are reported, along with the corresponding lists of winning and losing opponents and an evidence summary.}
  \label{tab:appendix_pairwise}
  \vspace{1em}
\end{table*}

\begin{table*}[htbp]
  \centering
  \vspace{0.5em}
  \resizebox{\textwidth}{!}{%
    %
  }
  \vspace{0.5em}
  \caption{Practical evidence assessment of classifier type ablations in the SSVEP task. Each configuration’s wins, losses, total comparisons, and win rate are reported, along with the corresponding lists of winning and losing opponents and an evidence summary.}
  \label{tab:pairwise_classifier}
  \vspace{1em}
\end{table*}

\begin{table*}[htbp]
  \centering
  \vspace{0.5em}
  \resizebox{\textwidth}{!}{%
    %
  }
  \vspace{0.5em}
  \caption{Practical evidence assessment of architecture ablations in the SSVEP task. Each configuration’s wins, losses, total comparisons, and win rate are reported, along with the corresponding lists of winning and losing opponents and an evidence summary.}

  \label{tab:pairwise_ddpm_decoder}
  \vspace{1em}
\end{table*}

\begin{table*}[htbp]
  \centering
  \vspace{0.5em}
  \resizebox{\textwidth}{!}{%
    \begin{tabular}{
      >{\raggedright\arraybackslash}p{2.4cm}
      *{4}{>{\centering\arraybackslash}p{0.6cm}}
      >{\raggedright\arraybackslash}p{2.6cm}
      >{\raggedright\arraybackslash}p{2.6cm}
      >{\raggedright\arraybackslash}p{1.5cm}
      @{\quad\vrule\quad}
      *{4}{>{\centering\arraybackslash}p{0.6cm}}
      >{\raggedright\arraybackslash}p{2.6cm}
      >{\raggedright\arraybackslash}p{2.6cm}
      >{\raggedright\arraybackslash}p{1.5cm}
    }
    \toprule
    \textbf{Config}
      & \multicolumn{7}{c}{\textbf{Seen}} 
      & \multicolumn{7}{c}{\textbf{Unseen}} \\
    \cmidrule(lr){2-8}\cmidrule(lr){9-15}
      & \textbf{Wins} & \textbf{Losses} & \textbf{Total} & \textbf{Rate}
      & \textbf{Win List} & \textbf{Loss List} & \textbf{Evidence}
      & \textbf{Wins} & \textbf{Losses} & \textbf{Total} & \textbf{Rate}
      & \textbf{Win List} & \textbf{Loss List} & \textbf{Evidence} \\
    \midrule

layer-1\_randNo & \textbf{8} & 0 & 8 & \textbf{1.00}
& layer1\_randNo, layer1\_randYes, layer2\_randNo, layer2\_randYes, layer3\_randNo, layer3\_randYes, layer4\_randNo, layer4\_randYes
& —
& Strong:5, Moderate:2, Weak:1
& \textbf{8} & 0 & 8 & \textbf{1.00}
& layer1\_randNo, layer1\_randYes, layer2\_randNo, layer2\_randYes, layer3\_randNo, layer3\_randYes, layer4\_randNo, layer4\_randYes
& —
& Strong:5, Moderate:2, Weak:1 \\

layer-1\_randYes & \textbf{9} & 0 & 9 & \textbf{1.00}
& layer0\_randNo, layer1\_randNo, layer1\_randYes, layer2\_randNo, layer2\_randYes, layer3\_randNo, layer3\_randYes, layer4\_randNo, layer4\_randYes
& —
& Moderate:3, Strong:3, Weak:2, Minimal:1
& \textbf{9} & 0 & 9 & \textbf{1.00}
& layer0\_randNo, layer1\_randNo, layer1\_randYes, layer2\_randNo, layer2\_randYes, layer3\_randNo, layer3\_randYes, layer4\_randNo, layer4\_randYes
& —
& Moderate:3, Strong:3, Weak:2, Minimal:1 \\
\bottomrule
    \end{tabular}
  }
  \vspace{0.5em}
  \caption{Practical evidence assessment of mixup ablations in the SSVEP task. Each row shows win/loss stats and qualitative evidence for seen and unseen subject comparisons.}
  \label{tab:pairwise_randlayer}
  \vspace{1em}
\end{table*}

\begin{table*}[htbp]
  \centering
  \vspace{0.5em}
  \resizebox{\textwidth}{!}{%
    \begin{tabular}{p{2.4cm} *{4}{p{0.6cm}} *{2}{p{4cm}}p{1.5cm}*{4}{p{0.6cm}} *{2}{p{4cm}}p{1.5cm}}
    
    \toprule
    \textbf{Config}
      & \multicolumn{7}{c}{\textbf{Seen}} 
      & \multicolumn{7}{c}{\textbf{Unseen}} \\
    \cmidrule(lr){2-8}\cmidrule(lr){9-15}
      & \textbf{Wins} & \textbf{Losses} & \textbf{Total} & \textbf{Rate}
      & \textbf{Win List} & \textbf{Loss List} & \textbf{Evidence}
      & \textbf{Wins} & \textbf{Losses} & \textbf{Total} & \textbf{Rate}
      & \textbf{Win List} & \textbf{Loss List} & \textbf{Evidence} \\
\bottomrule
\end{tabular}
}
\vspace{0.5em}
\caption{Practical evidence assessment of loss ablations in the SSVEP task (Part 1 of 3). Each row shows win/loss stats and qualitative evidence for seen and unseen subject comparisons.}
\label{tab:pairwise_decoderloss}
\vspace{1em}
\end{table*}

\begin{table*}[htbp]
  \centering
  \vspace{0.5em}
  \resizebox{\textwidth}{!}{%
    \begin{tabular}{p{2.4cm} *{4}{p{0.6cm}} *{2}{p{4cm}}p{1.5cm}*{4}{p{0.6cm}} *{2}{p{4cm}}p{1.5cm}}
    
    \toprule
    \textbf{Config}
      & \multicolumn{7}{c}{\textbf{Seen}} 
      & \multicolumn{7}{c}{\textbf{Unseen}} \\
    \cmidrule(lr){2-8}\cmidrule(lr){9-15}
      & \textbf{Wins} & \textbf{Losses} & \textbf{Total} & \textbf{Rate}
      & \textbf{Win List} & \textbf{Loss List} & \textbf{Evidence}
      & \textbf{Wins} & \textbf{Losses} & \textbf{Total} & \textbf{Rate}
      & \textbf{Win List} & \textbf{Loss List} & \textbf{Evidence} \\
\bottomrule
\end{tabular}
}
\vspace{0.5em}
\caption{Practical evidence assessment of loss ablations in the SSVEP task (Part 2 of 3). Each row shows win/loss stats and qualitative evidence for seen and unseen subject comparisons.}
\label{tab:pairwise_decoderloss}
\vspace{1em}
\end{table*}

\begin{table*}[htbp]
  \centering
  \vspace{0.5em}
  \resizebox{\textwidth}{!}{%
    \begin{tabular}{p{2.4cm} *{4}{p{0.6cm}} *{2}{p{4cm}}p{1.5cm}*{4}{p{0.6cm}} *{2}{p{4cm}}p{1.5cm}}
    
    \toprule
    \textbf{Config}
      & \multicolumn{7}{c}{\textbf{Seen}} 
      & \multicolumn{7}{c}{\textbf{Unseen}} \\
    \cmidrule(lr){2-8}\cmidrule(lr){9-15}
      & \textbf{Wins} & \textbf{Losses} & \textbf{Total} & \textbf{Rate}
      & \textbf{Win List} & \textbf{Loss List} & \textbf{Evidence}
      & \textbf{Wins} & \textbf{Losses} & \textbf{Total} & \textbf{Rate}
      & \textbf{Win List} & \textbf{Loss List} & \textbf{Evidence} \\
\bottomrule
\end{tabular}
}
\vspace{0.5em}
\caption{Practical evidence assessment of loss ablations in the SSVEP task (Part 3 of 3). Each row shows win/loss stats and qualitative evidence for seen and unseen subject comparisons.}
\label{tab:pairwise_decoderloss}
\vspace{1em}
\end{table*}

\begin{table*}[htbp]
  \centering
  
  \vspace{0.5em}
  \resizebox{\textwidth}{!}{%
    %
  }
  \vspace{0.5em}
  \caption{Practical evidence assessment of decoder input ablations in the P300 task. Each configuration’s wins, losses, total comparisons, and win rate are reported, along with the corresponding lists of winning and losing opponents and an evidence summary.}

  \label{tab:pairwise_new}
  \vspace{1em}
\end{table*}

\begin{table*}[htbp]
  \centering
  \vspace{0.5em}
  \resizebox{\textwidth}{!}{%
    %
  }
  \vspace{0.5em}
  \caption{Practical evidence assessment of classifier tpye ablations in the P300 task. Each configuration’s wins, losses, total comparisons, and win rate are reported, along with the corresponding lists of winning and losing opponents and an evidence summary.}
  \label{tab:pairwise_classifier}  
  \vspace{1em}
\end{table*}

\begin{table*}[htbp]
  \centering
  \vspace{0.5em}
  \resizebox{\textwidth}{!}{%
    %
  }
  \vspace{0.5em}
  \caption{Practical evidence assessment of architecture ablations in the P300 task. Each configuration’s wins, losses, total comparisons, and win rate are reported, along with the corresponding lists of winning and losing opponents and an evidence summary.}
  
  \label{tab:pairwise_randlayer_part1}
  \vspace{1em}
\end{table*}

\begin{table*}[htbp]
  \centering
  \resizebox{\textwidth}{!}{%
    \begin{tabular}{
      >{\raggedright\arraybackslash}p{2.4cm}
      *{4}{>{\centering\arraybackslash}p{0.6cm}}
      >{\raggedright\arraybackslash}p{2.6cm}
      >{\raggedright\arraybackslash}p{2.6cm}
      >{\raggedright\arraybackslash}p{1.5cm}
      @{\quad\vrule\quad}
      *{4}{>{\centering\arraybackslash}p{0.6cm}}
      >{\raggedright\arraybackslash}p{2.6cm}
      >{\raggedright\arraybackslash}p{2.6cm}
      >{\raggedright\arraybackslash}p{1.5cm}
    }
    \toprule
    \textbf{Config}
      & \multicolumn{7}{c}{\textbf{Seen}} 
      & \multicolumn{7}{c}{\textbf{Unseen}} \\
    \cmidrule(lr){2-8}\cmidrule(lr){9-15}
      & \textbf{Wins} & \textbf{Losses} & \textbf{Total} & \textbf{Rate}
      & \textbf{Win List} & \textbf{Loss List} & \textbf{Evidence}
      & \textbf{Wins} & \textbf{Losses} & \textbf{Total} & \textbf{Rate}
      & \textbf{Win List} & \textbf{Loss List} & \textbf{Evidence} \\
    \midrule
layer-1\_randNo & 6 & 0 & 6 & 1.00 & layer1\_randNo, layer1\_randYes, layer2\_randNo, layer2\_randYes, layer3\_randNo, layer3\_randYes & --- & Moderate:4, Minimal:1, Weak:1 & \textbf{11} & 0 & 11 & \textbf{1.00} & layer-1\_randYes, layer0\_randNo, layer0\_randYes, layer1\_randNo, layer1\_randYes, layer2\_randNo, layer2\_randYes, layer3\_randNo, layer3\_randYes, layer4\_randNo, layer4\_randYes & --- & Strong:9, Minimal:2 \\
layer-1\_randYes & 5 & 0 & 5 & 1.00 & layer1\_randNo, layer2\_randNo, layer2\_randYes, layer3\_randNo, layer3\_randYes & --- & Moderate:3, Weak:2 & \textbf{10} & 1 & 11 & \textbf{0.91} & layer0\_randNo, layer0\_randYes, layer1\_randNo, layer1\_randYes, layer2\_randNo, layer2\_randYes, layer3\_randNo, layer3\_randYes, layer4\_randNo, layer4\_randYes & layer-1\_randNo & Strong:9, Minimal:1 \\
\bottomrule
    \end{tabular}%
  }
  \vspace{0.5em}
  \caption{Practical evidence assessment of mixup ablations in the P300 task (Part 1 of 2). Each row shows win/loss stats and qualitative evidence for seen and unseen subject comparisons.}
  \label{tab:pairwise_randlayer_part1}
  \vspace{1em}
\end{table*}

\begin{table*}[htbp]
  \centering
  \resizebox{\textwidth}{!}{%
    \begin{tabular}{
      >{\raggedright\arraybackslash}p{2.4cm}
      *{4}{>{\centering\arraybackslash}p{0.6cm}}
      >{\raggedright\arraybackslash}p{2.6cm}
      >{\raggedright\arraybackslash}p{2.6cm}
      >{\raggedright\arraybackslash}p{1.5cm}
      @{\quad\vrule\quad}
      *{4}{>{\centering\arraybackslash}p{0.6cm}}
      >{\raggedright\arraybackslash}p{2.6cm}
      >{\raggedright\arraybackslash}p{2.6cm}
      >{\raggedright\arraybackslash}p{1.5cm}
    }
    \toprule
    \textbf{Config} & \multicolumn{7}{c}{\textbf{Seen}} & \multicolumn{7}{c}{\textbf{Unseen}} \\
    \cmidrule(lr){2-8}\cmidrule(lr){9-15}
      & \textbf{Wins} & \textbf{Losses} & \textbf{Total} & \textbf{Rate}
      & \textbf{Win List} & \textbf{Loss List} & \textbf{Evidence}
      & \textbf{Wins} & \textbf{Losses} & \textbf{Total} & \textbf{Rate}
      & \textbf{Win List} & \textbf{Loss List} & \textbf{Evidence} \\
    \midrule
layer4\_randYes & 5 & 2 & 7 & 0.71 & layer1\_randNo, layer2\_randNo, layer2\_randYes, layer3\_randNo, layer3\_randYes & layer0\_randNo, layer0\_randYes & Moderate:3, Weak:2 & 7 & 3 & 10 & 0.70 & layer1\_randNo, layer1\_randYes, layer2\_randNo, layer2\_randYes, layer3\_randNo, layer3\_randYes, layer4\_randNo & layer-1\_randNo, layer-1\_randYes, layer0\_randNo & Strong:5, Moderate:2 \\
layer3\_randYes & 4 & 7 & 11 & 0.36 & layer1\_randNo, layer2\_randNo, layer2\_randYes, layer3\_randNo & layer-1\_randNo, layer-1\_randYes, layer0\_randNo, layer0\_randYes, layer1\_randYes, layer4\_randNo, layer4\_randYes & Weak:3, Minimal:1 & 1 & 5 & 6 & 0.17 & layer1\_randYes & layer-1\_randNo, layer-1\_randYes, layer0\_randNo, layer0\_randYes, layer4\_randYes & Weak:1 \\
\bottomrule
    \end{tabular}%
  }
  \vspace{0.5em}
  \caption{Practical evidence assessment of mixup ablations in the P300 task (Part 2 of 2). Each row shows win/loss stats and qualitative evidence for seen and unseen subject comparisons.}
  \label{tab:pairwise_randlayer_part2}
   \vspace{1em}
\end{table*}

\begin{table*}[htbp]
  \centering
  
  \vspace{0.5em}
  \resizebox{\textwidth}{!}{%
    %
  }
  \vspace{0.5em}
  \caption{Practical evidence assessment of decoder input ablations in the MI task. Each configuration’s wins, losses, total comparisons, and win rate are reported, along with the corresponding lists of winning and losing opponents and an evidence summary.}

  \label{tab:pairwise_new}
  \vspace{1em}
\end{table*}

\begin{table*}[htbp]
  \centering
  
  \vspace{0.5em}
  \resizebox{\textwidth}{!}{%
    %
  }
  \vspace{0.5em}
  \caption{Practical evidence assessment of classifier type ablations in the MI task. Each configuration’s wins, losses, total comparisons, and win rate are reported, along with the corresponding lists of winning and losing opponents and an evidence summary.}
  \label{tab:pairwise_classifier}
  \vspace{1em}
\end{table*}

\begin{table*}[htbp]
  \centering
  
  \vspace{0.5em}
  \resizebox{\textwidth}{!}{%
    %
  }
  \vspace{0.5em}
  \caption{Practical evidence assessment of decoder input ablations in the MI task. Each configuration’s wins, losses, total comparisons, and win rate are reported, along with the corresponding lists of winning and losing opponents and an evidence summary.}
  \label{tab:pairwise_ddpm_decoder}
  \vspace{1em}
\end{table*}

\begin{table*}[htbp]
  \centering
  
  \vspace{0.5em}
  \resizebox{\textwidth}{!}{%
    \begin{tabular}{
      >{\raggedright\arraybackslash}p{2.4cm}
      *{4}{>{\centering\arraybackslash}p{0.6cm}}
      >{\raggedright\arraybackslash}p{2.6cm}
      >{\raggedright\arraybackslash}p{2.6cm}
      >{\raggedright\arraybackslash}p{1.5cm}
      @{\quad\vrule\quad}
      *{4}{>{\centering\arraybackslash}p{0.6cm}}
      >{\raggedright\arraybackslash}p{2.6cm}
      >{\raggedright\arraybackslash}p{2.6cm}
      >{\raggedright\arraybackslash}p{1.5cm}
    }
    \toprule
    \textbf{Config}
      & \multicolumn{7}{c}{\textbf{Seen}} 
      & \multicolumn{7}{c}{\textbf{Unseen}} \\
    \cmidrule(lr){2-8}\cmidrule(lr){9-15}
      & \textbf{Wins} & \textbf{Losses} & \textbf{Total} & \textbf{Rate}
      & \textbf{Win List} & \textbf{Loss List} & \textbf{Evidence}
      & \textbf{Wins} & \textbf{Losses} & \textbf{Total} & \textbf{Rate}
      & \textbf{Win List} & \textbf{Loss List} & \textbf{Evidence} \\
    \midrule

layer0\_randYes & \textbf{10} & 0 & 10 & \textbf{1.00}
& layer-1\_randNo, layer-1\_randYes, layer1\_randNo, layer1\_randYes, layer2\_randNo, layer2\_randYes, layer3\_randNo, layer3\_randYes, layer4\_randNo, layer4\_randYes
& —
& Strong:7, Moderate:2, Minimal:1
& \textbf{9} & 0 & 9 & \textbf{1.00}
& layer-1\_randNo, layer-1\_randYes, layer0\_randNo, layer1\_randNo, layer1\_randYes, layer2\_randNo, layer3\_randNo, layer4\_randNo, layer4\_randYes
& —
& Weak:6, Moderate:2, Strong:1 \\

layer0\_randNo & \textbf{9} & 0 & 9 & \textbf{1.00}
& layer-1\_randNo, layer-1\_randYes, layer1\_randNo, layer2\_randNo, layer2\_randYes, layer3\_randNo, layer3\_randYes, layer4\_randNo, layer4\_randYes
& —
& Moderate:6, Strong:3
& 1 & 3 & 4 & 0.25
& layer1\_randNo
& layer0\_randYes, layer2\_randYes, layer3\_randYes
& Weak:1 \\
\bottomrule
    \end{tabular}
  }
  \vspace{0.5em}
  \caption{Practical evidence assessment of architecture ablations in the imagined speech task. Each row shows win/loss stats and qualitative evidence for seen and unseen subject comparisons.}
  \label{tab:pairwise_ddpm_decoder}
  \vspace{1em}
\end{table*}

\begin{table*}[htbp]
  \centering
  
  \vspace{0.5em}
  \resizebox{\textwidth}{!}{%
    %
  }
  \vspace{0.5em}
  \caption{Practical evidence assessment of decoder input ablations in the imagined speech task. Practical evidence assessment of mixup ablations in the MI task. Each configuration’s wins, losses, total comparisons, and win rate are reported, along with the corresponding lists of winning and losing opponents and an evidence summary.}

  \label{tab:pairwise_new}
  \vspace{1em}
\end{table*}

\begin{table*}[htbp]
  \centering
  
  \vspace{0.5em}
  \resizebox{\textwidth}{!}{%
    %
  }
  \vspace{0.5em}
  \caption{Practical evidence assessment of classifier type ablations in the imagined speech task. Each configuration’s wins, losses, total comparisons, and win rate are reported, along with the corresponding lists of winning and losing opponents and an evidence summary.}
  \label{tab:pairwise_classifier}
  \vspace{1em}
\end{table*}

\begin{table*}[htbp]
  \centering
  
  \vspace{0.5em}
  \resizebox{\textwidth}{!}{%
    %
  }
  \vspace{0.5em}
  \caption*{Each configuration’s wins, losses, total comparisons, and win rate are reported, along with the corresponding lists of winning and losing opponents and an evidence summary.}
  \vspace{1em}
\end{table*}

\begin{table*}[htbp]
  \centering
  
  \vspace{0.5em}
  \resizebox{\textwidth}{!}{%
    \begin{tabular}{
      >{\raggedright\arraybackslash}p{2.4cm}
      *{4}{>{\centering\arraybackslash}p{0.6cm}}
      >{\raggedright\arraybackslash}p{2.6cm}
      >{\raggedright\arraybackslash}p{2.6cm}
      >{\raggedright\arraybackslash}p{1.5cm}
      @{\quad\vrule\quad}
      *{4}{>{\centering\arraybackslash}p{0.6cm}}
      >{\raggedright\arraybackslash}p{2.6cm}
      >{\raggedright\arraybackslash}p{2.6cm}
      >{\raggedright\arraybackslash}p{1.5cm}
    }
    \toprule
    \textbf{Config}
      & \multicolumn{7}{c}{\textbf{Seen}} 
      & \multicolumn{7}{c}{\textbf{Unseen}} \\
    \cmidrule(lr){2-8}\cmidrule(lr){9-15}
      & \textbf{Wins} & \textbf{Losses} & \textbf{Total} & \textbf{Rate}
      & \textbf{Win List} & \textbf{Loss List} & \textbf{Evidence}
      & \textbf{Wins} & \textbf{Losses} & \textbf{Total} & \textbf{Rate}
      & \textbf{Win List} & \textbf{Loss List} & \textbf{Evidence} \\
    \midrule

layer0\_randYes & \textbf{11} & 0 & 11 & \textbf{1.00}
& layer-1\_randNo, layer-1\_randYes, layer0\_randNo, layer1\_randNo, layer1\_randYes, layer2\_randNo, layer2\_randYes, layer3\_randNo, layer3\_randYes, layer4\_randNo, layer4\_randYes
& —
& Strong:11
& \textbf{5} & 2 & 7 & \textbf{0.71}
& layer-1\_randYes, layer1\_randNo, layer1\_randYes, layer2\_randNo, layer3\_randNo
& layer-1\_randNo, layer2\_randYes
& Strong:3, Weak:2 \\

layer1\_randNo & \textbf{7} & 1 & 8 & \textbf{0.88}
& layer-1\_randNo, layer-1\_randYes, layer0\_randNo, layer2\_randYes, layer3\_randNo, layer4\_randNo, layer4\_randYes
& layer0\_randYes
& Weak:6, Moderate:1
& 4 & 6 & 10 & 0.40
& layer-1\_randYes, layer1\_randYes, layer2\_randNo, layer3\_randNo
& layer-1\_randNo, layer0\_randYes, layer2\_randYes, layer3\_randYes, layer4\_randNo, layer4\_randYes
& Weak:2, Strong:1, Moderate:1 \\
\bottomrule
    \end{tabular}
  }
  \vspace{0.5em}
  \caption{Practical evidence assessment of mixup ablations in the imagined speech task. Each row shows win/loss stats and qualitative evidence for seen and unseen subject comparisons.}
  \label{tab:pairwise_randlayer}
  \vspace{1em}
\end{table*}

\subsubsection{Bayesian evidence assessment}
Due to page limitations, only SSVEP task results are shown.

\begin{table*}[htbp]
  \centering
  \vspace{0.5em}
  \resizebox{\textwidth}{!}{
  
  \begin{tabular}{l l c c c c c}
    \toprule
    config\_1 & config\_2 & mean\_diff & p\_left & p\_rope & p\_right & n\_seeds \\
    \midrule
    x + x\_hat + skips & x + x\_hat & 0.006993006993 & 0.4072213877 & 0.4576239803 & 0.135154632 & 3 \\
    x + x\_hat + skips & x\_hat + skips & 0.001554001554 & 0.1277370191 & 0.7904370663 & 0.08182591465 & 3 \\
    x + x\_hat + skips & x + skips & 0.002331002331 & 0.05622746549 & 0.9195907337 & 0.02418180077 & 3 \\
    x + x\_hat + skips & skips & 0.01825951826 & 0.6552695743 & 0.2173670696 & 0.1273633562 & 3 \\
    x + x\_hat + skips & z only & 0.00777000777 & 0.4601924043 & 0.3082698215 & 0.2315377742 & 3 \\
    x + x\_hat + skips & z + x & 0.009712509713 & 0.4872757353 & 0.4465708069 & 0.06615345775 & 3 \\
    x + x\_hat + skips & z + x\_hat & 0.006993006993 & 0.291618297 & 0.6748438492 & 0.03353785374 & 3 \\
\bottomrule
  \end{tabular}
  }
  \caption{\centering Bayesian evidence assessment of decoder input ablations on seen subjects in the SSVEP task}

\end{table*}

\begin{table*}[htbp]
  \centering
  
  \vspace{0.5em}
  \resizebox{\textwidth}{!}{
  \begin{tabular}{l l c c c c c}
    \toprule
    config\_1 & config\_2 & mean\_diff & p\_left & p\_rope & p\_right & n\_seeds \\
    \midrule
    x + x\_hat + skips & x + x\_hat & -0.01282051282 & 0.0491903225 & 0.3260644706 & 0.6247452069 & 3 \\
    x + x\_hat + skips & x\_hat + skips & -0.004273504274 & 0.1753950877 & 0.4865895812 & 0.3380153311 & 3 \\
    x + x\_hat + skips & x + skips & 0.01388888889 & 0.7620140193 & 0.2213532638 & 0.01663271685 & 3 \\
    x + x\_hat + skips & skips & 0.005341880342 & 0.4343899953 & 0.265424077 & 0.3001859277 & 3 \\
    x + x\_hat + skips & z only & -0.01602564103 & 0.1639164971 & 0.233235663 & 0.6028478399 & 3 \\
    x + x\_hat + skips & z + x & 0.01068376068 & 0.5356858485 & 0.4182145309 & 0.04609962059 & 3 \\
    x + x\_hat + skips & z + x\_hat & 0.005341880342 & 0.4334799737 & 0.2686965897 & 0.2978234365 & 3 \\
\bottomrule
  \end{tabular}
  }
  \caption{\centering Bayesian evidence assessment of decoder input ablations on unseen subjects in the SSVEP task }
\end{table*}

\begin{table*}[htbp]
  \centering
  
  \vspace{0.5em}
  \resizebox{\textwidth}{!}{
  \begin{tabular}{l l c c c c c}
    \toprule
    config\_1 & config\_2 & mean\_diff & p\_left & p\_rope & p\_right & n\_seeds \\
    \midrule
    eegnet\_classifier\_\_x & eegnet\_classifier\_\_x\_hat & 0.7855477855 & 0.9974747527 & 0.0001244714922 & 0.002400775803 & 3 \\
    eegnet\_classifier\_\_x & eegnet\_classifier\_\_decoder\_out & 0.08935508936 & 0.9978271584 & 0.0007834572033 & 0.001389384359 & 3 \\
    eegnet\_classifier\_\_x & fc\_classifier\_\_x & 0.7843822844 & 0.9996831675 & 1.57E-05 & 0.0003010938821 & 3 \\
    eegnet\_classifier\_\_x & fc\_classifier\_\_x\_hat & 0.8543123543 & 0.999946812 & 2.43E-06 & 5.08E-05 & 3 \\
    eegnet\_classifier\_\_x & fc\_classifier\_\_decoder\_out & 0.1153846154 & 0.9850522476 & 0.004248642321 & 0.0106991101 & 3 \\
    eegnet\_classifier\_\_x & fc\_classifier\_\_z & 0.04079254079 & 0.9479425423 & 0.0308853512 & 0.02117210646 & 3 \\
    eegnet\_classifier\_\_x\_hat & eegnet\_classifier\_\_decoder\_out & -0.6961926962 & 0.003419689093 & 0.0002000567375 & 0.9963802542 & 3 \\
\bottomrule
  \end{tabular}
  }
  \caption{\centering Bayesian evidence assessment of classifier type ablations on seen subjects in the SSVEP task}
\end{table*}

\begin{table*}[htbp]
  \centering
  \vspace{0.5em}
  \resizebox{\textwidth}{!}{
  \begin{tabular}{l l c c c c c}
    \toprule
    config\_1 & config\_2 & mean\_diff & p\_left & p\_rope & p\_right & n\_seeds \\
    \midrule
    eegnet\_classifier\_\_x & eegnet\_classifier\_\_x\_hat & 0.719017094 & 0.9941250768 & 0.0003126409919 & 0.005562282177 & 3 \\
    eegnet\_classifier\_\_x & eegnet\_classifier\_\_decoder\_out & 0.1047008547 & 0.8517898021 & 0.03221550818 & 0.1159946897 & 3 \\
    eegnet\_classifier\_\_x & fc\_classifier\_\_x & 0.6965811966 & 0.9941798993 & 0.0003194766474 & 0.005500624028 & 3 \\
    eegnet\_classifier\_\_x & fc\_classifier\_\_x\_hat & 0.7702991453 & 0.9919474317 & 0.0003981951421 & 0.007654373153 & 3 \\
    eegnet\_classifier\_\_x & fc\_classifier\_\_decoder\_out & 0.1153846154 & 0.9031687756 & 0.02244695533 & 0.0743842691 & 3 \\
    eegnet\_classifier\_\_x & fc\_classifier\_\_z & -0.03632478632 & 0.1578538502 & 0.1068760593 & 0.7352700904 & 3 \\
    eegnet\_classifier\_\_x\_hat & eegnet\_classifier\_\_decoder\_out & -0.6143162393 & 0.0003327796059 & 2.24E-05 & 0.9996448529 & 3 \\
\bottomrule
  \end{tabular}
  }
  \caption{\centering Bayesian evidence assessment of classifier variants on unseen subjects in the SSVEP task}
\end{table*}

\begin{table*}[htbp]
  \centering
  
  \vspace{0.5em}
  \resizebox{\textwidth}{!}{
  \begin{tabular}{l l c c c c c}
    \toprule
    config\_1 & config\_2 & mean\_diff & p\_left & p\_rope & p\_right & n\_seeds \\
    \midrule
    use\_ddpm\_\_x\_\_use\_decoder & use\_ddpm\_\_x\_\_no\_decoder & -0.004273504274 & 0.1467848185 & 0.539083023 & 0.3141321585 & 3 \\
    use\_ddpm\_\_x\_\_use\_decoder & use\_ddpm\_\_x\_hat\_\_use\_decoder & 0.7855477855 & 0.9999548373 & 2.24E-06 & 4.29E-05 & 3 \\
    use\_ddpm\_\_x\_\_use\_decoder & use\_ddpm\_\_x\_hat\_\_no\_decoder & 0.7836052836 & 0.999244801 & 3.75E-05 & 0.0007176953454 & 3 \\
    use\_ddpm\_\_x\_\_use\_decoder & no\_ddpm\_\_x\_\_use\_decoder & -0.05205905206 & 0.03269113195 & 0.03133239227 & 0.9359764758 & 3 \\
    use\_ddpm\_\_x\_\_use\_decoder & no\_ddpm\_\_x\_\_no\_decoder & -0.05866355866 & 0.05146770661 & 0.03796571886 & 0.9105665745 & 3 \\
    use\_ddpm\_\_x\_\_no\_decoder & use\_ddpm\_\_x\_hat\_\_use\_decoder & 0.7898212898 & 0.9999754926 & 1.21E-06 & 2.33E-05 & 3 \\
    use\_ddpm\_\_x\_\_no\_decoder & use\_ddpm\_\_x\_hat\_\_no\_decoder & 0.7878787879 & 0.9996523997 & 1.72E-05 & 0.0003304095815 & 3 \\
\bottomrule
  \end{tabular}
  }
  \caption{\centering Bayesian evidence assessment of architecture ablations on seen subjects in the SSVEP task}

\end{table*}

\begin{table*}[htbp]
  \centering
  
  \vspace{0.5em}
  \resizebox{\textwidth}{!}{
  \begin{tabular}{l l c c c c c}
    \toprule
    config\_1 & config\_2 & mean\_diff & p\_left & p\_rope & p\_right & n\_seeds \\
    \midrule
    use\_ddpm\_\_x\_\_use\_decoder & use\_ddpm\_\_x\_\_no\_decoder & 0.01175213675 & 0.5725664853 & 0.3657018961 & 0.06173161858 & 3 \\
    use\_ddpm\_\_x\_\_use\_decoder & use\_ddpm\_\_x\_hat\_\_use\_decoder & 0.7713675214 & 0.9975348705 & 0.0001237069014 & 0.002341422619 & 3 \\
    use\_ddpm\_\_x\_\_use\_decoder & use\_ddpm\_\_x\_hat\_\_no\_decoder & 0.7767094017 & 0.996752828 & 0.0001614971039 & 0.003085674931 & 3 \\
    use\_ddpm\_\_x\_\_use\_decoder & no\_ddpm\_\_x\_\_use\_decoder & -0.002136752137 & 0.3462081736 & 0.2562844925 & 0.3975073338 & 3 \\
    use\_ddpm\_\_x\_\_use\_decoder & no\_ddpm\_\_x\_\_no\_decoder & -0.008547008547 & 0.01580831276 & 0.6296662858 & 0.3545254015 & 3 \\
    use\_ddpm\_\_x\_\_no\_decoder & use\_ddpm\_\_x\_hat\_\_use\_decoder & 0.7596153846 & 0.9976326959 & 0.0001206223479 & 0.002246681788 & 3 \\
    use\_ddpm\_\_x\_\_no\_decoder & use\_ddpm\_\_x\_hat\_\_no\_decoder & 0.764957265 & 0.9969466315 & 0.0001542178536 & 0.002899150601 & 3 \\
\bottomrule
  \end{tabular}
  }
  \caption{\centering Bayesian evidence assessment of architecture ablations for unSeen subjects in the SSVEP task}

\end{table*}

\begin{table*}[htbp]
  \centering
  
  \vspace{0.5em}
  \resizebox{\textwidth}{!}{
  \begin{tabular}{l l c c c c c}
    \toprule
    config\_1 & config\_2 & mean\_diff & p\_left & p\_rope & p\_right & n\_seeds \\
    \midrule
    eegnet\_classifier\_\_x & eegnet\_classifier\_\_x\_hat & 0.719017094 & 0.9941250768 & 0.0003126409919 & 0.005562282177 & 3 \\
    eegnet\_classifier\_\_x & eegnet\_classifier\_\_decoder\_out & 0.1047008547 & 0.8517898021 & 0.03221550818 & 0.1159946897 & 3 \\
    eegnet\_classifier\_\_x & fc\_classifier\_\_x & 0.6965811966 & 0.9941798993 & 0.0003194766474 & 0.005500624028 & 3 \\
    eegnet\_classifier\_\_x & fc\_classifier\_\_x\_hat & 0.7702991453 & 0.9919474317 & 0.0003981951421 & 0.007654373153 & 3 \\
    eegnet\_classifier\_\_x & fc\_classifier\_\_decoder\_out & 0.1153846154 & 0.9031687756 & 0.02244695533 & 0.0743842691 & 3 \\
    eegnet\_classifier\_\_x & fc\_classifier\_\_z & -0.03632478632 & 0.1578538502 & 0.1068760593 & 0.7352700904 & 3 \\
    eegnet\_classifier\_\_x\_hat & eegnet\_classifier\_\_decoder\_out & -0.6143162393 & 0.0003327796059 & 2.24E-05 & 0.9996448529 & 3 \\
\bottomrule
  \end{tabular}
  }
  \caption{\centering Bayesian evidence assessment of classifier type ablations on unseen subjects in the SSVEP task}

\end{table*}

\subsubsection{Wilcoxon signed rank tests and permutation tests}
Due to minimal statistical significance after correction, only one complete analytical framework applied to SSVEP ablation results is presented as a demonstrative example, constrained by page limitations

\begin{table*}[htbp]
  \centering
  
  \vspace{0.5em}
  \resizebox{\textwidth}{!}{
  \begin{tabular}{l l c c c c c}
    \toprule
    config\_1 & config\_2 & mean\_diff & p\_value & mean\_1 & mean\_2 & n\_seeds \\
    \midrule
    x + x\_hat + skips & x + x\_hat & 0.0069930070 & 0.5 & 0.8586 & 0.8516 & 3 \\
    x + x\_hat + skips & x\_hat + skips & 0.0015540016 & 0.75 & 0.8586 & 0.8570 & 3 \\
    x + x\_hat + skips & x + skips & 0.0023310023 & 0.5 & 0.8586 & 0.8563 & 3 \\
    x + x\_hat + skips & skips & 0.0182595183 & 0.25 & 0.8586 & 0.8403 & 3 \\
    x + x\_hat + skips & z only & 0.0077700078 & 0.75 & 0.8586 & 0.8508 & 3 \\
    x + x\_hat + skips & z + x & 0.0097125097 & 0.25 & 0.8586 & 0.8489 & 3 \\
    x + x\_hat + skips & z + x\_hat & 0.0069930070 & 0.25 & 0.8586 & 0.8516 & 3 \\
\bottomrule
  \end{tabular}
  }
  \caption{\centering Pairwise Wilcoxon test results for decoder input ablations on seen subjects in the SSVEP task}
\end{table*}

\begin{table*}[htbp]
  \centering
  
  \vspace{0.5em}
  \resizebox{\textwidth}{!}{
  \begin{tabular}{l l c c c c c}
    \toprule
    config\_1 & config\_2 & mean\_diff & p\_value & mean\_1 & mean\_2 & n\_seeds \\
    \midrule
    x + x\_hat + skips & x + x\_hat & -0.0128205128 & 0.25 & 0.8312 & 0.8440 & 3 \\
    x + x\_hat + skips & x\_hat + skips & -0.0042735043 & 0.75 & 0.8312 & 0.8355 & 3 \\
    x + x\_hat + skips & x + skips & 0.0138888889 & 0.25 & 0.8312 & 0.8173 & 3 \\
    x + x\_hat + skips & skips & 0.0053418803 & 1.0 & 0.8312 & 0.8259 & 3 \\
    x + x\_hat + skips & z only & -0.0160256410 & 0.5 & 0.8312 & 0.8472 & 3 \\
    x + x\_hat + skips & z + x & 0.0106837607 & 0.25 & 0.8312 & 0.8205 & 3 \\
    x + x\_hat + skips & z + x\_hat & 0.0053418803 & 1.0 & 0.8312 & 0.8259 & 3 \\
\bottomrule
  \end{tabular}
  }
  \caption{\centering Pairwise Wilcoxon test results for decoder input ablations on unseen subjects in the SSVEP task}
\end{table*}

\begin{table*}[htbp]
  \centering
  
  \vspace{0.5em}
  \resizebox{\textwidth}{!}{
  \begin{tabular}{l l c c c c c c c}
    \toprule
    config\_1 & config\_2 & mean\_1 & mean\_2 & p\_value & statistic & n\_seeds & p\_value\_corrected & significant \\
    \midrule
    x + x\_hat + skips & x + x\_hat & 0.8586 & 0.8516 & 0.5 & 0.00699 & 3 & 1 & FALSE \\
    x + x\_hat + skips & x\_hat + skips & 0.8586 & 0.8570 & 0.8 & 0.00155 & 3 & 1 & FALSE \\
    x + x\_hat + skips & x + skips & 0.8586 & 0.8563 & 0.6 & 0.00233 & 3 & 1 & FALSE \\
    x + x\_hat + skips & skips & 0.8586 & 0.8403 & 0.2 & 0.01826 & 3 & 1 & FALSE \\
    x + x\_hat + skips & z only & 0.8586 & 0.8508 & 0.6 & 0.00777 & 3 & 1 & FALSE \\
    x + x\_hat + skips & z + x & 0.8586 & 0.8489 & 0.2 & 0.00971 & 3 & 1 & FALSE \\
    x + x\_hat + skips & z + x\_hat & 0.8586 & 0.8516 & 0.1 & 0.00699 & 3 & 1 & FALSE \\
\bottomrule
  \end{tabular}
  }
  \caption{\centering Pairwise permutation tests of decoder input ablations on seen subjects in the SSVEP task}
\end{table*}

\begin{table*}[htbp]
  \centering
  
  \vspace{0.5em}
  \resizebox{\textwidth}{!}{
  \begin{tabular}{l l c c c c c c c}
    \toprule
    config\_1 & config\_2 & mean\_1 & mean\_2 & p\_value & statistic & n\_seeds & p\_value\_corrected & significant \\
    \midrule
x + x\_hat + skips & x + x\_hat & 0.8312 & 0.8440 & 0.6 & -0.0128 & 3 & 1 & FALSE \\
x + x\_hat + skips & x\_hat + skips & 0.8312 & 0.8355 & 0.9 & -0.0043 & 3 & 1 & FALSE \\
x + x\_hat + skips & x + skips & 0.8312 & 0.8173 & 0.6 & 0.0139 & 3 & 1 & FALSE \\
x + x\_hat + skips & skips & 0.8312 & 0.8259 & 0.9 & 0.0053 & 3 & 1 & FALSE \\
x + x\_hat + skips & z only & 0.8312 & 0.8472 & 0.6 & -0.0160 & 3 & 1 & FALSE \\
x + x\_hat + skips & z + x & 0.8312 & 0.8205 & 0.5 & 0.0107 & 3 & 1 & FALSE \\
x + x\_hat + skips & z + x\_hat & 0.8312 & 0.8259 & 0.9 & 0.0053 & 3 & 1 & FALSE \\
\bottomrule
  \end{tabular}
  }
  \caption{\centering Pairwise permutation tests of decoder input ablations on unseen subjects in the SSVEP task}
\end{table*}

\end{document}